%% file: acl_latex.tex
\pdfoutput=1
\documentclass[11pt]{article}
\usepackage[final]{acl}
\usepackage{changepage}

\usepackage{times}
\usepackage{latexsym}
\usepackage[T1]{fontenc}
\usepackage[utf8]{inputenc}
\usepackage[russian,english]{babel}

\usepackage{microtype}
\usepackage{inconsolata}
\usepackage{graphicx}
\usepackage{tcolorbox}
\usepackage{amsmath}
\usepackage{subcaption}
\usepackage{url} 
\usepackage{booktabs}
\usepackage{amssymb}
\usepackage{todonotes}
\usepackage{multirow}
\usepackage{colortbl}
\usepackage{xcolor}
\usepackage{tikz}
\usepackage{array}
\usepackage{CJKutf8}
\usepackage{kotex}

\usepackage{textcomp}  
\usepackage{scalerel}  

\def\germanflag{\scalerel*{\includegraphics{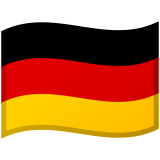}}{\textrm{\textbigcircle}}}
\def\ukflag{\scalerel*{\includegraphics{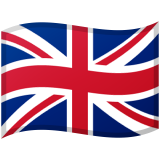}}{\textrm{\textbigcircle}}}

\definecolor{lightgreen}{RGB}{200,255,200}
\definecolor{mediumgreen}{RGB}{120,200,120}

\usepackage{CJKutf8}

\input{Chapters/macros}

\newcommand{\CLaSLogoSuperTiny}{
\begin{tikzpicture}[baseline=-0.25ex, scale=0.3]
    \definecolor{Ccol}{RGB}{52,152,219} 
    \definecolor{Lcol}{RGB}{231,76,60}  
    \definecolor{Scol}{RGB}{46,204,113} 
    \node[circle, draw=Ccol, fill=Ccol!20, minimum size=5mm, inner sep=0.5pt] (C) {\scriptsize C};
    \node[circle, draw=Lcol, fill=Lcol!20, minimum size=5mm, inner sep=0.5pt, right=0.3cm of C] (L) {\scriptsize L};
    \node[circle, draw=Scol, fill=Scol!20, minimum size=5mm, inner sep=0.5pt, right=0.3cm of L] (S) {\scriptsize S};
    
    \draw[->, Ccol] (C) to[out=30,in=150] (L);
    \draw[->, Lcol] (L) to[out=30,in=150] (S);

    \draw[thick, gray!60, rounded corners=2pt] (C.south) -- ++(0,-0.3) -- ++(2.4,0) -- ++(0,0.3);

\end{tikzpicture}
}
\title{\clasbench: \\ A Cross-Lingual Alignment and Steering Benchmark}
\newcommand{\affilsup}[1]{\rlap{\textsuperscript{\normalfont#1}}}

\author{
  Daniil Gurgurov\affilsup{1,2} \qquad Yusser Al Ghussin\affilsup{1,2} \qquad Tanja Bäumel\affilsup{1,2,3} \qquad Cheng-Ting Chou\affilsup{4} \\ 
  \textbf{Patrick Schramowski}\affilsup{2,5,6} \qquad  \textbf{Marius Mosbach}\affilsup{7,8} \qquad \textbf{Josef van Genabith}\affilsup{1,2} \qquad \textbf{Simon Ostermann}\affilsup{1,2,3} \\
  \\
  \small $^1$ Saarland University \enspace $^2$ German Research Center for Artificial Intelligence (DFKI) \\ 
  \small $^3$ Centre for European Research in Trusted AI (CERTAIN) \enspace $^4$ University of Illinois Urbana-Champaign \enspace $^5$ TU Darmstadt \\ 
  \small $^6$ hessian.AI \enspace $^7$ Mila - Quebec Artificial Intelligence Institute \enspace $^8$ McGill University \\
  \texttt{\small daniil.gurgurov@dfki.de}
}

\begin{document}
\maketitle

\begin{abstract}
Understanding and controlling the behavior of large language models (LLMs) is an increasingly important topic in multilingual NLP. Beyond prompting or fine-tuning, \emph{language steering}, i.e.,~manipulating internal representations during inference, has emerged as a more efficient and interpretable technique for adapting models to a target language. Yet, no dedicated benchmarks or evaluation protocols exist to quantify the effectiveness of steering techniques. 
We introduce \clasbench, a lightweight parallel-question benchmark for evaluating language-forcing behavior in LLMs across 32 languages, enabling systematic evaluation of multilingual steering methods. We evaluate a broad array of steering techniques, including residual-stream DiffMean interventions, probe-derived directions, language-specific neurons, PCA/LDA vectors, Sparse Autoencoders, and prompting baselines. Steering performance is measured along two axes: language control and semantic relevance, combined into a single harmonic-mean steering score. We find that across languages simple residual-based DiffMean method consistently outperforms all other methods. Moreover, a layer-wise analysis reveals that language-specific structure emerges predominantly in later layers and steering directions cluster based on language family. 
\clasbench is the first standardized benchmark for multilingual steering, enabling both rigorous scientific analysis of language representations and practical evaluation of steering as a low-cost adaptation alternative.\looseness-1

\end{abstract}

\input{Chapters/01_Introduction}
\input{Chapters/02_Data_new}
\input{Chapters/03_SteeringMethods_new}

\input{Chapters/05_Results}
\input{Chapters/06_RelatedWork}
\input{Chapters/08_Conclusion}

\section*{Limitations}
Our work has several limitations. First, due to computational constraints, we use varying amounts of data across methods: DiffMean and LAPE process 10M tokens per language, while PCA and LDA use 500K and 100K tokens respectively, as these methods require computing covariance matrices that scale quadratically with sample size. We follow established practices for each method \cite{marks2023geometry, tang2024language}, but this variation may affect comparability. Second, SAE-based steering is limited to layers with publicly available pretrained SAEs (layers 4, 12, 18, 20, 25 for \texttt{Llama-3.1-8B-Instruct}), preventing exhaustive layer-wise analysis. Additionally, we were unable to evaluate SAE-based steering for \texttt{Aya-Expanse-8B} due to the absence of publicly available pretrained SAEs for this model. Third, while \clasbench covers 32 typologically diverse languages, many of the world's languages remain unrepresented, particularly those with limited digital resources. Finally, we evaluate only instruction-tuned models; base models may exhibit different steering dynamics.

\section*{Acknowledgments}

This research was supported by \textit{lorAI - Low Resource Artificial Intelligence}, a project funded by the European Union under \href{https://doi.org/10.3030/101136646}{GA No.101136646}, and by the German Federal Ministry of Research, Technology and Space (BMFTR) as part of the project TRAILS (01IW24005). We also thank Masha Fedzechkina for her valuable feedback on an early draft of the paper.

\bibliography{custom}

\appendix
\newpage

\input{Appendices/A}

\end{document}

%% file: Chapters/macros.tex
\usepackage{xspace}
\newcommand{\clasbench}{\texttt{CLaS-Bench}\xspace}

%% file: Chapters/01_Introduction.tex
\section{Introduction}
\label{sec:introduction}

\begin{figure}[t!]
\centering
\begin{tikzpicture}[
    node distance=1.8cm,
    every node/.style={font=\footnotesize},
    input/.style={rectangle, rounded corners=4pt, draw=black!40, fill=blue!12, minimum width=3.2cm, align=center, inner sep=0.4cm},
    process/.style={rectangle, rounded corners=4pt, draw=black!40, fill=yellow!12, minimum width=2.2cm, minimum height=1.2cm, align=center, inner sep=0.35cm},
    metric/.style={rectangle, rounded corners=4pt, draw=black!40, minimum height=0.7cm, align=center, inner sep=0.25cm, font=\footnotesize\bfseries},
    arrow/.style={->, thick, black!50, shorten >=2pt, shorten <=2pt}
]

  \node[input, fill=blue!15] (inputs) at (1.6, 2.4) {
  \textbf{\CLaSLogoSuperTiny}  \\
  \rule{3cm}{0.5pt}\\
  \textbf{32L} $\times$ \textbf{70Q} = \textbf{2.24K} \textit{(Evals per Lang)}
};

  \draw[arrow] (1.6, 1.5) -- (1.6, 0.4);

  \node[process, fill=yellow!15] (task) at (1.6, 0.1) {
    \textbf{Steering \ukflag $\to$ \germanflag}\\
    \footnotesize How are you? \textit{(Source \ukflag)} \\
    {\tiny $\curvearrowleft\!\lfloor$ \texttt{Language Model} $\rfloor\!\curvearrowright$} \\
    {\tiny Steering Methods: {\tiny \color{blue}$\mathcal{E}$ \color{green}$\mathcal{T}$ \color{red}$\odot$ \color{orange}$\vec{\Delta}$ \color{purple}$\mathbf{w}$ \color{cyan}$\mathbf{u}$ \color{blue}${\mathbf{v}}$ \color{brown}$\vec{\Delta}_{S}$}}\\
    {Mir geht es gut. \textit{(Target \germanflag)}}
  };

  \node[metric, fill=orange!15, minimum width=0.9cm] (m1) at (0.3, -1.9) {$\mathbf{F}$};
  \node[metric, fill=purple!15, minimum width=0.9cm] (m2) at (1.6, -1.9) {$\mathbf{R}$};
  \node[metric, fill=red!15, draw=black!60, line width=1pt, minimum width=0.9cm] (m3) at (2.9, -1.9) {$\mathbf{S}$};

  \draw[arrow] (1.2, -1.1) -- (0.3, -1.5);
  \draw[arrow] (1.6, -1.1) -- (1.6, -1.6);
  \draw[arrow] (2.0, -1.1) -- (2.9, -1.5);
  
  \node[font=\tiny, black!60, align=center] at (0.3, -2.4) {Forcing};
  \node[font=\tiny, black!60, align=center] at (1.6, -2.4) {Relevance};
  
  \node[font=\tiny, black!60, align=center] at (2.9, -2.4) {$\text{HM}(\mathbf{F}, \mathbf{R})$};
\end{tikzpicture}

\caption{
\clasbench pipeline:
Multilingual inputs consisting of 70 parallel questions (\textbf{Q}) across 32 languages (\textbf{L}) are evaluated per target language. Each input is passed to an LLM, which is steered with a selected method. 
The steered model output is evaluated along two axes: \textbf{language forcing F} (whether generation switches to the intended target language) and \textbf{output relevance R} (whether response is related to the input). These metrics are combined via a harmonic mean into a single \textbf{steering score S}.}
\label{fig:benchmark_overview}
\end{figure}
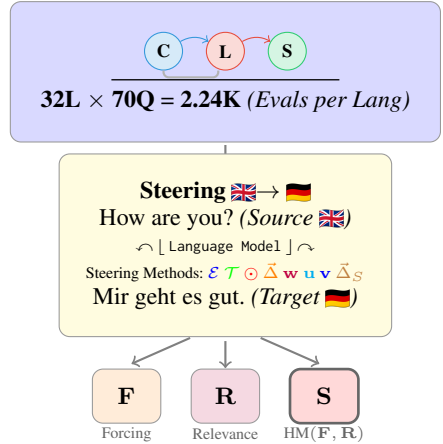

As our understanding of the internal mechanisms of large language models (LLMs) advances, increasing attention is given to methods that exploit these internal mechanisms to control model behavior. This research direction, often referred to as \textit{actionable interpretability} \cite{mosbach2024insights}, has increasingly incorporated techniques collectively called \textit{steering}, i.e., the manipulation of model weights or activations to guide models toward desired outputs \cite{subramani2022extracting}. Unlike related techniques such as fine-tuning, steering methods are typically applied \textit{at inference time}, positioning them as a more lightweight, direct alternative for adapting models without re-training. Representation-based steering, in particular, intervenes directly on a model's hidden representations (e.g., residual streams, or latent features) to induce desired behaviors and has been applied successfully to mitigate sycophancy \cite{panickssery2023steering}, improve truthfulness \cite{li2024inferencetimeinterventionelicitingtruthful}, and reduce toxicity \cite{suau2024whispering}, showing that internal representations can be used for controllability.


\begin{table*}[th!]
\centering
\small
\begin{tabular}{p{1.25cm} p{3.5cm} p{10cm}}
\toprule
\textbf{Domain} & \textbf{English Prompt} & \textbf{Translations (Sampling)} \\
\midrule
\multirow{2}{*}{\textbf{Knowledge}} 
& \multirow{2}{*}{\parbox{3.5cm}{\small ``Can you explain the basics of quantum computing?''}}
& \textbf{Spanish:} ``¿Puedes explicar los conceptos básicos de la computación cuántica?'' \\
& & \textbf{Norwegian:} ``Kan du forklare grunnleggende kvanteberegning?'' \\
\midrule
\multirow{2}{*}{\textbf{Reasoning}}
& \multirow{2}{*}{\parbox{3.5cm}{\small ``How many words are spoken daily on Earth? Try to explain your answer. Your explanation should take the reader through your reasoning step-by-step.''}}
& \textbf{German:} ``Wie viele Wörter werden täglich auf Erden gesprochen? Versuche, deine Antwort zu erklären. Deine Erklärung sollte den Leser Schritt für Schritt durch dein Denken führen.'' \\
& & \textbf{Japanese:} \begin{CJK}{UTF8}{min}``地球上で毎日いくつの単語が発せられていますか？あなたの答えを説明してみてください。あなたの説明は読み手をあなたの思考回路に一歩ずつ案内していくようなものであるべきです。''\end{CJK} \\
\midrule
\multirow{2}{*}{\textbf{Creative}}
& \multirow{2}{*}{\parbox{3.5cm}{\small ``How would you introduce yourself as a medieval knight at a royal banquet?''}}
& \textbf{French:} ``Comment vous présenteriez-vous en tant que chevalier médiéval dans un banquet royal?'' \\
& & \textbf{Korean:} \begin{CJK}{UTF8}{mj}``중세 기사로서 왕실 연회에서 자신을 어떻게 소개하시겠습니까?''\end{CJK} \\
\midrule
\multirow{2}{*}{\textbf{Opinion}}
& \multirow{2}{*}{\parbox{3.5cm}{\small ``Do we have a moral obligation to explore space, or should we focus on solving Earth's problems first?''}}
& \textbf{Dutch:} ``Har vi en moralsk forpligtelse til at udforske rummet, eller skal vi først fokusere på at løse Jordens problemer?'' \\
& & \textbf{Polish:} ``Czy mamy moralny obowiązek odkrywania przestrzeni kosmicznej, czy powinniśmy najpierw skupić się na rozwiązywaniu problemów Ziemi?'' \\
\midrule
\multirow{2}{*}{\textbf{Writing}}
& \multirow{2}{*}{\parbox{3.5cm}{\small ``Can you help me write a formal email to a potential business partner proposing a joint venture?''}}
& \textbf{Chinese:} \begin{CJK}{UTF8}{gbsn}``你能帮我写一封正式邮件，向潜在的商业伙伴提议合资吗？''\end{CJK} \\
& & \textbf{Russian:} \foreignlanguage{russian}{``Можете ли вы помочь мне написать формальное письмо потенциальному бизнес-партнёру с предложением совместного предприятия?''} \\
\bottomrule
\end{tabular}
\caption{Examples of questions included in the \clasbench benchmark spanning all domains.}
\label{tab:eval_examples_app}
\end{table*}

A prominent application of representation steering is control over the language of generation. Recent work probing LLMs has revealed language-specific features and neurons \cite{tang2024language, zhao2024largelanguagemodelshandle, kojima2024multilingual}, and these insights have been used both to improve downstream performance and control cross-lingual behavior \cite{gurgurov2025languagearithmeticssystematiclanguage, chou2025causal}. Steering is particularly promising for multilingual adaptation, enabling targeted language control without requiring costly retraining. However, despite such advances, there is still \textbf{no standard evaluation} framework for language steering in language models. Existing benchmarks \cite{wu2025axbench, mueller2025mib} focus exclusively on conceptual steering tasks in English, leaving multilingual and cross-lingual settings unexplored.

This gap motivates two central research questions. First, how effective is steering for controlling output language compared to established approaches? Second, does steering perform equally well across languages, given that most LLMs are predominantly pretrained on English? More broadly, mechanistic interpretability research remains predominantly English-centric: most analyses, circuits, and intervention techniques are developed for and evaluated only in English. Extending these to a broad set of languages is thus essential for scientific understanding and enabling actionable interpretability in truly multilingual settings.

To close this gap, we introduce \clasbench, a lightweight benchmark for evaluating multilingual and cross-lingual language steering. \clasbench covers 32 languages and 70 parallel questions with answers per language (i.e., over \textbf{71,680} potential cross-lingual question-answer pairs), enabling controlled, language-by-language evaluation of 
\textit{steering}, i.e., the informed manipulation of language components, such as neurons or latent directions. Crucially, our evaluation emphasizes \textit{cross-lingual} steering, where the prompt (source) and desired output (target) languages differ--capturing interesting multilingual use cases.

We use \clasbench to compare a broad set of steering approaches, including neuron-level interventions \cite{tang2024language}, residual-stream difference-in-means vectors \cite{marks2023geometry}, probe-derived directions \cite{li2024inferencetimeinterventionelicitingtruthful}, and vectors from LDA \cite{balakrishnama1998linear}, PCA \cite{abdi2010principal}, and Sparse Autoencoders \cite{bricken2023monosemanticity}, against prompting baselines across several LLMs. We measure steering success along two orthogonal dimensions: (1) \textbf{Forcing}, whether the model produces text in the intended language, and (2) \textbf{Relevance}, whether the output remains conceptually appropriate; these are combined into an overall steering score via the harmonic mean.

Our experiments reveal three key findings. First, representation-based steering, particularly DiffMean on residual activations, consistently outperforms all other tested methods, including prompting baselines, across most evaluated languages. Second, prompting exhibits failures for specific languages, while DiffMean succeeds across most languages; moreover, we find steering earlier layers is effective with low intervention strengths, whereas later layers require higher strengths. Third, language-specific representations concentrate in later layers (roughly layers 16--32), and typologically related languages cluster geometrically in representation space. 

%% file: Chapters/02_Data_new.tex
\section{Benchmark Design}
\label{sec:design}


\paragraph{Languages.}
\clasbench covers a typologically and geographically diverse subset of 32 languages. We provide details on the languages covered in Appendix~\ref{app:langs}. The selection aims to balance high-resource and low-resource languages while spanning different language families and scripts, which enables evaluation across a broad linguistic spectrum. This diversity ensures that \clasbench evaluates language steering across a wide range of typological phenomena, scripts, and resource levels.  The diversity also enables focusing on challenging cross-lingual settings, particularly for low-resource and non-Latin-script languages, where LLMs often struggle \cite{joshi-etal-2020-state}.

\paragraph{Evaluation data.}
We use a curated subset of 70 diverse open-ended questions from the Vicuna dataset \cite{vicuna2023}, originally introduced by \citet{tang2024language}. These questions (see Table \ref{tab:eval_examples_app} for examples) cover a wide range of conversational domains, which we identify and label manually (reasoning, knowledge, personal opinion, creative, and professional writing), and are designed to elicit multi-sentence outputs. 

We translate all English questions into the remaining 31 languages using the Google Translate API \cite{wu2016googlesneuralmachinetranslation}, resulting in a parallel dataset of 70 questions per language across 32 languages. All translations are proofread and corrected by native speakers to ensure fluency, idiomaticity, and semantic fidelity to the English source prompts (see Appendix~\ref{app:data_curation} for details on quality assurance and the proofreading protocol). For evaluation, each question in each source language can be paired with an answer in any target language, yielding $70 \times 32 = 2{,}240$ instances per target language. 

\paragraph{Task definition.}

Let $\mathcal{L}$ be the set of all languages \clasbench covers. Given a question $x_s$ in source language $s \in \mathcal{L}$, the task is to generate an answer $y_t$ in target language $t \in \mathcal{L}$.
$M_\theta(x)$ is a language model with fixed parameters $\theta$ and $h_{\ell}(x) \in \mathbb{R}^{d_{\ell}}$ is the hidden representation at layer $\ell$ computed from input $x$, where $d_{\ell}$ is the dimensionality. We use $h_{\ell}[i]$ to index the $i$-th element (neuron) of $h_{\ell}$.

A steering method $S(\cdot)$ modifies the generation process either indirectly by changing the input or directly by intervening on hidden representations:
\[
\hat{y}_t = M_\theta(S(x_s))~.
\]
Here, $S$ either transforms the input: $x_s \to x_s'$, or intervenes on the hidden representations at layer $\ell$, i.e., replaces the hidden representation by the intervention $\delta_{\ell}$. The parameters $\theta$ remain fixed. The goal is to ensure $\hat{y}_t$ is in the target language $t$ while preserving the semantic content of $x_s$.
The overall pipeline is illustrated in Figure~\ref{fig:benchmark_overview}.

\paragraph{Evaluation metrics.}
We assess steering effectiveness along two complementary dimensions:

\begin{itemize}
    \item \textbf{Language Forcing Success (LFS).} 
    This measure indicates the overall success of a method to force a specific language. We apply the FastText LID classifier \cite{joulin2016bag} to detect the language of generated outputs, which provides good coverage for the languages in our benchmark. We report both overall success rate and per-language breakdown:
    \[
    LFS = \frac{\text{\# outputs in target language}}{\text{total \# outputs}} \in [0,1].
    \]

    \item \textbf{Output Relevance (OR).} 
    This score measures the semantic fidelity of the answer to the question. We compute this using an LLM-as-a-judge evaluation with Qwen-3-8B \cite{qwen3technicalreport}, which demonstrates strong multilingual performance. Each output is scored 0 (unrelated or gibberish), 1 (partially relevant or incomplete), or 2 (clearly relevant and coherent), and we report the normalized average relevance:
    \[
    OR = \frac{1}{N} \sum_{i=1}^{N} \frac{\text{score}_i}{2} \in [0,1],
    \]
    where \(N\) is the number of evaluated outputs. The judging protocol employed is similar to the one from \citet{wu2025axbench} and is presented in Appendix \ref{app:judge}.

\end{itemize}

We combine these in the \textbf{Language Steering Score} which computes the harmonic mean of LFS and OR:
    \[
    \text{LSS} = \frac{2 \cdot LFS \cdot OR}{LFS + OR},
    \]
    which penalizes cases where one of the two metrics is very low relative to the other.

%% file: Chapters/03_SteeringMethods_new.tex
\section{Experimental Setup}

\paragraph{Models.}

We evaluate \clasbench on two LLMs: \texttt{Llama-3.1-8B-Instruct} \cite{grattafiori2024llama}, a widely used mid-sized foundation model, and \texttt{Aya-Expanse-8B} \cite{dang2024ayaexpansecombiningresearch}, a widely used multilingual alternative.

\paragraph{Steering methods.}
We benchmark a multitude of steering methods, spanning both prompting-based and representation-based interventions. The data for designing representation-based interventions is sourced from CulturaX \cite{nguyen2023culturaxcleanedenormousmultilingual}. Below, $\alpha$ denotes the \textit{steering strength} in all methods.

\vspace{0.2em}
\noindent\textbf{(I)} \quad \(\color{blue}{\mathcal{E}}\) \quad \textbf{Prompting with Language Specification (Baseline-I).} Adding explicit instructions to respond in the target language with the instructions in English, e.g., \textit{Question + "Respond in German"} for steering towards German.

\vspace{0.2em}
\noindent\textbf{(II)} \quad \(\color{green}{\mathcal{T}}\) \quad \textbf{Prompting with Language Specification (Baseline-II).} Adding explicit instructions to respond in the target language with instructions in the target language, e.g., \textit{Question + "Antworte auf Deutsch"} for steering towards German. 

\vspace{0.2em}
\noindent\textbf{(III)} \quad \(\color{red}{\odot}\) \quad \textbf{Neuron-Based Steering (LAPE).} Identifying language-sensitive neurons \cite{tang2024language} by analyzing activation patterns across 10M tokens per language. We compute activation probabilities $p_{\ell,h}^{\text{lang}}$ for each neuron $h_{\ell}[i]$ in layer $\ell$ across languages, then apply entropy filtering to select language sensitive-neurons $\mathcal{N}_{\mathrm{selected}}$ with low cross-lingual entropy (high language selectivity). The intervention is defined as 
\[\delta_{\ell} = \sum_{i \in \mathcal{N}_{\mathrm{selected}}} \delta_{\ell,i} \cdot \mathbf{e}_i~,\] 
where $\mathbf{e}_i$ is the standard basis vector. Let $\bar{a}_{\ell,h}^{\text{lang}}$ be the average activation of neuron $h_{\ell}[i]$ in layer $\ell$ for the target language. Selected neurons are manipulated via two intervention mechanisms: 
\begin{enumerate}
    \item \textit{additive}: $\delta_{\ell,i} = \alpha \cdot \bar{a}_{\ell,i}^{\mathrm{target}} + h_{\ell}[i]$
    \item \textit{replacement}: $\delta_{\ell,i} = \alpha \cdot \bar{a}_{\ell,i}^{\mathrm{target}}$
\end{enumerate}
Non-target language neurons are optionally deactivated by zeroing them out. 

\vspace{0.2em}
\noindent\textbf{(IV)} \quad \(\color{orange}{\vec{\Delta}}\) \quad \textbf{DiffMean Steering Vectors on Residual Activations.} 
Computing language-specific average activations across the residual stream \cite{marks2023geometry} for 10M tokens per language. We define the hidden intervention as \[\delta_{\ell} = h_{\ell} + \alpha \cdot \frac{\vec{\Delta}_{\ell}}{\|\vec{\Delta}_{\ell}\|_2} ~,\]
where $\vec{\Delta}_{\ell} = \bar{h}_{\ell}^{\mathrm{target}} - \bar{h}_{\ell}^{\mathrm{source}}$ and $\bar{h}_{\ell}^{\mathrm{lang}}$ is the average activation at layer $\ell$ for language \textit{lang}.


\vspace{0.2em}
\noindent\textbf{(V)} \quad \(\color{purple}{\mathbf{w}}\) \quad \textbf{Probe-based Steering Vectors on Residual Streams.} Training linear probes \cite{li2024inferencetimeinterventionelicitingtruthful} to classify target language representations against negative languages. For each layer $\ell$, we train a binary classifier $\text{Probe}_{\ell}: \mathbb{R}^{d} \to [0,1]$ on balanced datasets of target language activations (positive class) and negative language activations (negative class), each consisting of 100K samples, optimizing binary cross-entropy loss. The probe weight vector $\mathbf{w}_{\ell} \in \mathbb{R}^{d}$ encodes the direction in the residual stream that discriminates the target language. We then define the intervention as 
\[\delta_{\ell} = h_{\ell} +\alpha \cdot \frac{\mathbf{w}_{\ell}}{\|\mathbf{w}_{\ell}\|_2} ~.\] 


\vspace{0.2em}
\noindent\textbf{(VI)} \quad \(\color{cyan}{\mathbf{u}}\) \quad \textbf{PCA-based Steering Vectors on Residual Streams.} Computing language-specific subspaces through Principal Component Analysis (PCA) \cite{abdi2010principal} on residual stream activations. We collect activations from each target language across 500K tokens and center them by subtracting the mean. For each layer $\ell$, we apply PCA to obtain the top $k=20$ principal components $U_{\ell} \in \mathbb{R}^{k \times d}$, which span the subspace of maximum variance for that language. During inference, given a hidden state $h_{\ell} \in \mathbb{R}^{d}$, we project onto the language subspace via $\text{proj}_{\ell} = U_{\ell} h_{\ell}^T \in \mathbb{R}^{k}$, then reconstruct in the original space: $\mathbf{u}_{\ell} = U_{\ell}^T \text{proj}_{\ell} \in \mathbb{R}^{d}$. 
We normalize $\mathbf{u}_{\ell}$ to decouple steering magnitude from component strength. 
The intervention is then defined as
\[
\delta_{\ell} = h_{\ell} + \alpha \cdot \frac{\mathbf{u}_{\ell}} {\|\mathbf{u}_{\ell}\|_2} ~.
\]


\begin{table*}[t!]
\centering
\small
\begin{tabular}{l|cccccccc}
\toprule
\textbf{Lang.} & $\color{blue}\mathcal{E}$ Base.-I & $\color{green}\mathcal{T}$ Base.-II & $\color{orange}\vec{\Delta}$ DiffM. & $\color{purple}\mathbf{w}$ Probe & $\color{cyan}\mathbf{u}$ PCA & $\color{brown}\vec{\Delta}$ SAE-DM. & $\color{blue}\mathbf{v}$ LDA & $\color{red}\odot$ LAPE \\
\midrule
ar & \cellcolor{gray!40}62.9 & \cellcolor{gray!40}54.0 & \cellcolor{yellow!25}\textbf{88.8} & 14.5 & 16.3 & 49.5 & 34.9 & \cellcolor{orange!20}77.2 \\
bo & \cellcolor{gray!40}33.5 & \cellcolor{gray!40}\textbf{38.2} & \cellcolor{yellow!25}8.5 & 5.6 & 6.7 & 6.1 & 4.8 & \cellcolor{orange!20}7.6 \\
cs & \cellcolor{gray!40}69.1 & \cellcolor{gray!40}75.2 & \cellcolor{yellow!25}\textbf{92.3} & 28.6 & 16.6 & 39.5 & 27.3 & \cellcolor{orange!20}89.1 \\
da & \cellcolor{gray!40}80.6 & \cellcolor{gray!40}47.7 & \cellcolor{yellow!25}\textbf{90.1} & 35.8 & 17.4 & 32.7 & 8.8 & \cellcolor{orange!20}53.7 \\
de & \cellcolor{gray!40}70.6 & \cellcolor{gray!40}56.6 & \cellcolor{yellow!25}95.4 & 83.4 & 14.7 & 51.2 & 0.0 & \cellcolor{orange!20}\textbf{96.5} \\
el & \cellcolor{gray!40}82.3 & \cellcolor{gray!40}76.4 & \cellcolor{yellow!25}\textbf{88.4} & 9.0 & 15.5 & 39.0 & 6.1 & \cellcolor{orange!20}87.1 \\
en & \cellcolor{gray!40}7.1 & \cellcolor{gray!40}7.1 & \cellcolor{yellow!25}95.9 & 95.7 & 11.0 & 51.2 & 51.3 & \cellcolor{orange!20}\textbf{98.8} \\
es & \cellcolor{gray!40}74.7 & \cellcolor{gray!40}65.3 & \cellcolor{yellow!25}\textbf{94.7} & 78.7 & 20.6 & 57.7 & 47.6 & \cellcolor{orange!20}94.0 \\
fa & \cellcolor{gray!40}38.2 & \cellcolor{gray!40}53.3 & \cellcolor{yellow!25}79.0 & 13.2 & 13.9 & 47.4 & 10.3 & \cellcolor{orange!20}\textbf{85.1} \\
fr & \cellcolor{gray!40}66.8 & \cellcolor{gray!40}66.6 & \cellcolor{yellow!25}96.4 & 89.7 & 19.3 & 64.6 & 23.3 & \cellcolor{orange!20}\textbf{96.5} \\
hi & \cellcolor{gray!40}52.5 & \cellcolor{gray!40}68.0 & \cellcolor{yellow!25}90.5 & 13.4 & 11.8 & 53.8 & 0.0 & \cellcolor{orange!20}\textbf{93.2} \\
id & \cellcolor{gray!40}65.7 & \cellcolor{gray!40}76.8 & \cellcolor{yellow!25}\textbf{94.7} & 59.0 & 18.3 & 17.9 & 0.0 & \cellcolor{orange!20}\textbf{94.7} \\
it & \cellcolor{gray!40}66.2 & \cellcolor{gray!40}71.1 & \cellcolor{yellow!25}94.7 & 77.5 & 23.4 & 53.4 & 15.8 & \cellcolor{orange!20}\textbf{95.9} \\
ja & \cellcolor{gray!40}48.0 & \cellcolor{gray!40}72.6 & \cellcolor{yellow!25}\textbf{84.6} & 35.5 & 14.3 & 10.0 & 16.9 & \cellcolor{orange!20}83.5 \\
ka & \cellcolor{gray!40}\textbf{80.3} & \cellcolor{gray!40}67.9 & \cellcolor{yellow!25}32.9 & 6.1 & 11.8 & 6.2 & 3.8 & \cellcolor{orange!20}27.4 \\
kk & \cellcolor{gray!40}\textbf{63.7} & \cellcolor{gray!40}56.6 & \cellcolor{yellow!25}42.1 & 6.3 & 6.1 & 8.8 & 6.2 & \cellcolor{orange!20}28.7 \\
ko & \cellcolor{gray!40}42.7 & \cellcolor{gray!40}66.4 & \cellcolor{yellow!25}\textbf{89.8} & 39.8 & 16.9 & 35.6 & 20.9 & \cellcolor{orange!20}79.8 \\
mt & \cellcolor{gray!40}\textbf{68.7} & \cellcolor{gray!40}64.5 & \cellcolor{yellow!25}15.6 & 5.8 & 7.3 & 6.2 & 5.7 & \cellcolor{orange!20}20.8 \\
nl & \cellcolor{gray!40}67.7 & \cellcolor{gray!40}66.8 & \cellcolor{yellow!25}\textbf{95.3} & 45.9 & 13.8 & 43.5 & 7.7 & \cellcolor{orange!20}95.0 \\
no & \cellcolor{gray!40}63.4 & \cellcolor{gray!40}55.8 & \cellcolor{yellow!25}\textbf{88.8} & 29.3 & 18.0 & 38.7 & 4.4 & \cellcolor{orange!20}69.5 \\
pl & \cellcolor{gray!40}53.9 & \cellcolor{gray!40}53.9 & \cellcolor{yellow!25}\textbf{92.5} & 33.9 & 10.7 & 41.4 & 46.5 & \cellcolor{orange!20}89.9 \\
pt & \cellcolor{gray!40}61.1 & \cellcolor{gray!40}78.9 & \cellcolor{yellow!25}\textbf{95.8} & 84.6 & 18.9 & 48.3 & 55.9 & \cellcolor{orange!20}95.6 \\
ro & \cellcolor{gray!40}70.2 & \cellcolor{gray!40}83.7 & \cellcolor{yellow!25}\textbf{93.0} & 57.9 & 13.8 & 35.0 & 6.4 & \cellcolor{orange!20}77.3 \\
ru & \cellcolor{gray!40}57.4 & \cellcolor{gray!40}69.2 & \cellcolor{yellow!25}\textbf{97.0} & 82.5 & 16.7 & 67.6 & 56.8 & \cellcolor{orange!20}95.5 \\
sk & \cellcolor{gray!40}\textbf{83.5} & \cellcolor{gray!40}52.9 & \cellcolor{yellow!25}77.7 & 31.8 & 7.8 & 17.9 & 10.9 & \cellcolor{orange!20}65.9 \\
sv & \cellcolor{gray!40}63.7 & \cellcolor{gray!40}63.6 & \cellcolor{yellow!25}\textbf{93.7} & 47.6 & 17.0 & 39.3 & 60.2 & \cellcolor{orange!20}89.8 \\
sw & \cellcolor{gray!40}\textbf{83.8} & \cellcolor{gray!40}79.6 & \cellcolor{yellow!25}49.8 & 8.4 & 2.1 & 6.2 & 6.2 & \cellcolor{orange!20}52.7 \\
th & \cellcolor{gray!40}73.4 & \cellcolor{gray!40}86.0 & \cellcolor{yellow!25}\textbf{91.2} & 24.2 & 12.4 & 48.0 & 9.8 & \cellcolor{orange!20}84.5 \\
tr & \cellcolor{gray!40}77.5 & \cellcolor{gray!40}73.0 & \cellcolor{yellow!25}\textbf{84.5} & 23.3 & 14.7 & 19.8 & 7.2 & \cellcolor{orange!20}84.1 \\
uk & \cellcolor{gray!40}67.3 & \cellcolor{gray!40}80.7 & \cellcolor{yellow!25}\textbf{92.9} & 13.5 & 15.4 & 69.7 & 16.1 & \cellcolor{orange!20}91.6 \\
vi & \cellcolor{gray!40}79.9 & \cellcolor{gray!40}88.4 & \cellcolor{yellow!25}\textbf{96.9} & 56.1 & 14.7 & 11.3 & 11.1 & \cellcolor{orange!20}95.0 \\
zh & \cellcolor{gray!40}60.4 & \cellcolor{gray!40}76.6 & \cellcolor{yellow!25}\textbf{97.6} & 92.3 & 12.2 & 85.4 & 87.6 & \cellcolor{orange!20}96.2 \\
\midrule
Avg. & \cellcolor{gray!40}67.7 & \cellcolor{gray!40}67.3 & \cellcolor{yellow!25}\textbf{84.5} & 48.6 & 15.1 & 42.3 & 23.6 & \cellcolor{orange!20}80.1 \\
\bottomrule
\end{tabular}
\caption{Language steering scores 
across all methods for 32 languages for \texttt{Llama-3.1-8B-Instruct}. Individual scores for language forcing and output relevance metrics are in Appendix \ref{app:scores}.
\textcolor{gray!80}{Gray} columns correspond to baselines. \textcolor{yellow!80}{Yellow} and \textcolor{orange!80}{orange} to the best performing methods.
}
\label{tab:methods_languages}
\end{table*}

\vspace{0.2em}
\noindent\textbf{(VII)} \quad \(\color{blue}{\mathbf{v}}\) \quad \textbf{LDA-based Steering Vectors on Residual Streams.} 
Computing language-discriminative steering vectors through Linear Discriminant Analysis (LDA) \cite{balakrishnama1998linear} on residual stream activations. We collect activations from the target language and multiple negative languages across 100K tokens each. For each layer $\ell$, we formulate a binary classification problem where the positive class consists of target language activations and the negative class consists of equal balanced samples from all other languages. LDA finds the optimal linear direction that maximizes class separability by computing: $\mathbf{v}_{\ell} = \Sigma_w^{-1} (\boldsymbol{\mu}_{\text{tgt}} - \boldsymbol{\mu}_{\text{other}})$, where $\Sigma_w$ is the within-class covariance matrix and $\boldsymbol{\mu}_{\text{tgt}}, \boldsymbol{\mu}_{\text{other}}$ are the class means. 
The intervention is then defined as
\[
\delta_{\ell} = h_{\ell} + \alpha \cdot \frac{\mathbf{v}_{\ell}}{\|\mathbf{v}_{\ell}\|_2} ~.
\]
    
\vspace{0.2em}
\noindent\textbf{(VIII)} \quad \(\color{brown}{\vec{\Delta}_{S}}\) \quad \textbf{DiffMean on Sparse Autoencoder Layer.} Computing language-specific average activations in the sparse autoencoder (SAE) latent space for 10M tokens per language. We utilize pre-trained SAEs from \citet{li2025trainingsuperiorsparseautoencoders} for Llama-3.1-8B-Instruct. For each SAE layer $\ell \in \mathcal{L}$ (where $\mathcal{L}$ indexes the subset of model layers with trained SAEs), we encode residual stream activations $h_\ell$ into sparse representations via a JumpReLU encoder: $f_\ell = \text{JumpReLU}(W_{\text{enc}} h_\ell + b_{\text{enc}}) \in \mathbb{R}^{d_{\text{SAE}}}$, where JumpReLU$(z) = z \cdot \mathbf{1}[z > \theta]$ with learned threshold $\theta$. The steering vector is computed as the difference between target and source language means in this sparse space: $\vec{\Delta}_{\ell} = \bar{f}_{\ell}^{\text{target}} - \bar{f}_{\ell}^{\text{source}}$. During inference, we hook into the input to intercept the residual stream from layer $\ell$. The combined hidden state is encoded, steered in SAE latent space with $\ell_2$-normalized strength $\alpha$, then decoded back: 
\[
\delta_{\ell} = W_{\text{dec}}\left(f_\ell + \alpha \cdot \frac{\vec{\Delta}_{\ell}}{\|\vec{\Delta}_{\ell}\|_2}\right) 
+ \epsilon ~.
\]
Here, $\epsilon$ 
is a reconstruction error correction term that preserves information not captured by the SAE.\footnote{Bias terms are left out for the sake of readability.} 


%% file: Chapters/05_Results.tex
\begin{figure*}[t!]
\centering
\setlength{\tabcolsep}{1pt}
\renewcommand{\arraystretch}{0.6} 
\resizebox{\textwidth}{!}{
\begin{tabular}{lcccccc}
& {\scriptsize\(\color{orange}{\vec{\Delta}}\) \textbf{DiffMean}} 
& {\scriptsize\(\color{purple}{\mathbf{w}}\) \textbf{Probe}} 
& {\scriptsize\(\color{blue}{\mathbf{v}}\) \textbf{LDA}}
& {\scriptsize\(\color{red}{\odot}\) \textbf{LAPE}} 
& {\scriptsize\(\color{cyan}{\mathbf{u}}\) \textbf{PCA}} 
& {\scriptsize\(\color{brown}{\vec{\Delta}_{S}}\) \textbf{SAE-DiffM}} \\[0pt]
\raisebox{0.3cm}{\rotatebox{90}{\scriptsize \textbf{Lang. Forcing}}} &
\includegraphics[width=0.16\textwidth]{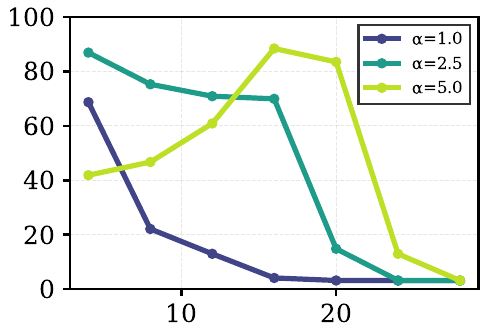} &
\includegraphics[width=0.16\textwidth]{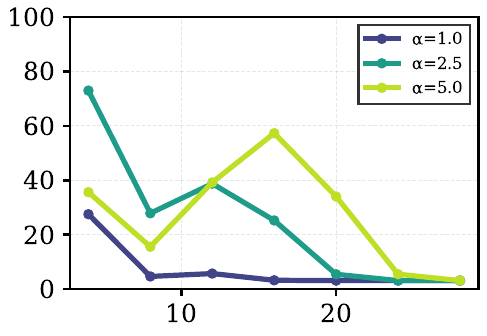} &
\includegraphics[width=0.16\textwidth]{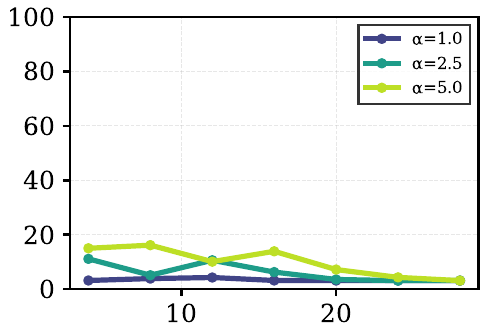} &
\includegraphics[width=0.16\textwidth]{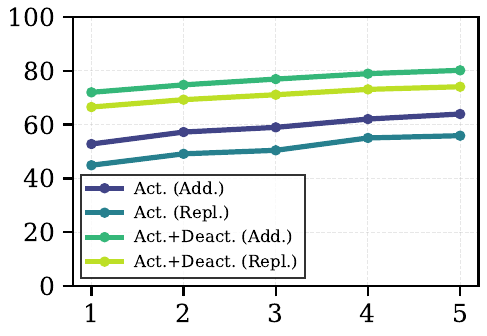} &
\includegraphics[width=0.16\textwidth]{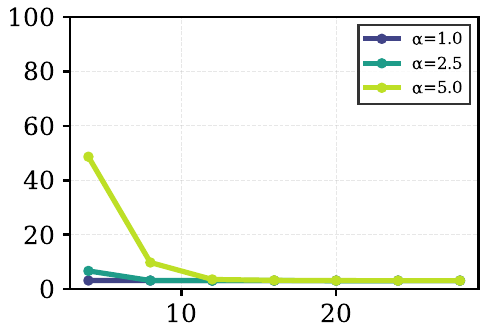} &
\includegraphics[width=0.16\textwidth]{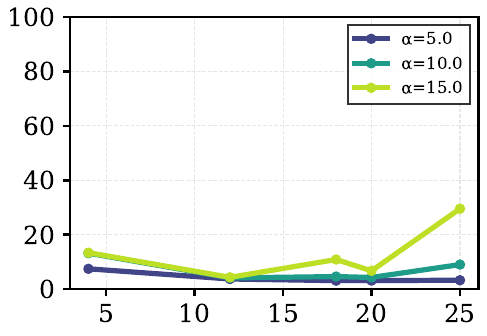} \\[-1pt]
\raisebox{0.3cm}{\rotatebox{90}{\scriptsize \textbf{Judge Quality}}} &
\includegraphics[width=0.16\textwidth]{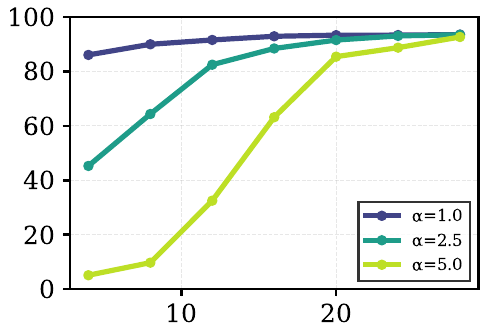} &
\includegraphics[width=0.16\textwidth]{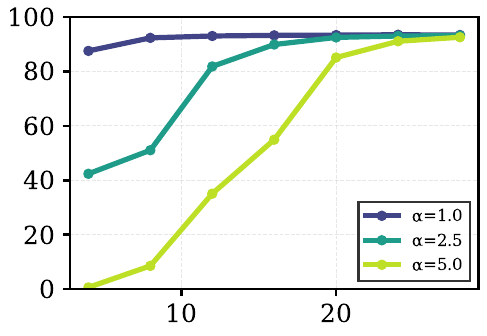} &
\includegraphics[width=0.16\textwidth]{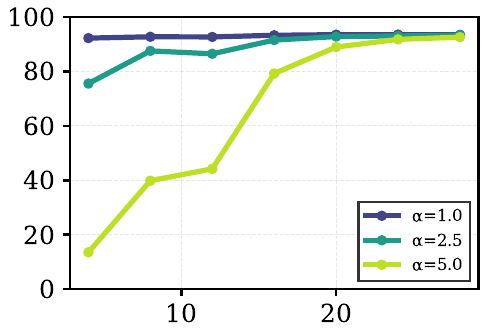} &
\includegraphics[width=0.16\textwidth]{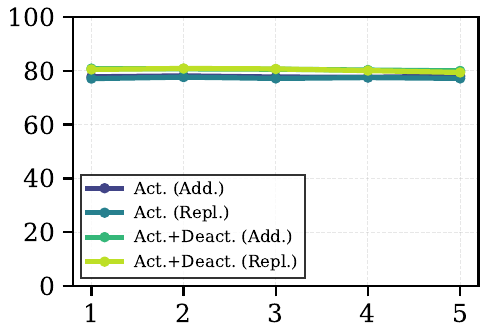} &
\includegraphics[width=0.16\textwidth]{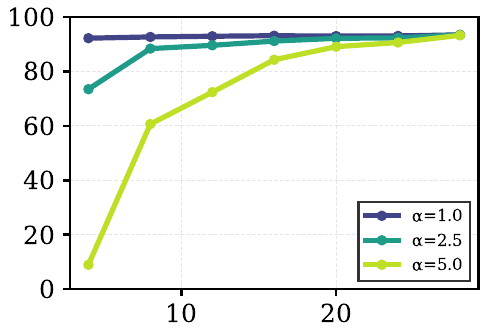} &
\includegraphics[width=0.16\textwidth]{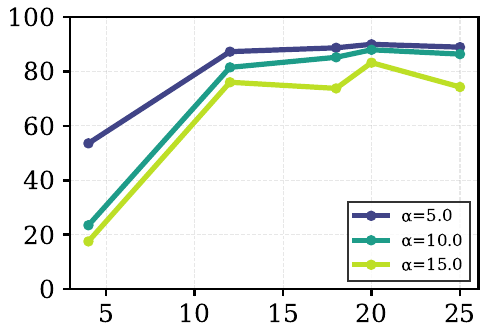} \\[-1pt]
\raisebox{0.3cm}{\rotatebox{90}{\scriptsize \textbf{Steering Score}}} &
\includegraphics[width=0.16\textwidth]{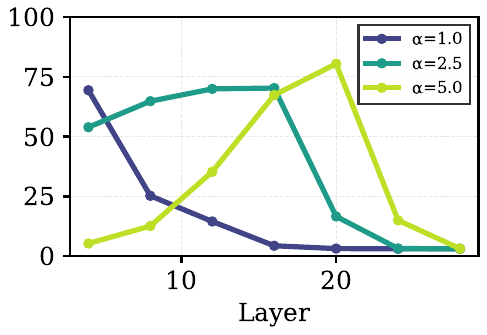} &
\includegraphics[width=0.16\textwidth]{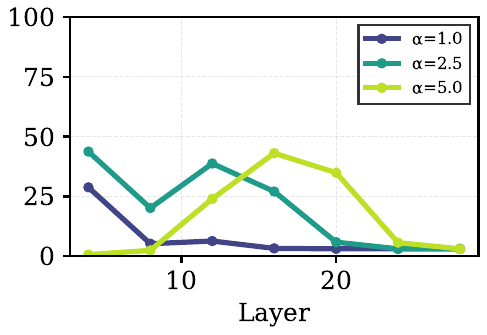} &
\includegraphics[width=0.16\textwidth]{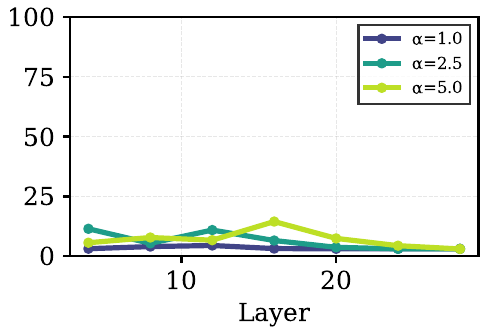} &
\includegraphics[width=0.16\textwidth]{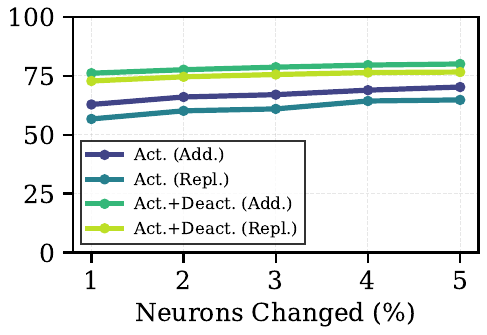} &
\includegraphics[width=0.16\textwidth]{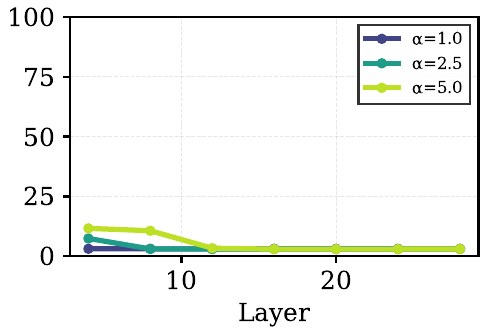} &
\includegraphics[width=0.16\textwidth]{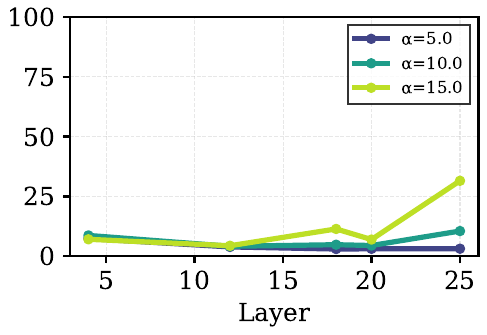}
\end{tabular}
}
\caption{Analysis of steering methods across evaluation metrics for \texttt{Llama-3.1-8B-Instruct}. Columns show different methods. Rows represent: forcing success rate, judge relevance, and overall steering score.}
\label{fig:method_comparison}
\end{figure*}

\section{Steering Results}
\label{sec:results}

We present the results of our cross-lingual steering evaluation in Table~\ref{tab:methods_languages}, reporting the steering score across 32 typologically diverse languages for \texttt{Llama-3.1-8B-Instruct} (results and discussion for \texttt{Aya-Expanse-8B} are in Appendix \ref{app:scores}). The results for each method are reported for the best steering configurations, as identified in Section~\ref{sec:ablation} and specified in Appendix~\ref{app:hyperparams}.

\paragraph{Residual-based steering outperforms prompting.}
DiffMean $\color{orange}\vec{\Delta}$ steering on residual activations achieves the highest average steering score (84.5\%), substantially outperforming both prompting baselines $\color{blue}\mathcal{E}$ (67.7\%) and $\color{green}\mathcal{T}$ (67.3\%). DiffMean maintains scores exceeding 90\% for 19 out of 32 languages, indicating robust cross-lingual generalization. LAPE $\color{red}\odot$ achieves the second-highest average (80.1\%), demonstrating that language-specific neurons can effectively control output language, though with greater variability across languages. In contrast, the prompting baselines reveal interesting inconsistencies: $\color{blue}\mathcal{E}$, for example, fails for English, Tibetan, and Farsi, while $\color{green}\mathcal{T}$ mostly struggles with English as a target. These failures highlight a potential limitation--models ignore or misinterpret explicit language directives, particularly for lower-resource languages. Table~\ref{tab:example} illustrates this concretely: given a Russian prompt with an explicit ``Respond in English'' instruction, the prompt-based method produces Russian output, while DiffMean successfully generates English. 

\paragraph{SAE-based steering lags behind.}
DiffMean steering in sparse autoencoder latent space ($\color{brown}\vec{\Delta}$SAE-DM.) achieves moderate performance (42.3\% average), outperforming several residual-stream methods but falling short of residual DiffMean steering. The method performs well for high-resource languages like German, Spanish, and Hindi, but struggles with Japanese and Slovak. This gap may stem from reconstruction error inherent in SAE encoding or from SAEs' training data not covering all languages extensively. 

\paragraph{Supervised methods underperform unsupervised approaches.}
Surprisingly, supervised methods such as probe-based $\color{purple}\mathbf{w}$ steering (48.6\%) and LDA $\color{blue}\mathbf{v}$ (23.6\%) underperform unsupervised DiffMean and LAPE approaches. Probe-based steering exhibits extreme variance, achieving 95.7\% for English but only 9.0\% for Greek and 14.5\% for Arabic. PCA-based $\color{cyan}\mathbf{u}$ methods perform the worst, averaging only 15.1\%. These results suggest that supervised objectives may overfit to language-specific characteristics in training data, while the unsupervised difference-of-means captures more generalizable language directions.

\begin{table}[t]
\centering
\scriptsize
\begin{tabular}{p{3.5cm}p{3.5cm}}
\toprule
\textbf{Language Instruction $\color{blue}\mathcal{E}$} & \textbf{DiffMean Steering} $\color{orange}\vec{\Delta}$ \\
\midrule
\multicolumn{2}{p{7.2cm}}{\textbf{Input:} \foreignlanguage{russian}{\textit{Как я могу развить свои навыки критического мышления?}}} \\
\midrule
\foreignlanguage{russian}{Развитие навыков критического мышления требует времени, усилий и практики. Вот несколько советов...} & Developing your critical thinking skills is a valuable investment in your personal and professional growth. Here are some tips... \\
\midrule
\cellcolor{red!20}\foreignlanguage{russian}{Russian} & \cellcolor{green!20}English \\
\bottomrule
\end{tabular}
\caption{Prompt-based instruction (left) outputs Russian despite explicit ``Respond in English'' directive. Activation-based steering (right) produces English without prompt modification. This holds for many source and target languages as indicated in Table \ref{tab:methods_languages}. More examples are provided in Appendix \ref{app:examples}.}
\label{tab:example}
\end{table}

\section{Ablation Analysis}
\label{sec:ablation}
We conduct systematic ablations to understand how steering effectiveness varies with intervention layer, steering strength, and method-specific parameters. Figure~\ref{fig:method_comparison} presents results for \texttt{Llama-3.1-8B-Instruct} across three evaluation dimensions: language forcing success, output relevance, and overall steering score. The results for \texttt{Aya-Expanse-8B} are in Appendix \ref{app:ablations}.

\paragraph{Layer and steering strength interact.}
A key finding is that optimal steering strength depends on intervention depth. For DiffMean, low strength ($\alpha=1.0$) suffices at early layers, achieving over 80\% steering score, while later layers require progressively higher strengths to be effective: $\alpha=5.0$ peaks around layer 20. Crucially, output quality remains stable across late-layer interventions even with high steering strengths, whereas early-layer interventions with strong $\alpha$ severely degrade coherence. This suggests that later layers encode language information in a more modular fashion, allowing targeted manipulation without disrupting other generation capabilities. 

Probe-based steering shows a similar but more pronounced pattern, with $\alpha=\{1.0, 2.5\}$ effective only at very early layers and higher strengths required beyond layer 10. LDA exhibits weak steering regardless of layer or strength, once exceeding 15\% success. For LAPE, the combined activation-plus-deactivation outperforms activation-only intervention strategy, and performance increases slightly when intervening on more neurons (from 1\% to 5\%). PCA shows modest steering with higher strengths ($\alpha=2.5$--$5.0$) in early layers but remains ineffective in other layers, likely due to those layers capturing less variance with the selected principal components. SAE-based steering, operating at higher strengths ($\alpha \in \{5.0, 10.0, 15.0\}$), shows a distinctive pattern: steering performance is best at layer 25 with alpha $\alpha=15.0$, indicating better language control in the sparse activation space of higher layers.

\section{Interpretability Insights}
\label{sec:mech_interp}

Beyond evaluating steering performance, \clasbench motivates investigation into how multilingual representations are organized within LLMs. We analyze the structural properties of language-specific components discovered through various methods, revealing consistent patterns. 

\begin{figure}[t]
\centering

\begin{subfigure}[b]{0.45\linewidth}
    \centering
    \includegraphics[width=\linewidth]{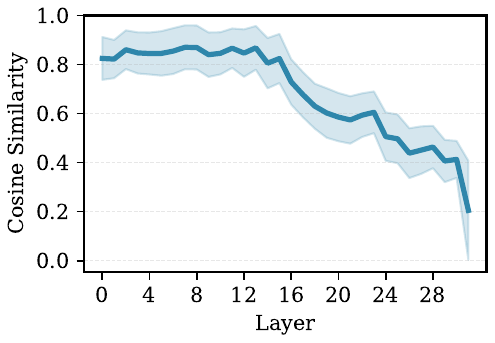}
    \caption{Lang Vectors}
\end{subfigure}
\begin{subfigure}[b]{0.45\linewidth}
    \centering
    \includegraphics[width=\linewidth]{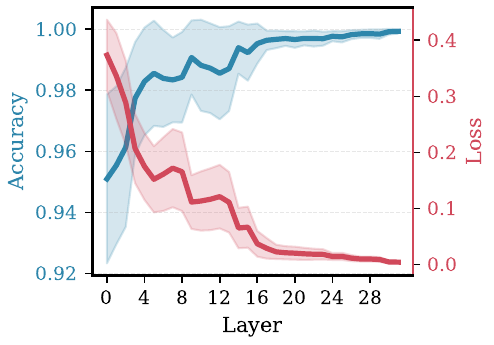}
    \caption{Probes}
\end{subfigure}


\begin{subfigure}[b]{0.45\linewidth}
    \centering
    \includegraphics[width=\linewidth]{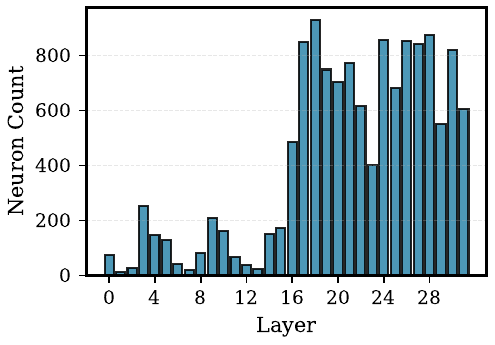}
    \caption{MLP Neurons}
\end{subfigure}
\begin{subfigure}[b]{0.45\linewidth}
    \centering
    \includegraphics[width=\linewidth]{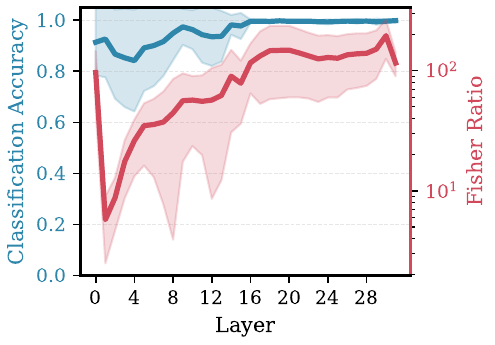}
    \caption{LDA}
\end{subfigure}

\caption{Insights into language-specific components across interpretation tools for \texttt{Llama-3.1-8B-Instruct}. (a) reveals average cosine similarity patterns across all language vectors. (b) demonstrates probe learning dynamics through loss and accuracy trajectories. (c) identifies the distribution of language-specific neurons across layers. (d) provides LDA classification accuracy and Fisher Ratio (the degree of separability between classes). 
}
\label{fig:mech_interp}
\end{figure}

\paragraph{Language-specific information concentrates in later layers.}
Converging evidence from multiple analysis methods indicates that language-specific representations emerge predominantly in layers 16--32. Figure~\ref{fig:mech_interp}a shows that cosine similarity between residual-based language vectors across language pairs decreases monotonically through the network, reaching minimum values (maximum separability) in layers 22--32. Linear probes (Figure~\ref{fig:mech_interp}b) achieve >99\% language classification accuracy from layer 14 onward, with the probe loss also reaching its minimum in the deeper layers. LAPE-identified language-specific neurons (Figure~\ref{fig:mech_interp}c) cluster predominantly in layers 24--28, with counts increasing sharply from layer 16. Finally, LDA classification accuracy and Fisher ratio \cite{fisher1936use} (Figure~\ref{fig:mech_interp}d) both peak in late layers. This convergence suggests a hierarchical processing view: early layers encode language-agnostic features, while later layers encode language-specific generation patterns. 

\paragraph{Language families exhibit geometric clustering.}
Analysis of steering vector similarities (Appendix Figures~\ref{fig:residual_overlap_llama}--\ref{fig:lda_overlap_aya}) reveals that typologically related languages cluster in representation space. Romance languages (Spanish, French, Portuguese, Italian, Romanian) exhibit high mutual cosine similarity, as do Germanic (German, Dutch, Swedish, Danish, Norwegian) and Slavic languages (Russian, Polish, Ukrainian, Czech, Slovak). Figure~\ref{fig:neuron_overlap_llama} further shows that LAPE-identified neurons are largely language-specific, with substantial overlap only within language families. This geometric structure has practical implications: steering between related languages might require smaller interventions, while cross-family steering (e.g., Japanese to Arabic) could demand larger modifications and be more susceptible to quality degradation.

\paragraph{Implications for multilingual interpretability.}
Our findings support the view that language control operates through geometrically structured, linearly accessible representations. The success of DiffMean over statistically sophisticated methods (LDA, PCA) suggests that language directions are well-approximated by simple difference vectors. This is encouraging for interpretability: language control does not require modeling complex nonlinear interactions, but rather identifying appropriate linear subspaces in relevant layers. \clasbench thus provides a foundation for systematic investigation of multilingual representations, enabling researchers to probe not only \emph{whether} steering works, but \emph{why} and \emph{where} it succeeds or fails.

%% file: Chapters/06_RelatedWork.tex
\section{Related Work}
\label{sec:relatedWork}

\subsection{Representation-based Steering} 
A broad line of work explores steering language models by directly manipulating their hidden representations, rather than relying solely on prompts or fine-tuning. Typical approaches include adding fixed vectors to activations, selectively activating neurons, or constraining intermediate states.

Several works have explored representation-based paradigm across different tasks. For instance, \citet{panickssery2023steering} and \citet{turner2023steering} demonstrate that simple vector-based interventions can steer models toward more truthful or less sycophantic behaviors. \citet{marks2023geometry} formalize geometric methods such as DiffMean, enabling systematic manipulation of residual-stream activations. Sparse latent-space methods \cite{cunningham2023sparse} use interpretable autoencoder directions for controllability. \citet{liu2023context} combine representation interventions with in-context learning to guide semantic properties. Collectively, these studies show that internal representations are a viable interface for direct behavioral control.

\subsection{Language-specific Dimensions} 
Recent work has specifically focused on the identification and steerability of language-specific components in LLMs, with two main lines of research: neuron-based and SAE-based methods.

\paragraph{Neuron-based methods.} Neuron-based approaches focus on detecting neurons sensitive to particular languages and testing their causal role through interventions. For example, \citet{zhao2024largelanguagemodelshandle} identify language-sensitive neurons in both attention and MLP layers via ablation and hidden state perturbation analysis, showing that setting these neurons to zero suppresses the corresponding language. \citet{kojima2024multilingual} instead train binary classifiers on neuron activations to rank neurons by their discriminative power, and propose replacement-based manipulation to steer model outputs. Similarly, \citet{tang2024language} introduce the Language Activation Probability Entropy (LAPE) method, demonstrating language steering by deactivating source neurons (zeroing) and activating target neurons (setting to an average). 


\paragraph{SAE-based methods.} SAEs have emerged as a powerful tool for uncovering interpretable, language-specific features and manipulating model behavior through interventions in sparse latent spaces. \citet{deng2025unveilinglanguagespecificfeatureslarge} introduce a metric for monolinguality of SAE features and demonstrate that ablating language-specific features selectively impairs performance in one language and that these features can enhance the construction of steering vectors to control language generation. \citet{chou2025causal} propose leveraging SAE features to achieve causal language control in LLMs; by modulating the activation of a single SAE feature in mid-to-late transformer layers, they steer generation toward target languages with up to 90\% success, while preserving semantic fidelity. 

\subsection{Benchmarking Steering}

Several benchmarks have been proposed to evaluate steering and interpretable control in LLMs. AxBench \cite{wu2025axbench} provides a large-scale benchmark for steering and concept detection in English, comparing prompting, fine-tuning, and representation-based methods (e.g., SAEs, DiffMean, ReFT-r1), and finds that prompting generally outperforms existing approaches. MIB \cite{mueller2025mib} assesses mechanistic interpretability through circuit and causal variable localization tasks, showing that attribution and supervised distributed alignment search (DAS) \cite{geiger2024finding} methods outperform SAE-based features for recovering causal components. Steer-Bench \cite{chen2025steer} evaluates population-specific steerability with in-context learning and fine-tuning by testing whether LLMs can adapt outputs to align with the norms, perspectives, and communication styles of 30 contrasting subreddit pairs. 

While these benchmarks advance evaluation of English-language steering and interpretability, \textbf{neither assesses steering in multilingual or cross-lingual settings}. This leaves unanswered how well steering generalizes across languages, how methods perform on low-resource or typologically distant languages, and how multilingual representations can be systematically probed. \clasbench fills this gap in the literature.

%% file: Chapters/08_Conclusion.tex
\section{Conclusion}
\label{sec:conclusion}

We introduce \clasbench, the first benchmark for standardized evaluation of cross-lingual language steering in LLMs.\footnote{The code and data are publicly available at \\ \url{https://github.com/d-gurgurov/CLaS-Bench}.} Covering 32 diverse languages with 70 parallel high-quality open-ended questions each, \clasbench establishes a structured framework for measuring the effectiveness of steering methods in controlling output language. Unlike prior work that primarily focuses on English and conceptual attributes, our benchmark positions multilingualism at the center, highlighting both the strengths and limitations of existing approaches.

Our evaluation setup enables cross-lingual experiments, revealing whether steering works consistently across languages and how it compares to prompting. Designed to be lightweight and easily extendable, the benchmark allows new languages to be incorporated simply by translating the questions and applying the same evaluation protocol. While our current focus is on 32 languages, \clasbench can naturally grow into a broader multilingual resource. By providing a common ground for comparing steering methods, we aim to accelerate research at the intersection of interpretability and multilingual NLP, ultimately advancing our understanding of how LLMs represent language and supporting the development of user-adaptive multilingual systems that operate reliably across diverse linguistic contexts.

%% file: Appendices/A.tex
\clearpage
\section*{Appendix}
\section{Data Curation and Native Speaker Validation}
\label{app:data_curation}

\subsection{Translation and Proofreading Protocol}

The initial translation of the 70 English prompts into 34 additional languages was performed using the Google Translate API to ensure consistency and comprehensive coverage. To guarantee semantic fidelity, fluency, and idiomaticity, all translations underwent a systematic proofreading process conducted by native speakers of the target languages.

We recruited volunteer native speakers from our institution's campus community, representing all 34 target languages. Participants were provided with access to a dedicated web interface displaying the English source prompts alongside their machine-translated versions. The interface allowed annotators to review, correct, and refine translations while maintaining semantic equivalence with the original English questions. Proofreaders were instructed to prioritize:
\begin{itemize}
    \item \textbf{Semantic fidelity:} Ensuring the translated prompts retained the intended meaning and conversational intent of the English source
    \item \textbf{Fluency and idiomaticity:} Correcting grammatical errors and replacing awkward phrasing with natural, idiomatic expressions appropriate for native speakers
    \item \textbf{Domain consistency:} Maintaining the conversational tone and style across all linguistic domains (reasoning, knowledge, personal opinions, creative, and professional writing)
\end{itemize}

Each native speaker volunteer spent less than one hour completing the proofreading task for their respective language(s). No compensation was offered, and participation was entirely voluntary.

\subsection{Annotator Background}

All proofreaders were native speakers of their respective target languages with fluency in English, enabling them to accurately assess translation quality. The majority of participants had backgrounds in linguistics, computer science, or language technology. Annotator information was collected and stored anonymously to protect participant privacy.

\subsection{Ethical Considerations}

Prior to participation, all volunteers were informed about the purpose of the data curation task and provided explicit consent for the corrected translations to be used in subsequent research and made available for public release (with appropriate anonymization of annotator identities). No ethics review board approval was sought, as the proofreading task did not fall under institutional requirements for formal ethical review. The study involved minimal risk to participants, consisted of standard proofreading activities, and did not require collection of sensitive personal information beyond basic language background.

\section{Judge Prompt}
\label{app:judge}
\begin{tcolorbox}[
  colback=white,
  colframe=black!50,
  arc=1.5mm,
  boxrule=0.8pt,
  left=8pt,
  right=8pt,
  top=8pt,
  bottom=8pt
]
\small
\noindent\textbf{Judge Prompt.} You are an impartial judge evaluating multilingual responses from a language steering experiment. A question was asked in \{\textit{source\_language}\}, and the model was steered to respond in \{\textit{target\_language}\}. 

\vspace{4pt}

\noindent\textbf{\textcolor{black!70}{Important:}} Do NOT evaluate which language the response is in or whether steering succeeded. Ignore the target language entirely for scoring---evaluate the final response solely on its content, regardless of the language used.

\vspace{6pt}

\noindent{\color{black!80}\rule{\linewidth}{0.4pt}}

\vspace{4pt}

\noindent\textbf{Evaluation Criteria:}
\vspace{3pt}

\noindent\hspace{0.5cm}(1) \textbf{Relevance:} Is the response a relevant answer to the question?

\noindent\hspace{0.5cm}(2) \textbf{Coherence:} Is the response coherent text, or is it gibberish, repetitive, or useless?

\vspace{6pt}

\noindent\textbf{Scoring Rubric:}
\vspace{3pt}

\noindent\hspace{0.5cm}$\bullet$ \textbf{0:} Completely unrelated or gibberish/useless text.

\vspace{2pt}

\noindent\hspace{0.5cm}$\bullet$ \textbf{1:} Somewhat related but incomplete, partially off-target, or minimally useful.

\vspace{2pt}

\noindent\hspace{0.5cm}$\bullet$ \textbf{2:} Clearly related to the question, coherent, and useful text.

\vspace{6pt}

\noindent{\color{black!80}\rule{\linewidth}{0.4pt}}

\vspace{4pt}

\noindent\textit{Output format:} \texttt{Rating: [[score]]}
\end{tcolorbox}

\clearpage

\begin{table*}[t]
\section{Selected Languages}
\label{app:langs}
\centering
\footnotesize
\begin{tabular}{lllll}
\toprule
\textbf{Language} & \textbf{ISO} & \textbf{Glottolog Family} & \textbf{Script} & \textbf{Resource Level} \\
\midrule
Tibetan & bo & Sino-Tibetan (Bodic) & Tibetan & 1 \\
Maltese & mt & Afro-Asiatic (Semitic) & Latin & 2 \\
Italian & it & Indo-European (Romance) & Latin & 4 \\
Spanish & es & Indo-European (Romance) & Latin & 5 \\
German & de & Indo-European (Germanic) & Latin & 5 \\
Japanese & ja & Japonic & Japanese & 5 \\
Arabic & ar & Afro-Asiatic (Semitic) & Arabic & 5 \\
Chinese & zh & Sino-Tibetan (Chinese) & Han & 5 \\
Dutch & nl & Indo-European (Germanic) & Latin & 4 \\
French & fr & Indo-European (Romance) & Latin & 5 \\
Portuguese & pt & Indo-European (Romance) & Latin & 4 \\
Russian & ru & Indo-European (Slavic) & Cyrillic & 4 \\
Korean & ko & Koreanic & Hangul & 4 \\
Hindi & hi & Indo-European (Indo-Aryan) & Devanagari & 4 \\
Turkish & tr & Turkic & Latin & 4 \\
Polish & pl & Indo-European (Slavic) & Latin & 4 \\
Swedish & sv & Indo-European (Germanic) & Latin & 4 \\
Danish & da & Indo-European (Germanic) & Latin & 3 \\
Norwegian & no & Indo-European (Germanic) & Latin & 1 \\
English & en & Indo-European (Germanic) & Latin & 5 \\
Slovak & sk & Indo-European (Slavic) & Latin & 3 \\
Greek & el & Indo-European (Hellenic) & Greek & 3 \\
Swahili & sw & Atlantic-Congo (Bantu) & Latin & 2 \\
Kazakh & kk & Turkic & Cyrillic & 3 \\
Georgian & ka & Kartvelian & Georgian & 2 \\
Ukrainian & uk & Indo-European (Slavic) & Cyrillic & 3 \\
Persian & fa & Indo-European (Iranian) & Arabic-Persian & 4 \\
Thai & th & Kra-Dai & Thai & 3 \\
Indonesian & id & Austronesian & Latin & 3 \\
Vietnamese & vi & Austroasiatic & Latin & 4 \\
Czech & cs & Indo-European (Slavic) & Latin & 4 \\
Romanian & ro & Indo-European (Romance) & Latin & 3 \\
\bottomrule
\end{tabular}
\caption{Languages included in CLaS-Bench, with ISO codes, Glottolog family assignments \cite{glottolog}, writing systems, and resource levels \cite{joshi-etal-2020-state}.}
\label{tab:languages}
\end{table*}
\clearpage

\begin{figure*}[ht]
\section{Language Neurons from MLPs}
    \centering
    \includegraphics[width=0.8\linewidth]{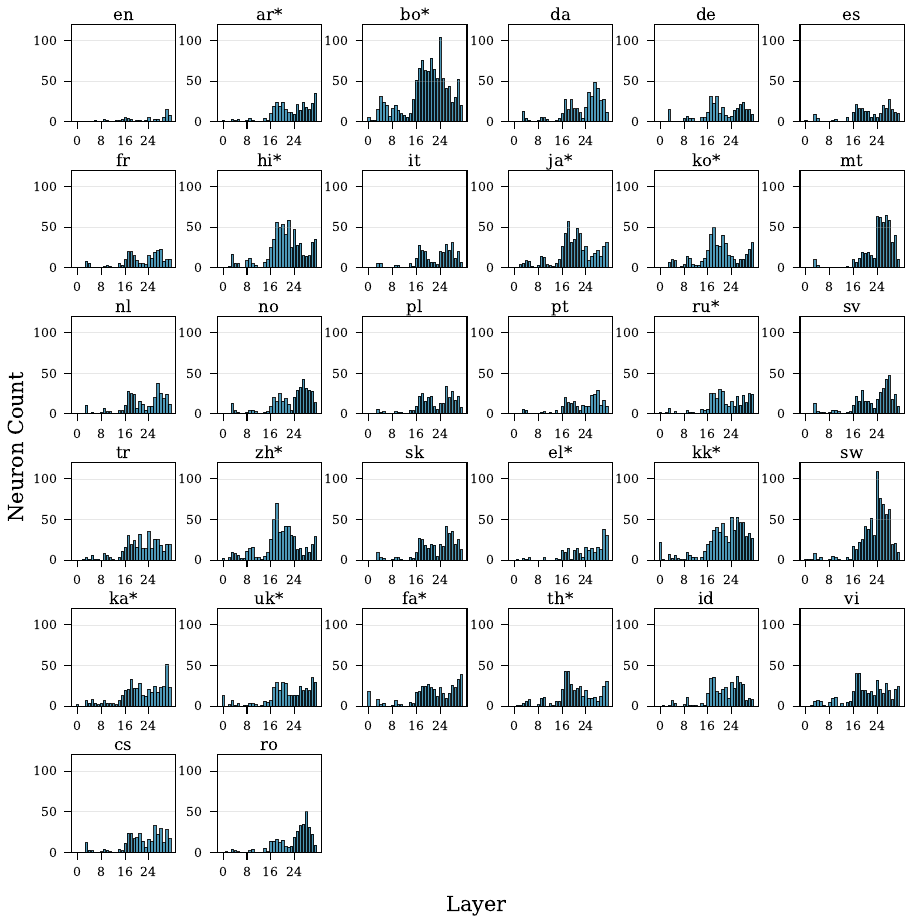}
    \caption{\textbf{Distribution} of LAPE identified \textbf{language-specific neurons} over layers in \texttt{Llama-3.1-Instruct} for all 32 languages.}
    \label{fig:neuron_dists_llama}
\end{figure*}

\begin{figure*}[ht]
    \centering
    \includegraphics[width=0.8\linewidth]{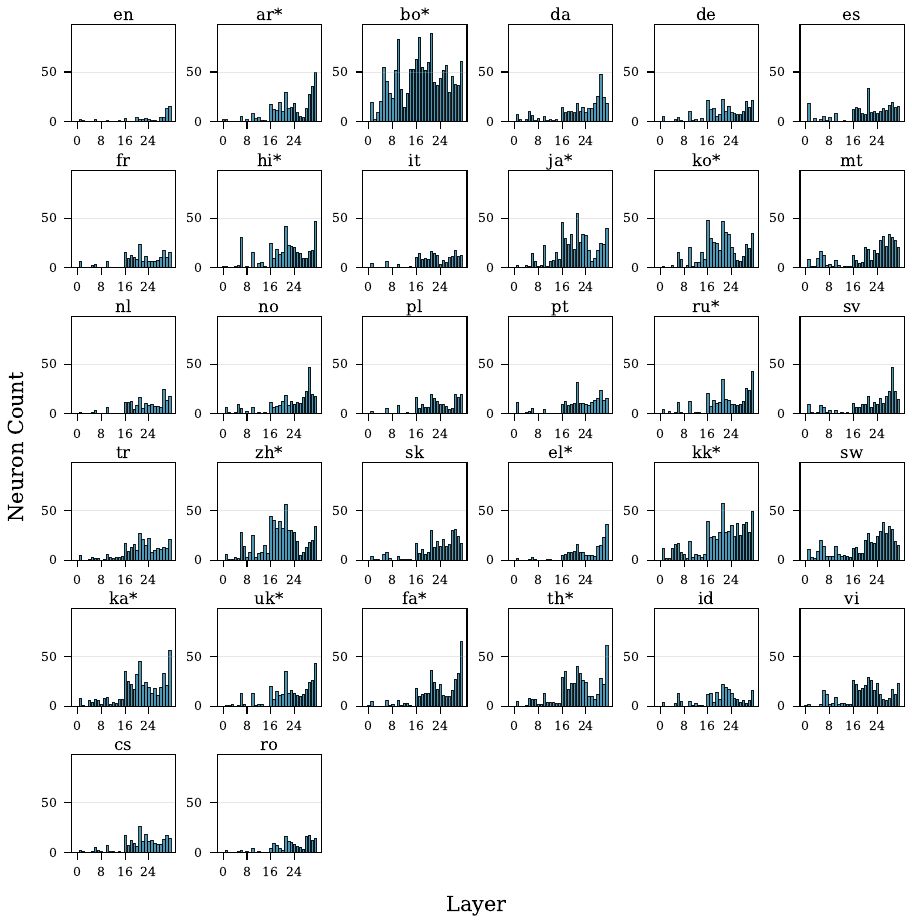}
    \caption{\textbf{Distribution} of LAPE identified \textbf{language-specific neurons} over layers in \texttt{Aya-Expanse-8B} for all 32 languages.}
    \label{fig:neuron_dists_aya}
\end{figure*}

\begin{figure*}[ht]
    \centering
    \includegraphics[width=0.8\linewidth]{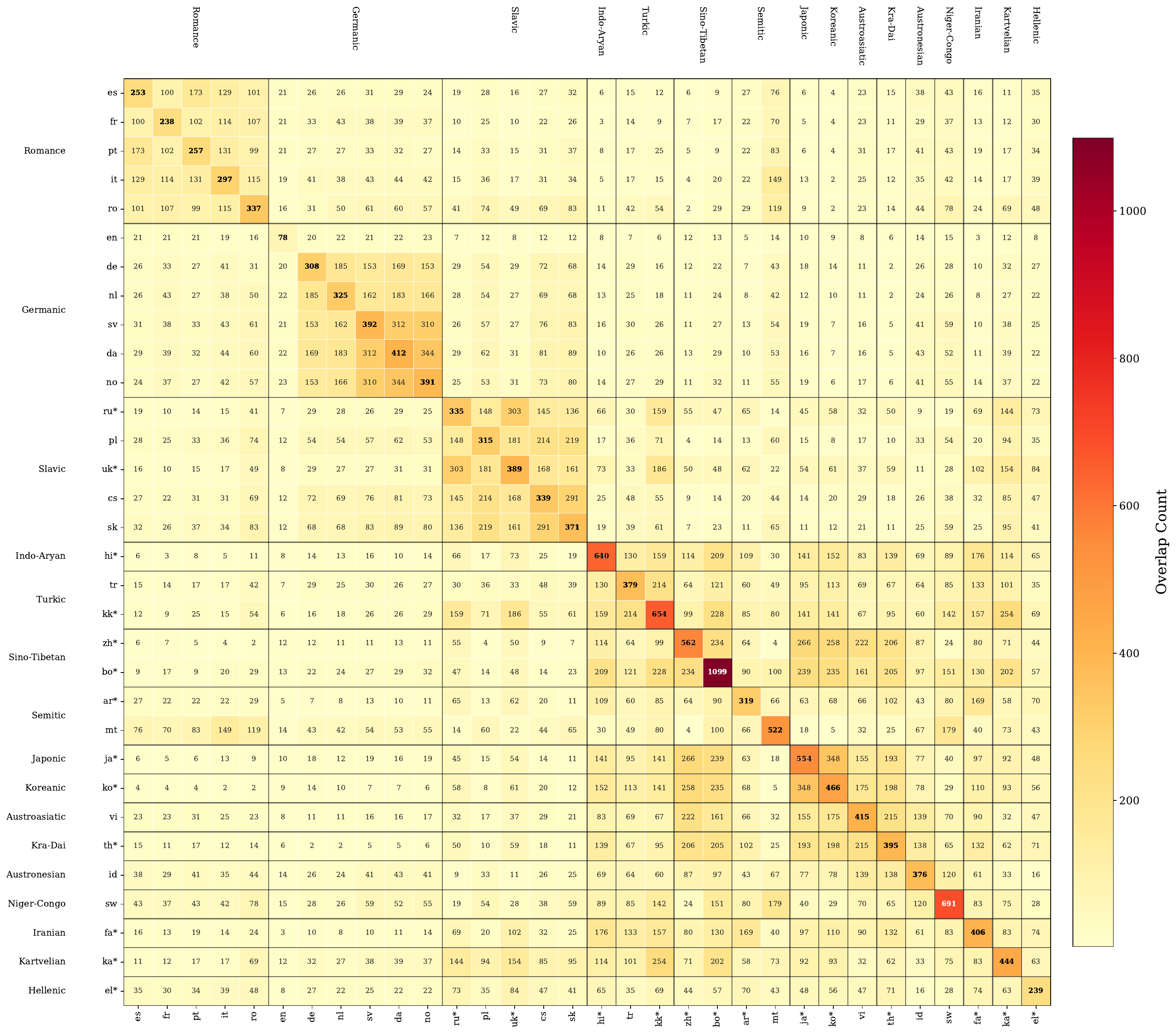}
    \caption{\textbf{Overlap} of all LAPE identified \textbf{language-specific neurons} in \texttt{Llama-3.1-Instruct} for the selected 32 languages.}
    \label{fig:neuron_overlap_llama}
\end{figure*}

\begin{figure*}[ht]
    \centering
    \includegraphics[width=0.8\linewidth]{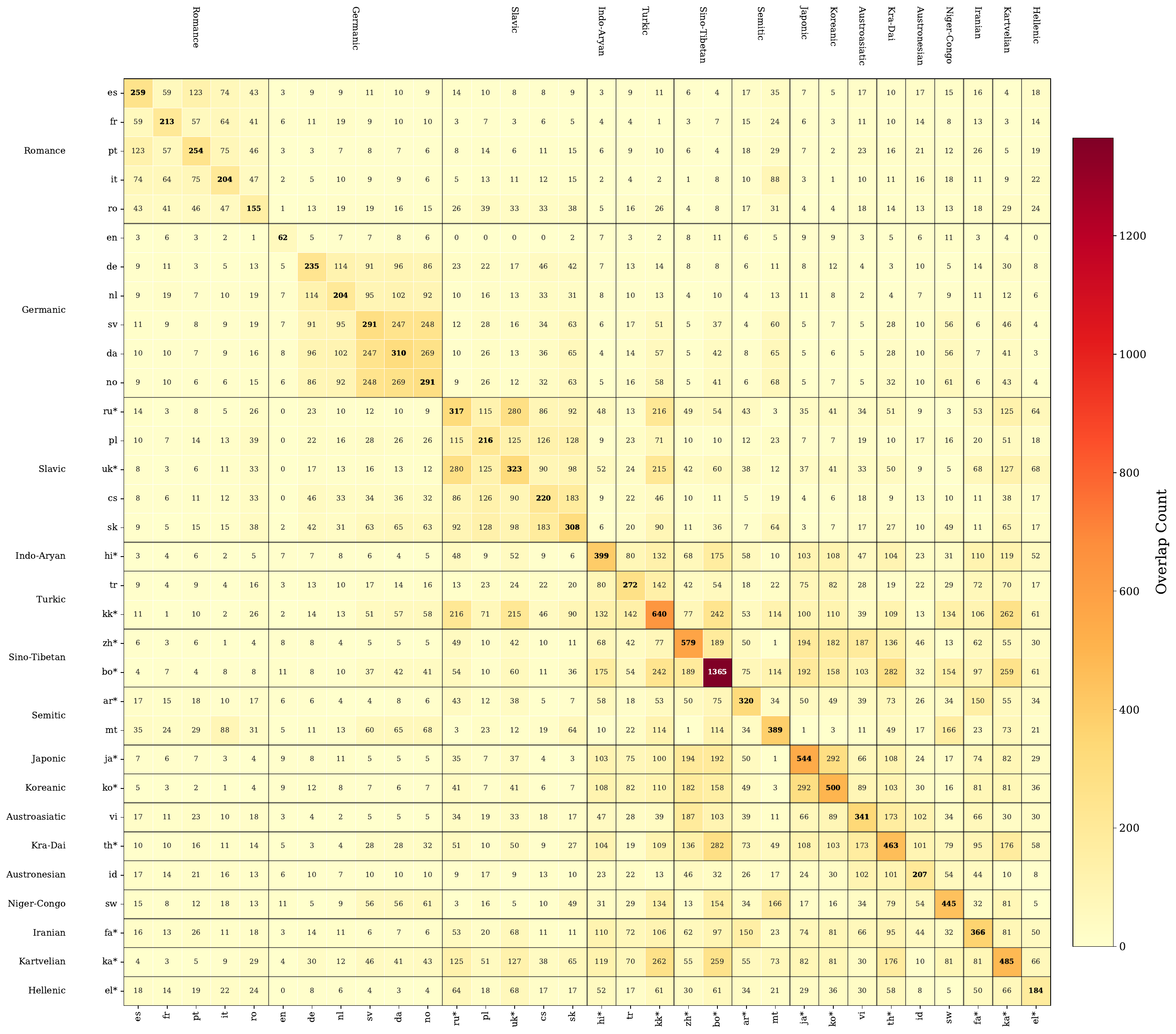}
    \caption{\textbf{Overlap} of all LAPE identified \textbf{language-specific neurons} in \texttt{Aya-Expanse-8B} for the selected 32 languages.}
    \label{fig:neuron_overlap_aya}
\end{figure*}

\begin{figure*}[ht]
\section{Language Vectors from Residuals}
    \centering
    \includegraphics[width=0.8\linewidth]{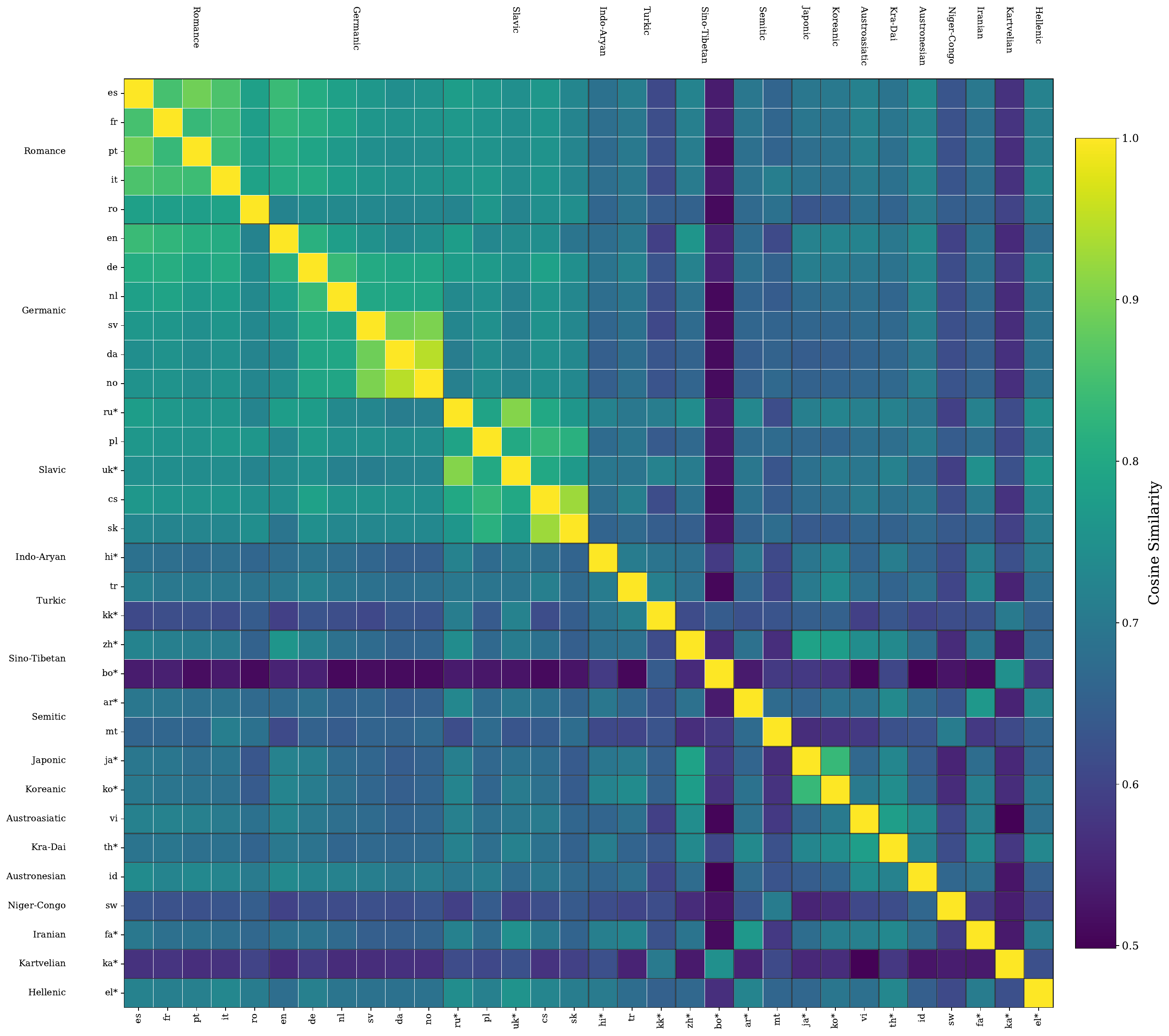}
    \caption{\textbf{Cosine similarity} between the \textbf{residual-based vectors} for all 32 selected languages in \texttt{Llama-3.1-Instruct} averaged over all layers.}
    \label{fig:residual_overlap_llama}
\end{figure*}

\begin{figure*}[ht]
    \centering
    \includegraphics[width=0.8\linewidth]{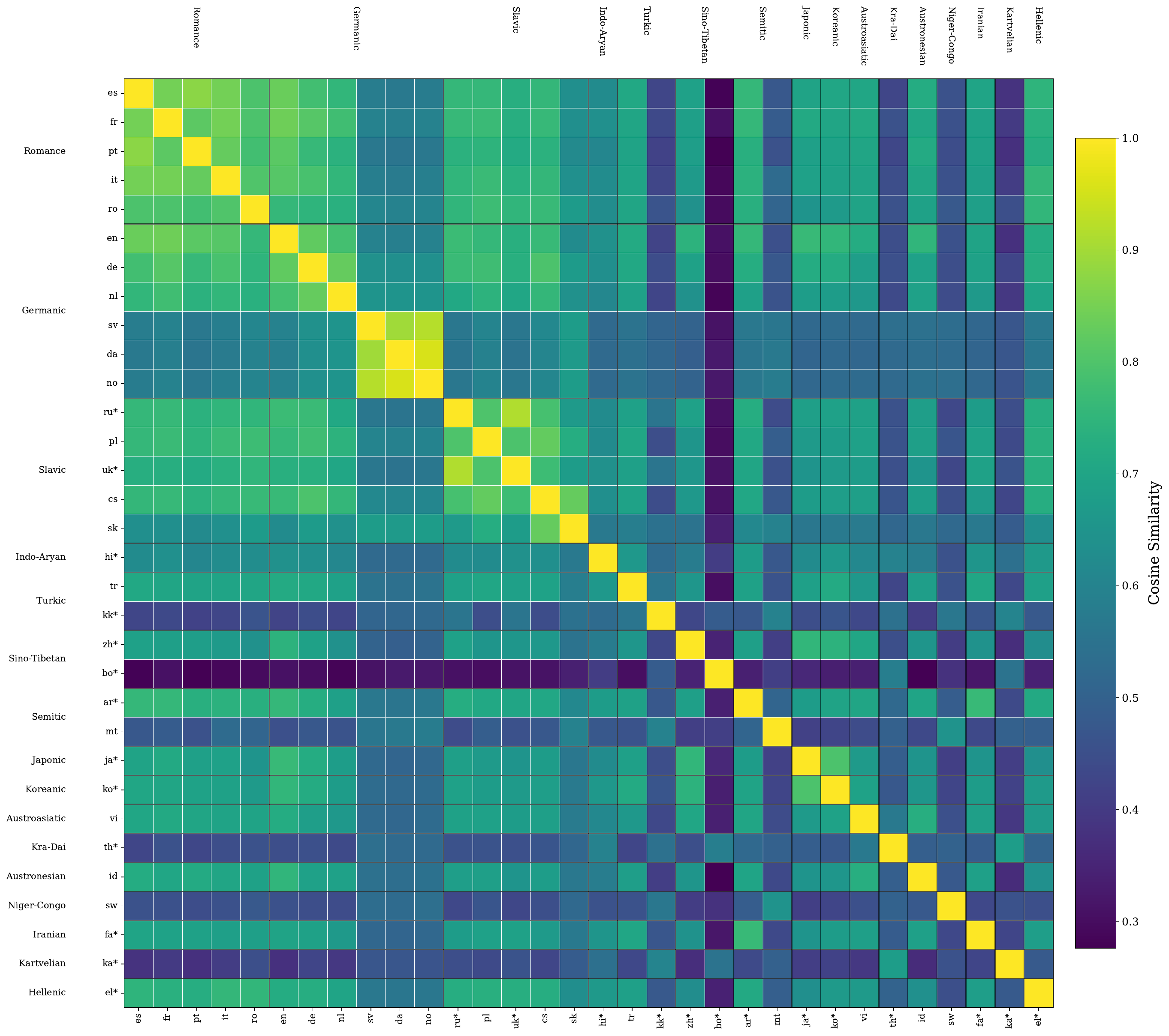}
    \caption{\textbf{Cosine similarity} between the \textbf{residual-based vectors} for all 32 selected languages in \texttt{Aya-Expanse-8B} averaged over all layers.}
    \label{fig:residual_overlap_aya}
\end{figure*}

\begin{figure*}[ht]
\section{Language Vectors from Probes}
    \centering
    \includegraphics[width=0.8\linewidth]{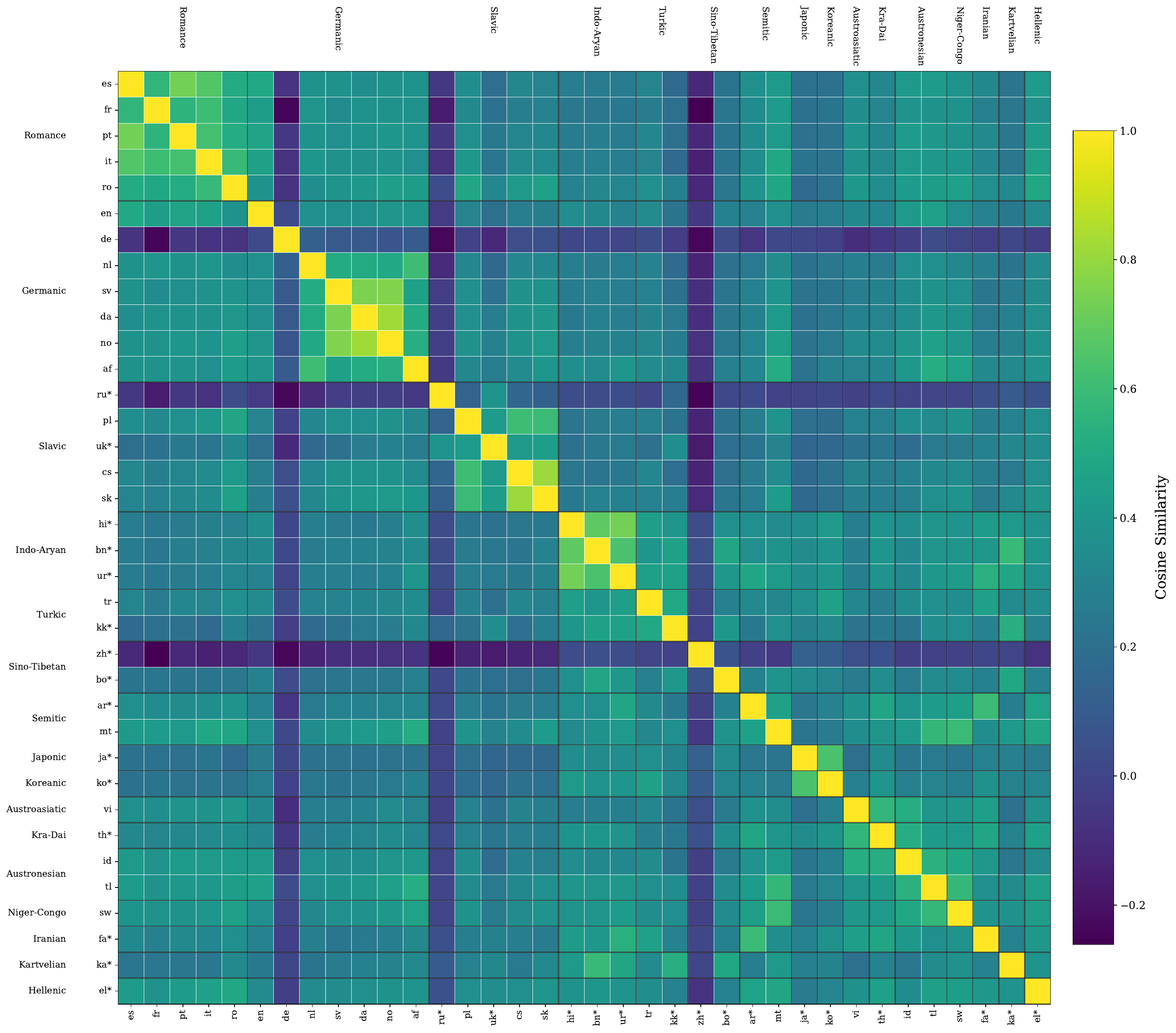}
    \caption{\textbf{Cosine similarity} between the \textbf{probe-based vectors} for all 32 selected languages in \texttt{Llama-3.1-Instruct} averaged over all layers.}
    \label{fig:probe_overlap_llama}
\end{figure*}

\begin{figure*}[ht]
    \centering
    \includegraphics[width=0.8\linewidth]{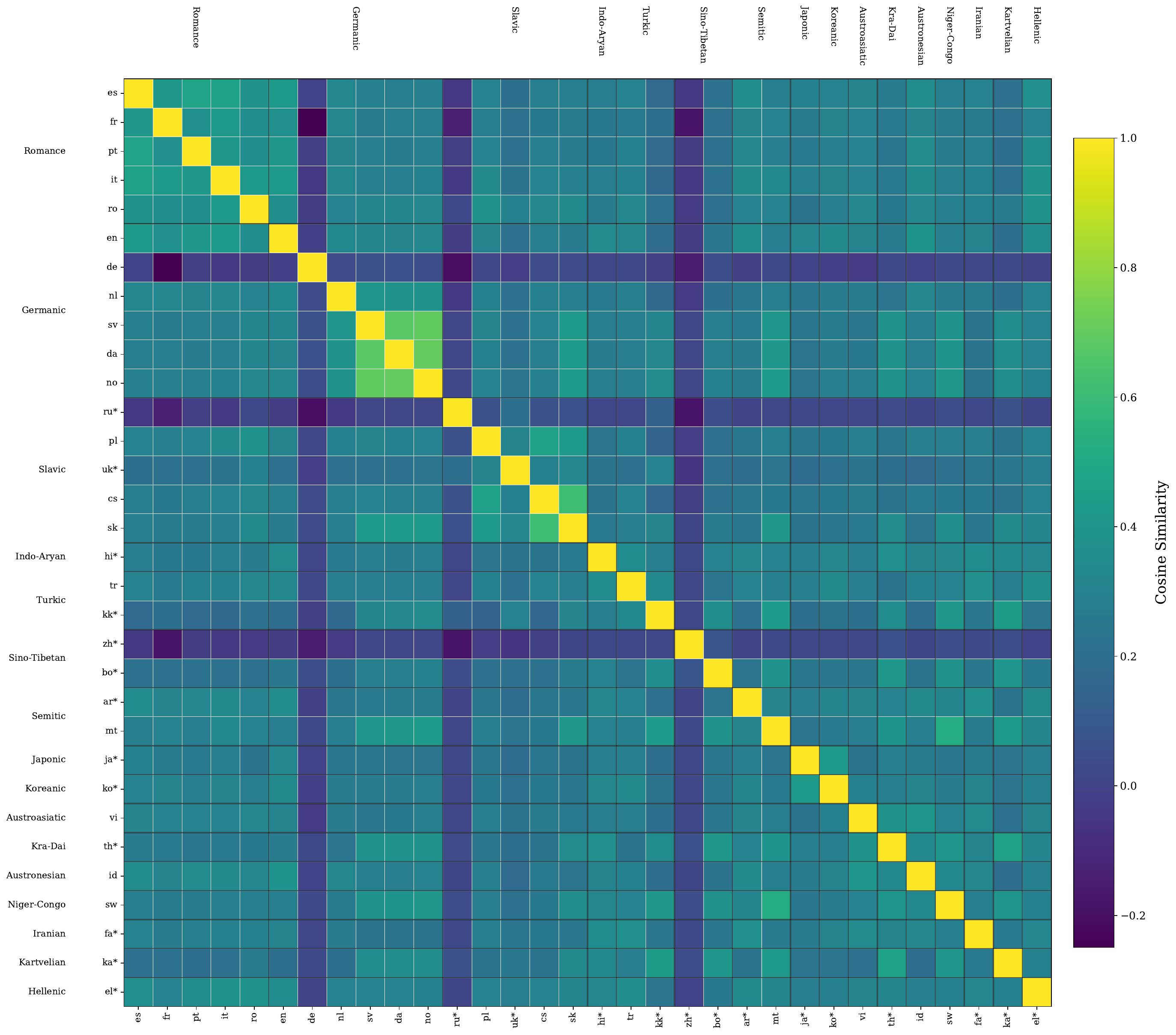}
    \caption{\textbf{Cosine similarity} between the \textbf{probe-based vectors} for all 32 selected languages in \texttt{Aya-Expanse-8B} averaged over all layers.}
    \label{fig:probe_overlap_aya}
\end{figure*}

\begin{figure*}[ht]
\section{Language Vectors from LDA}
    \centering
    \includegraphics[width=0.8\linewidth]{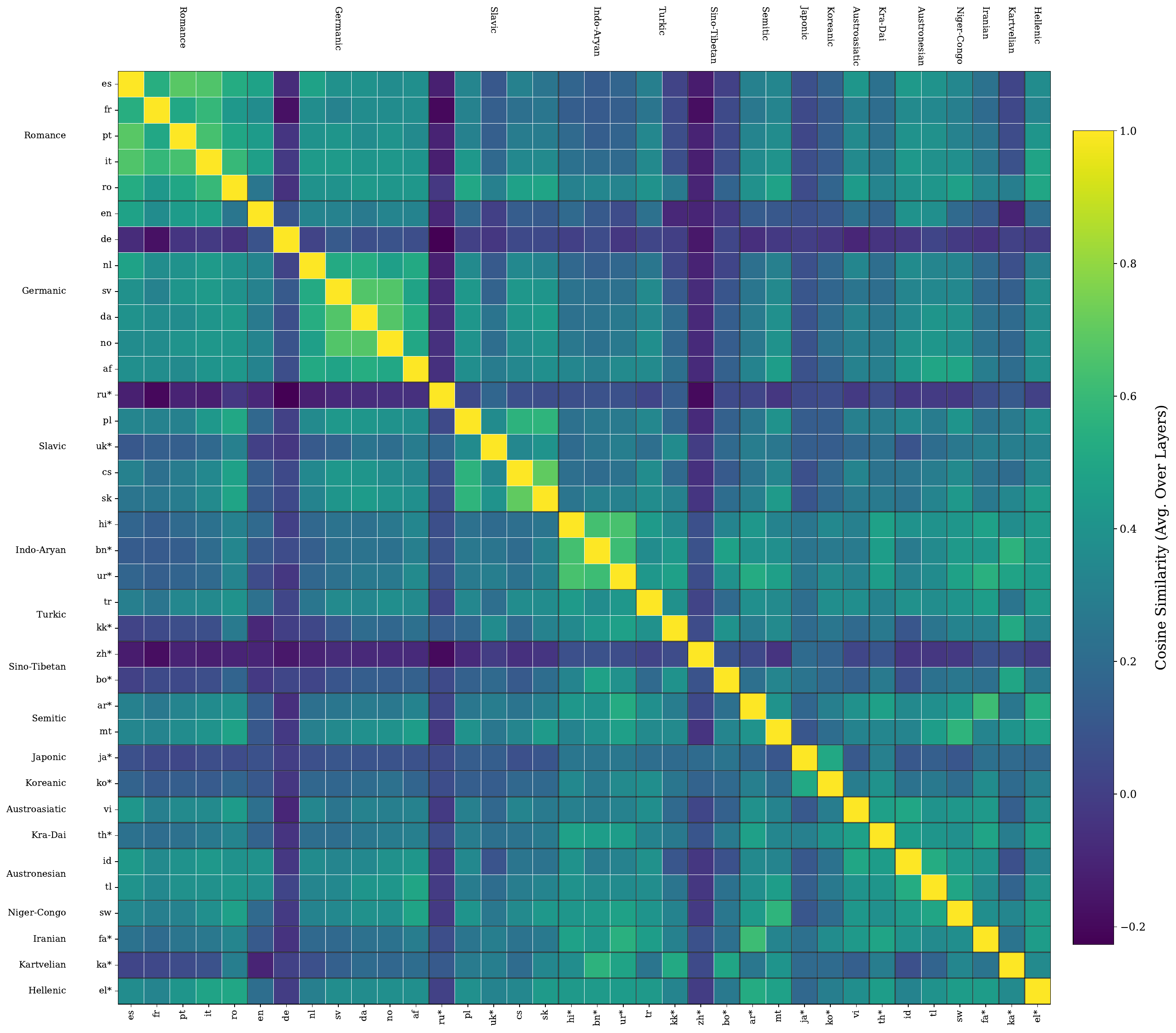}
    \caption{Cosine similarity between the LDA-based vectors for all 32 selected languages in \texttt{Llama-3.1-Instruct} averaged over all layers.}
    \label{fig:lda_overlap_llama}
\end{figure*}

\begin{figure*}[ht]
    \centering
    \includegraphics[width=0.8\linewidth]{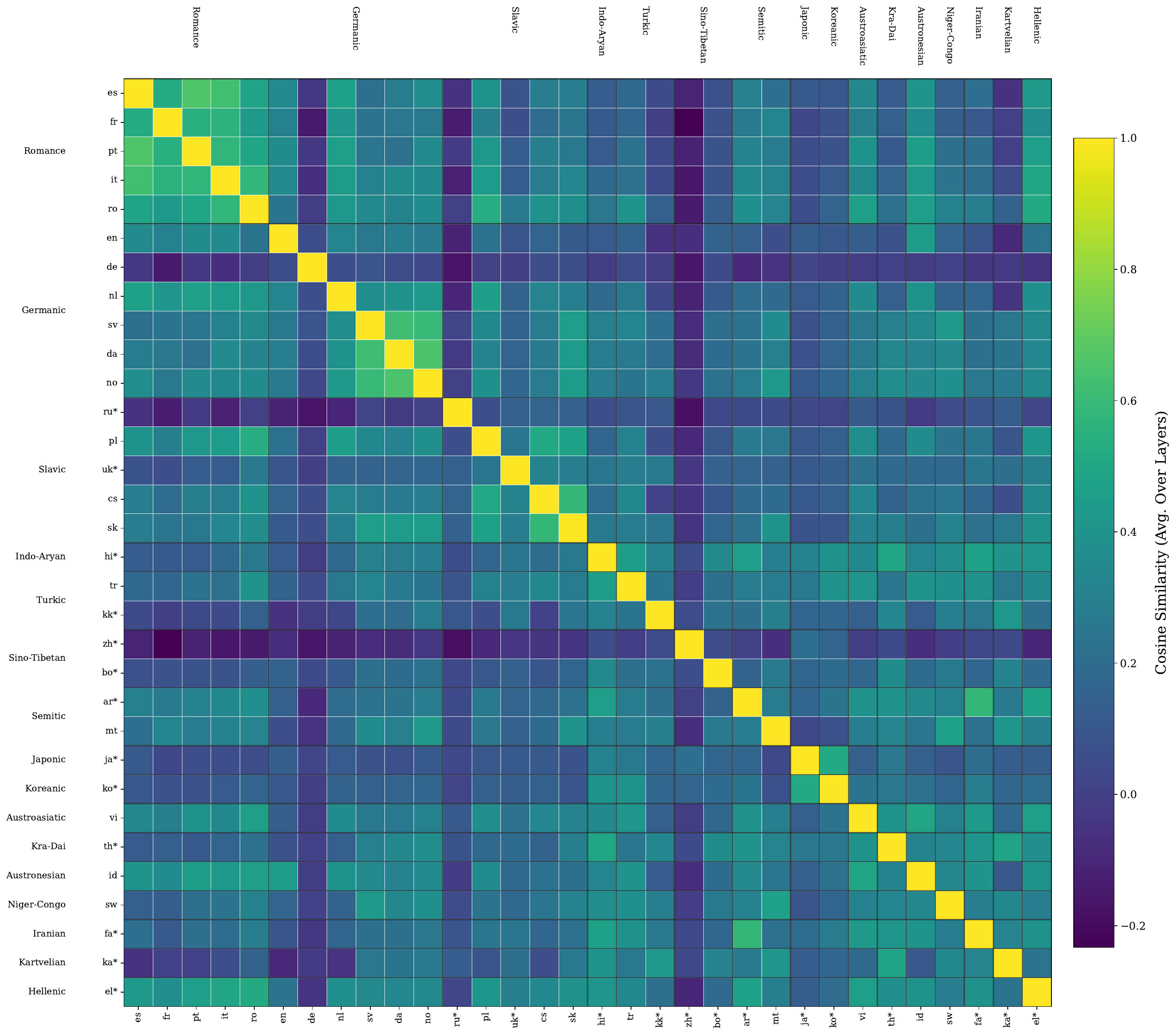}
    \caption{Cosine similarity between the LDA-based vectors for all 32 selected languages in \texttt{Aya-Expanse-8B} averaged over all layers.}
    \label{fig:lda_overlap_aya}
\end{figure*}
\clearpage

\begin{figure*}[t]
\section{Mechanistic Insights into Language-specific Components}
\centering

\begin{subfigure}[b]{0.38\linewidth}
    \centering
    \includegraphics[width=\linewidth]{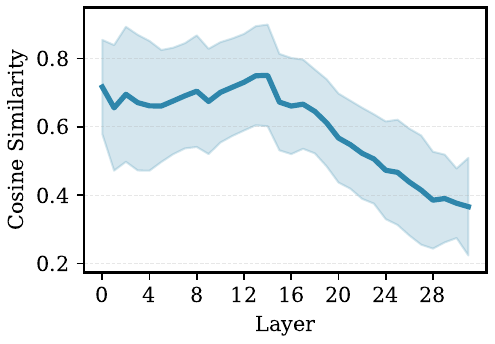}
    \caption{DiffMean}
\end{subfigure}
\begin{subfigure}[b]{0.38\linewidth}
    \centering
    \includegraphics[width=\linewidth]{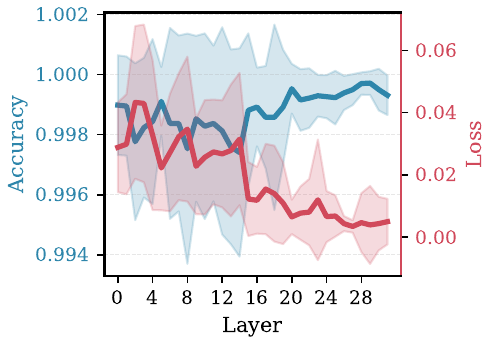}
    \caption{Probes}
\end{subfigure}

\vspace{1em}

\begin{subfigure}[b]{0.38\linewidth}
    \centering
    \includegraphics[width=\linewidth]{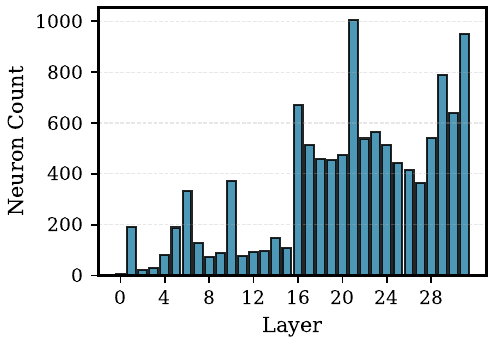}
    \caption{MLP Neurons}
\end{subfigure}
\begin{subfigure}[b]{0.38\linewidth}
    \centering
    \includegraphics[width=\linewidth]{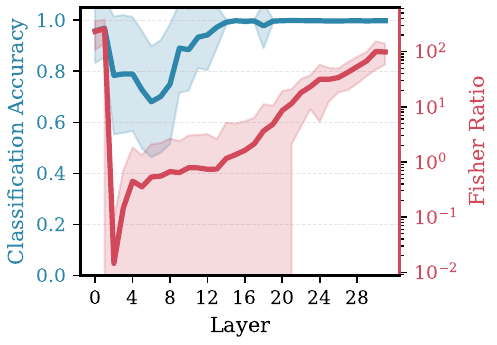}
    \caption{LDA}
\end{subfigure}

\caption{Mechanistic insights into language-specific components across interpretation tools for \texttt{Aya-Expanse-8B}. (a) DiffMean reveals average cosine similarity patterns across all languages. (b) Probes demonstrate learning dynamics through loss and accuracy trajectories. (c) LDA provides classification accuracy and Fisher Ratio (the degree of separability between two classes considered for LDA). (d) LAPE identifies the distribution of language-specific neurons across layers. Across all four methods, a consistent pattern emerges: language specificity concentrates in later layers, suggesting that higher-level representations encode language-dependent information.}
\label{fig:mech_interp_aya}
\end{figure*}
\clearpage

\begin{figure*}[t!]
\section{Ablation Results}
\label{app:ablations}
\centering
\setlength{\tabcolsep}{1pt}
\renewcommand{\arraystretch}{0.6} 
\resizebox{\textwidth}{!}{
\begin{tabular}{lccccc}
& {\scriptsize\(\color{orange}{\vec{\Delta}}\) \textbf{DiffMean}} 
& {\scriptsize\(\color{purple}{\mathbf{w}}\) \textbf{Probe}} 
& {\scriptsize\(\color{blue}{\mathbf{v}}\) \textbf{LDA}}
& {\scriptsize\(\color{red}{\odot}\) \textbf{LAPE}} 
& {\scriptsize\(\color{cyan}{\mathbf{u}}\) \textbf{PCA}} \\[0pt]
\raisebox{0.3cm}{\rotatebox{90}{\scriptsize \textbf{Lang. Forcing}}} &
\includegraphics[width=0.16\textwidth]{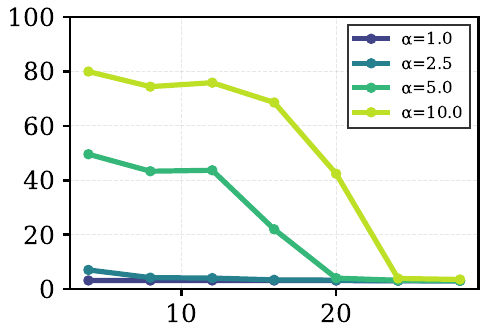} &
\includegraphics[width=0.16\textwidth]{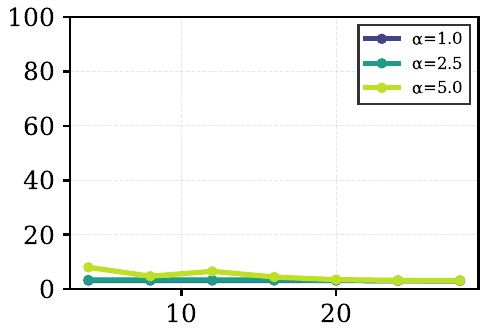} &
\includegraphics[width=0.16\textwidth]{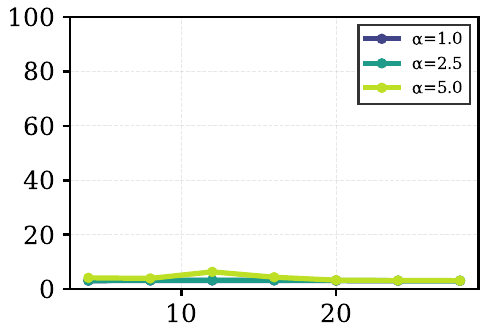} &
\includegraphics[width=0.16\textwidth]{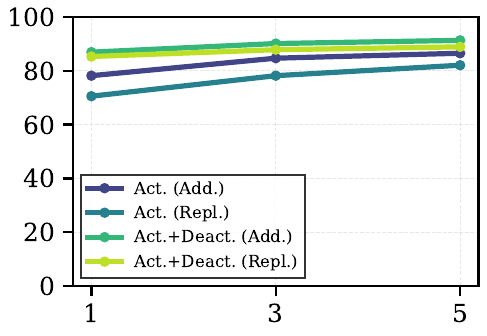} &
\includegraphics[width=0.16\textwidth]{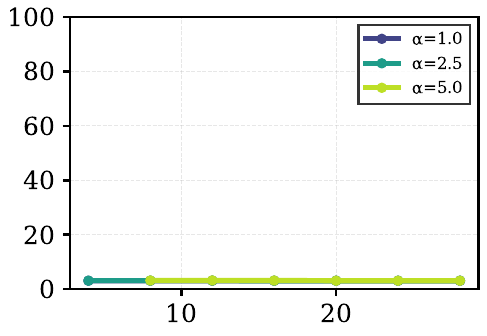} \\[-1pt]
\raisebox{0.3cm}{\rotatebox{90}{\scriptsize \textbf{Judge Quality}}} &
\includegraphics[width=0.16\textwidth]{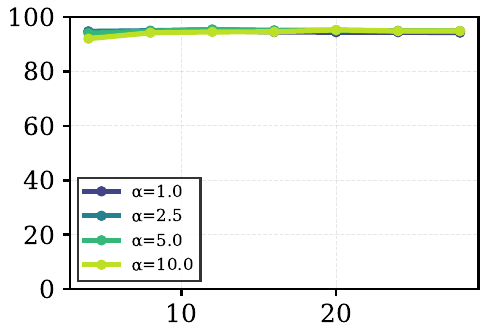} &
\includegraphics[width=0.16\textwidth]{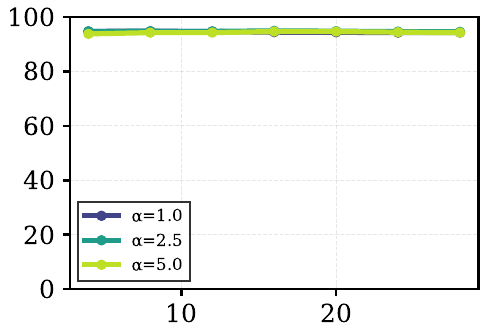} &
\includegraphics[width=0.16\textwidth]{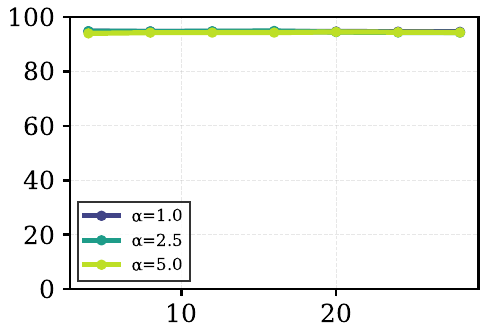} &
\includegraphics[width=0.16\textwidth]{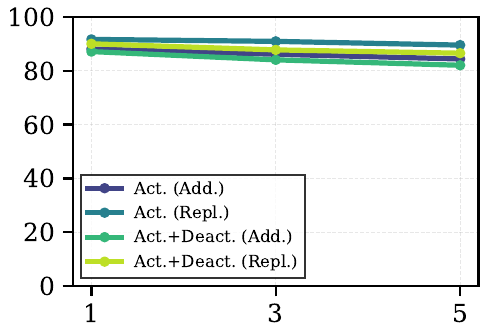} &
\includegraphics[width=0.16\textwidth]{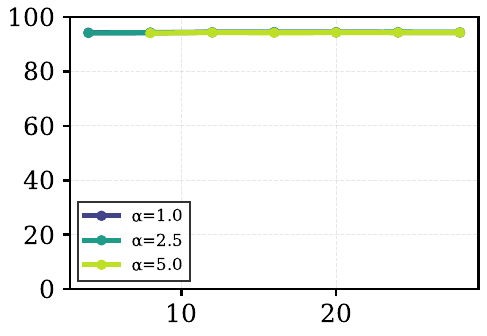} \\[-1pt]
\raisebox{0.3cm}{\rotatebox{90}{\scriptsize \textbf{Steering Score}}} &
\includegraphics[width=0.16\textwidth]{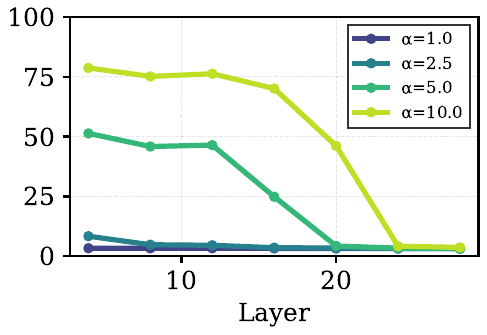} &
\includegraphics[width=0.16\textwidth]{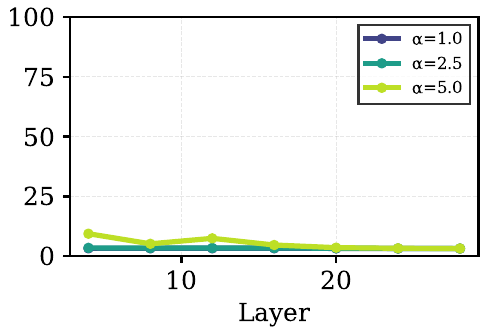} &
\includegraphics[width=0.16\textwidth]{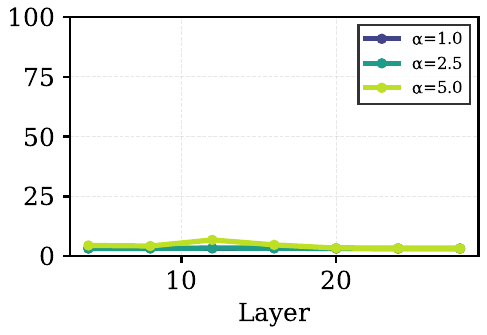} &
\includegraphics[width=0.16\textwidth]{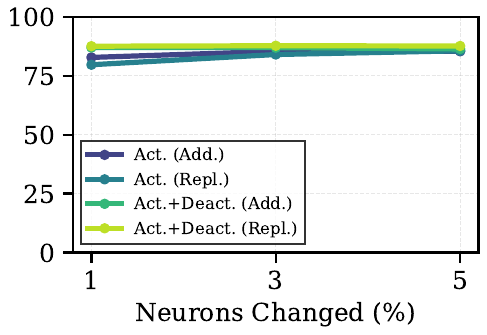} &
\includegraphics[width=0.16\textwidth]{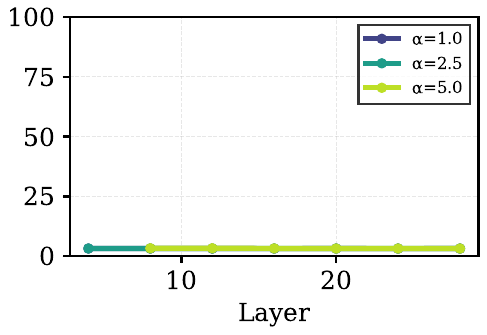}
\end{tabular}
}
\caption{Comparative analysis of steering methods across evaluation metrics for \texttt{Aya-Expanse-8B}. Columns show different methods (DiffMean, Probe, LDA, LAPE, PCA). Rows represent: language forcing success rate, judge relevance quality, and overall steering score.}
\label{fig:method_comparison_aya}
\end{figure*}
\clearpage

\begin{table*}[t]
\section{Selected Layers and Intervention Strengths}
\label{app:hyperparams}
\centering
\begin{tabular}{lcc}
\toprule
\textbf{Method} & \textbf{Selected Layer} & \textbf{Alpha Strength} \\
\midrule
$\mathcal{E}$ Base.-I & - & - \\
$\mathcal{E}$ Base.-II & - & - \\
$\vec{\Delta}$ DiffM. & 20 & 5.0 \\
$\mathbf{w}$ Probe & 20 & 5.0 \\
$\mathbf{u}$ PCA & 4 & 5.0 \\
$\vec{\Delta}$ SAE-DM. & 25 & 5.0 \\
$\mathbf{v}$ LDA & 16 & 5.0 \\
$\odot$ LAPE & - & 1.0 \\
\bottomrule
\end{tabular}
\caption{Hyperparameters--selected layer and alpha strength--for each steering method, based on the ablation results in Section \ref{sec:ablation}, for \texttt{Llama-3.1-8B}.}
\label{tab:hyperparameters_llama}
\end{table*}

\begin{table*}[t]
\centering
\begin{tabular}{lcc}
\toprule
\textbf{Method} & \textbf{Selected Layer} & \textbf{Alpha Strength} \\
\midrule
$\mathcal{E}$ Base.-I & - & - \\
$\mathcal{E}$ Base.-II & - & - \\
$\vec{\Delta}$ DiffM. & 4 & 10.0 \\
$\mathbf{w}$ Probe & 4 & 5.0 \\
$\mathbf{u}$ PCA & 12 & 5.0 \\
$\mathbf{v}$ LDA & 12 & 5.0 \\
$\odot$ LAPE & - & 1.0 \\
\bottomrule
\end{tabular}
\caption{Hyperparameters--selected layer and alpha strength--for each steering method, based on the ablation results in Appendix \ref{app:ablations}, for \texttt{Aya-Expanse-8B}.}
\label{tab:hyperparameters_aya}
\end{table*}

\begin{table*}[t!]
\section{Per-language Forcing and Judge Scores}
\label{app:scores}
\subsection{Llama-3.1-8B}
\centering
\small
\begin{tabular}{l|cccccccc}
\toprule
\textbf{Lang.} & $\color{blue}\mathcal{E}$ Base.-I & $\color{green}\mathcal{T}$ Base.-II & $\color{orange}\vec{\Delta}$ DiffM. & $\color{purple}\mathbf{w}$ Probe & $\color{cyan}\mathbf{u}$ PCA & $\color{brown}\vec{\Delta}$ SAE-DM. & $\color{blue}\mathbf{v}$ LDA & $\color{red}\odot$ LAPE \\
\midrule
ar & \cellcolor{gray!40}47.4 & \cellcolor{gray!40}38.2 & \cellcolor{yellow!25}\textbf{97.1} & 7.9 & 35.0 & 51.1 & 23.1 & \cellcolor{orange!20}92.9 \\
bo & \cellcolor{gray!40}\textbf{64.6} & \cellcolor{gray!40}26.7 & \cellcolor{yellow!25}4.5 & 2.9 & 6.5 & 3.2 & 2.5 & \cellcolor{orange!20}4.2 \\
cs & \cellcolor{gray!40}54.2 & \cellcolor{gray!40}62.7 & \cellcolor{yellow!25}\textbf{94.7} & 17.3 & 76.3 & 27.2 & 16.8 & \cellcolor{orange!20}89.3 \\
da & \cellcolor{gray!40}69.0 & \cellcolor{gray!40}32.1 & \cellcolor{yellow!25}\textbf{92.8} & 22.9 & 85.5 & 20.9 & 4.7 & \cellcolor{orange!20}38.2 \\
de & \cellcolor{gray!40}55.7 & \cellcolor{gray!40}40.6 & \cellcolor{yellow!25}\textbf{97.4} & 82.0 & 75.8 & 38.7 & - & \cellcolor{orange!20}96.6 \\
el & \cellcolor{gray!40}73.3 & \cellcolor{gray!40}65.3 & \cellcolor{yellow!25}\textbf{96.0} & 4.8 & 31.5 & 29.4 & 3.2 & \cellcolor{orange!20}94.7 \\
en & \cellcolor{gray!40}3.7 & \cellcolor{gray!40}3.7 & \cellcolor{yellow!25}96.3 & 96.0 & 31.5 & 36.1 & 48.3 & \cellcolor{orange!20}\textbf{99.5} \\
es & \cellcolor{gray!40}61.1 & \cellcolor{gray!40}49.8 & \cellcolor{yellow!25}\textbf{96.5} & 71.7 & 64.7 & 44.7 & 32.9 & \cellcolor{orange!20}91.9 \\
fa & \cellcolor{gray!40}24.0 & \cellcolor{gray!40}37.2 & \cellcolor{yellow!25}72.4 & 7.2 & 15.5 & 39.1 & 5.5 & \cellcolor{orange!20}\textbf{84.4} \\
fr & \cellcolor{gray!40}51.3 & \cellcolor{gray!40}51.8 & \cellcolor{yellow!25}\textbf{98.2} & 89.0 & 62.9 & 53.7 & 13.6 & \cellcolor{orange!20}96.6 \\
hi & \cellcolor{gray!40}36.5 & \cellcolor{gray!40}53.2 & \cellcolor{yellow!25}96.4 & 7.3 & 35.8 & 48.0 & - & \cellcolor{orange!20}\textbf{97.3} \\
id & \cellcolor{gray!40}49.8 & \cellcolor{gray!40}64.3 & \cellcolor{yellow!25}\textbf{97.0} & 46.0 & 33.2 & 10.1 & - & \cellcolor{orange!20}95.1 \\
it & \cellcolor{gray!40}50.8 & \cellcolor{gray!40}56.8 & \cellcolor{yellow!25}\textbf{97.5} & 73.6 & 77.7 & 41.5 & 8.8 & \cellcolor{orange!20}96.3 \\
ja & \cellcolor{gray!40}32.4 & \cellcolor{gray!40}59.1 & \cellcolor{yellow!25}85.9 & 22.5 & 32.8 & 5.3 & 9.4 & \cellcolor{orange!20}\textbf{92.4} \\
ka & \cellcolor{gray!40}\textbf{91.9} & \cellcolor{gray!40}59.1 & \cellcolor{yellow!25}21.3 & 3.2 & 11.2 & 3.2 & 2.0 & \cellcolor{orange!20}31.3 \\
kk & \cellcolor{gray!40}\textbf{75.2} & \cellcolor{gray!40}45.6 & \cellcolor{yellow!25}32.3 & 3.3 & 17.2 & 4.7 & 3.2 & \cellcolor{orange!20}71.4 \\
ko & \cellcolor{gray!40}27.7 & \cellcolor{gray!40}51.8 & \cellcolor{yellow!25}\textbf{99.5} & 26.0 & 50.7 & 23.1 & 11.9 & \cellcolor{orange!20}94.7 \\
mt & \cellcolor{gray!40}\textbf{94.8} & \cellcolor{gray!40}61.7 & \cellcolor{yellow!25}8.7 & 3.0 & 8.6 & 3.2 & 3.0 & \cellcolor{orange!20}12.9 \\
nl & \cellcolor{gray!40}52.3 & \cellcolor{gray!40}51.8 & \cellcolor{yellow!25}\textbf{97.7} & 31.6 & 64.9 & 30.4 & 4.1 & \cellcolor{orange!20}96.0 \\
no & \cellcolor{gray!40}47.6 & \cellcolor{gray!40}39.7 & \cellcolor{yellow!25}\textbf{90.3} & 17.8 & 83.0 & 27.3 & 2.3 & \cellcolor{orange!20}56.9 \\
pl & \cellcolor{gray!40}37.6 & \cellcolor{gray!40}38.0 & \cellcolor{yellow!25}\textbf{95.9} & 21.3 & 79.2 & 30.1 & 33.6 & \cellcolor{orange!20}91.2 \\
pt & \cellcolor{gray!40}44.8 & \cellcolor{gray!40}67.1 & \cellcolor{yellow!25}\textbf{97.7} & 81.9 & 48.3 & 34.5 & 41.2 & \cellcolor{orange!20}95.9 \\
ro & \cellcolor{gray!40}55.5 & \cellcolor{gray!40}75.5 & \cellcolor{yellow!25}\textbf{96.5} & 44.0 & 70.9 & 23.2 & 3.3 & \cellcolor{orange!20}71.3 \\
ru & \cellcolor{gray!40}41.3 & \cellcolor{gray!40}54.8 & \cellcolor{yellow!25}\textbf{99.2} & 78.3 & 55.2 & 62.6 & 43.1 & \cellcolor{orange!20}97.6 \\
sk & \cellcolor{gray!40}75.9 & \cellcolor{gray!40}37.1 & \cellcolor{yellow!25}77.9 & 19.9 & \textbf{82.0} & 10.4 & 5.9 & \cellcolor{orange!20}57.8 \\
sv & \cellcolor{gray!40}47.7 & \cellcolor{gray!40}48.0 & \cellcolor{yellow!25}\textbf{95.9} & 33.4 & 78.2 & 26.5 & 46.1 & \cellcolor{orange!20}86.3 \\
sw & \cellcolor{gray!40}\textbf{95.8} & \cellcolor{gray!40}83.3 & \cellcolor{yellow!25}51.7 & 4.4 & 60.3 & 3.2 & 3.2 & \cellcolor{orange!20}49.3 \\
th & \cellcolor{gray!40}60.5 & \cellcolor{gray!40}80.6 & \cellcolor{yellow!25}\textbf{99.1} & 14.1 & 37.8 & 36.6 & 5.2 & \cellcolor{orange!20}96.3 \\
tr & \cellcolor{gray!40}66.6 & \cellcolor{gray!40}60.8 & \cellcolor{yellow!25}93.3 & 13.6 & 32.4 & 11.4 & 3.8 & \cellcolor{orange!20}\textbf{94.7} \\
uk & \cellcolor{gray!40}52.2 & \cellcolor{gray!40}70.7 & \cellcolor{yellow!25}93.2 & 7.3 & 63.8 & 76.4 & 8.9 & \cellcolor{orange!20}\textbf{98.7} \\
vi & \cellcolor{gray!40}68.8 & \cellcolor{gray!40}82.7 & \cellcolor{yellow!25}\textbf{99.3} & 42.8 & 31.8 & 6.0 & 5.9 & \cellcolor{orange!20}96.2 \\
zh & \cellcolor{gray!40}44.1 & \cellcolor{gray!40}63.5 & \cellcolor{yellow!25}\textbf{99.4} & 92.3 & 16.1 & 83.5 & 94.4 & \cellcolor{orange!20}97.4 \\
\midrule
Avg. & \cellcolor{gray!40}54.8 & \cellcolor{gray!40}53.5 & \cellcolor{yellow!25}\textbf{83.5} & 34.0 & 48.6 & 29.5 & 13.9 & \cellcolor{orange!20}80.2 \\
\bottomrule
\end{tabular}
\caption{\textbf{Language forcing scores} across all methods for 32 ablation languages for \texttt{Llama-3.1-8B-Instruct}. }
\label{tab:forcing_llama}
\end{table*}

\begin{table*}[t!]
\centering
\small
\begin{tabular}{l|cccccccc}
\toprule
\textbf{Lang.} & $\color{blue}\mathcal{E}$ Base.-I & $\color{green}\mathcal{T}$ Base.-II & $\color{orange}\vec{\Delta}$ DiffM. & $\color{purple}\mathbf{w}$ Probe & $\color{cyan}\mathbf{u}$ PCA & $\color{brown}\vec{\Delta}$ SAE-DM. & $\color{blue}\mathbf{v}$ LDA & $\color{red}\odot$ LAPE \\
\midrule
ar & \cellcolor{gray!40}\textbf{93.5} & \cellcolor{gray!40}91.7 & \cellcolor{yellow!25}81.8 & 85.2 & 10.6 & 47.9 & 71.2 & \cellcolor{orange!20}66.0 \\
bo & \cellcolor{gray!40}22.6 & \cellcolor{gray!40}67.3 & \cellcolor{yellow!25}79.1 & 84.2 & 6.8 & \textbf{85.4} & 74.0 & \cellcolor{orange!20}43.1 \\
cs & \cellcolor{gray!40}\textbf{95.4} & \cellcolor{gray!40}93.8 & \cellcolor{yellow!25}90.2 & 82.7 & 9.3 & 71.9 & 73.9 & \cellcolor{orange!20}88.8 \\
da & \cellcolor{gray!40}\textbf{96.8} & \cellcolor{gray!40}92.5 & \cellcolor{yellow!25}87.5 & 82.6 & 9.7 & 75.8 & 77.3 & \cellcolor{orange!20}90.1 \\
de & \cellcolor{gray!40}96.3 & \cellcolor{gray!40}93.3 & \cellcolor{yellow!25}93.5 & 84.8 & 8.2 & 75.8 & 88.0 & \cellcolor{orange!20}\textbf{96.3} \\
el & \cellcolor{gray!40}\textbf{93.9} & \cellcolor{gray!40}92.0 & \cellcolor{yellow!25}81.9 & 87.9 & 10.3 & 57.7 & 78.0 & \cellcolor{orange!20}80.7 \\
en & \cellcolor{gray!40}92.6 & \cellcolor{gray!40}92.5 & \cellcolor{yellow!25}95.4 & 95.5 & 6.7 & 87.9 & 54.5 & \cellcolor{orange!20}\textbf{98.1} \\
es & \cellcolor{gray!40}96.1 & \cellcolor{gray!40}94.9 & \cellcolor{yellow!25}92.9 & 87.2 & 12.2 & 81.3 & 85.8 & \cellcolor{orange!20}\textbf{96.1} \\
fa & \cellcolor{gray!40}\textbf{93.9} & \cellcolor{gray!40}93.8 & \cellcolor{yellow!25}86.9 & 84.3 & 12.6 & 60.2 & 77.0 & \cellcolor{orange!20}85.9 \\
fr & \cellcolor{gray!40}95.5 & \cellcolor{gray!40}93.5 & \cellcolor{yellow!25}94.8 & 90.4 & 11.4 & 81.1 & 82.8 & \cellcolor{orange!20}\textbf{96.4} \\
hi & \cellcolor{gray!40}93.0 & \cellcolor{gray!40}94.0 & \cellcolor{yellow!25}85.2 & 84.8 & 7.0 & 61.1 & \textbf{94.3} & \cellcolor{orange!20}89.4 \\
id & \cellcolor{gray!40}\textbf{96.4} & \cellcolor{gray!40}95.3 & \cellcolor{yellow!25}92.5 & 82.3 & 12.7 & 81.7 & 95.7 & \cellcolor{orange!20}94.2 \\
it & \cellcolor{gray!40}94.9 & \cellcolor{gray!40}95.0 & \cellcolor{yellow!25}92.2 & 81.8 & 13.8 & 74.8 & 79.1 & \cellcolor{orange!20}\textbf{95.5} \\
ja & \cellcolor{gray!40}92.4 & \cellcolor{gray!40}\textbf{94.0} & \cellcolor{yellow!25}83.4 & 84.7 & 9.1 & 83.1 & 82.1 & \cellcolor{orange!20}76.1 \\
ka & \cellcolor{gray!40}71.2 & \cellcolor{gray!40}79.8 & \cellcolor{yellow!25}71.9 & \textbf{86.5} & 12.4 & 85.7 & 45.4 & \cellcolor{orange!20}24.3 \\
kk & \cellcolor{gray!40}55.3 & \cellcolor{gray!40}74.5 & \cellcolor{yellow!25}60.7 & \textbf{87.3} & 3.7 & 69.0 & 81.4 & \cellcolor{orange!20}17.9 \\
ko & \cellcolor{gray!40}\textbf{93.2} & \cellcolor{gray!40}92.6 & \cellcolor{yellow!25}81.8 & 85.0 & 10.2 & 77.2 & 82.6 & \cellcolor{orange!20}69.0 \\
mt & \cellcolor{gray!40}53.8 & \cellcolor{gray!40}67.6 & \cellcolor{yellow!25}72.9 & \textbf{85.2} & 6.4 & 79.4 & 69.7 & \cellcolor{orange!20}53.7 \\
nl & \cellcolor{gray!40}\textbf{96.0} & \cellcolor{gray!40}94.2 & \cellcolor{yellow!25}93.0 & 83.7 & 7.7 & 76.2 & 84.1 & \cellcolor{orange!20}94.0 \\
no & \cellcolor{gray!40}\textbf{94.7} & \cellcolor{gray!40}93.8 & \cellcolor{yellow!25}87.5 & 81.8 & 10.1 & 66.1 & 86.9 & \cellcolor{orange!20}89.3 \\
pl & \cellcolor{gray!40}\textbf{95.2} & \cellcolor{gray!40}92.6 & \cellcolor{yellow!25}89.3 & 83.1 & 5.8 & 66.3 & 75.2 & \cellcolor{orange!20}88.7 \\
pt & \cellcolor{gray!40}\textbf{95.8} & \cellcolor{gray!40}95.7 & \cellcolor{yellow!25}94.0 & 87.5 & 11.7 & 80.9 & 86.5 & \cellcolor{orange!20}95.3 \\
ro & \cellcolor{gray!40}\textbf{95.4} & \cellcolor{gray!40}93.8 & \cellcolor{yellow!25}89.8 & 84.6 & 7.6 & 71.1 & 77.6 & \cellcolor{orange!20}84.3 \\
ru & \cellcolor{gray!40}94.3 & \cellcolor{gray!40}94.0 & \cellcolor{yellow!25}\textbf{94.8} & 87.1 & 9.9 & 73.6 & 83.3 & \cellcolor{orange!20}93.4 \\
sk & \cellcolor{gray!40}\textbf{92.8} & \cellcolor{gray!40}92.2 & \cellcolor{yellow!25}77.5 & 79.1 & 4.1 & 65.4 & 70.6 & \cellcolor{orange!20}76.5 \\
sv & \cellcolor{gray!40}\textbf{95.8} & \cellcolor{gray!40}94.2 & \cellcolor{yellow!25}91.6 & 83.3 & 9.5 & 75.9 & 86.9 & \cellcolor{orange!20}93.6 \\
sw & \cellcolor{gray!40}74.5 & \cellcolor{gray!40}76.3 & \cellcolor{yellow!25}48.0 & 83.6 & 1.1 & \textbf{84.6} & 77.8 & \cellcolor{orange!20}56.6 \\
th & \cellcolor{gray!40}\textbf{93.4} & \cellcolor{gray!40}92.3 & \cellcolor{yellow!25}84.5 & 84.6 & 7.4 & 69.9 & 85.9 & \cellcolor{orange!20}75.2 \\
tr & \cellcolor{gray!40}\textbf{92.5} & \cellcolor{gray!40}91.4 & \cellcolor{yellow!25}77.2 & 82.7 & 9.5 & 73.0 & 83.6 & \cellcolor{orange!20}75.6 \\
uk & \cellcolor{gray!40}\textbf{94.6} & \cellcolor{gray!40}93.9 & \cellcolor{yellow!25}92.6 & 87.6 & 8.8 & 64.1 & 82.2 & \cellcolor{orange!20}85.4 \\
vi & \cellcolor{gray!40}\textbf{95.3} & \cellcolor{gray!40}94.9 & \cellcolor{yellow!25}94.6 & 81.4 & 9.6 & 85.5 & 81.3 & \cellcolor{orange!20}93.8 \\
zh & \cellcolor{gray!40}95.9 & \cellcolor{gray!40}\textbf{96.6} & \cellcolor{yellow!25}95.9 & 92.4 & 9.8 & 87.4 & 81.8 & \cellcolor{orange!20}95.1 \\
\midrule
Avg. & \cellcolor{gray!40}88.5 & \cellcolor{gray!40}\textbf{90.4} & \cellcolor{yellow!25}85.5 & 85.2 & 8.9 & 74.3 & 79.3 & \cellcolor{orange!20}80.0 \\
\bottomrule
\end{tabular}
\caption{\textbf{Output relevance scores} across all methods for 32 languages for \texttt{Llama-3.1-8B-Instruct}.}
\label{tab:judge_llama}
\end{table*}

\begin{table*}[t!]
\subsection{Aya-Expanse-8B}
\centering
\small
\begin{tabular}{l|ccccccc}
\toprule
\textbf{Lang.} & $\color{blue}\mathcal{E}$ Base.-I & $\color{green}\mathcal{T}$ Base.-II & $\color{orange}\vec{\Delta}$ DiffM. & $\color{purple}\mathbf{w}$ Probe & $\color{cyan}\mathbf{u}$ PCA & $\color{blue}\mathbf{v}$ LDA & $\color{red}\odot$ LAPE \\
\midrule
ar & \cellcolor{gray!40}\textbf{100.0} & \cellcolor{gray!40}100.0 & \cellcolor{yellow!25}90.4 & 15.0 & 3.2 & 8.4 & \cellcolor{orange!20}99.6 \\
bo & \cellcolor{gray!40}5.8 & \cellcolor{gray!40}4.6 & \cellcolor{yellow!25}\textbf{9.2} & 4.6 & - & - & \cellcolor{orange!20}1.8 \\
cs & \cellcolor{gray!40}99.9 & \cellcolor{gray!40}99.5 & \cellcolor{yellow!25}96.9 & 43.0 & 1.7 & 1.3 & \cellcolor{orange!20}\textbf{100.0} \\
da & \cellcolor{gray!40}99.5 & \cellcolor{gray!40}96.4 & \cellcolor{yellow!25}68.8 & 3.6 & 3.5 & 3.0 & \cellcolor{orange!20}\textbf{99.7} \\
de & \cellcolor{gray!40}\textbf{100.0} & \cellcolor{gray!40}100.0 & \cellcolor{yellow!25}97.3 & 7.4 & 3.2 & 4.6 & \cellcolor{orange!20}100.0 \\
el & \cellcolor{gray!40}\textbf{100.0} & \cellcolor{gray!40}100.0 & \cellcolor{yellow!25}96.9 & 8.6 & 3.2 & 3.2 & \cellcolor{orange!20}100.0 \\
en & \cellcolor{gray!40}89.1 & \cellcolor{gray!40}89.1 & \cellcolor{yellow!25}96.7 & 6.5 & 5.0 & 77.8 & \cellcolor{orange!20}\textbf{100.0} \\
es & \cellcolor{gray!40}99.5 & \cellcolor{gray!40}\textbf{100.0} & \cellcolor{yellow!25}97.0 & 8.0 & 3.2 & 3.3 & \cellcolor{orange!20}100.0 \\
fa & \cellcolor{gray!40}\textbf{99.4} & \cellcolor{gray!40}98.8 & \cellcolor{yellow!25}98.4 & 8.7 & 3.1 & 3.2 & \cellcolor{orange!20}99.3 \\
fr & \cellcolor{gray!40}\textbf{100.0} & \cellcolor{gray!40}100.0 & \cellcolor{yellow!25}99.2 & 8.9 & 3.2 & 3.2 & \cellcolor{orange!20}100.0 \\
hi & \cellcolor{gray!40}\textbf{99.9} & \cellcolor{gray!40}99.4 & \cellcolor{yellow!25}99.7 & 6.7 & 5.8 & 6.5 & \cellcolor{orange!20}99.8 \\
id & \cellcolor{gray!40}99.8 & \cellcolor{gray!40}\textbf{100.0} & \cellcolor{yellow!25}98.8 & 4.5 & 3.4 & 3.3 & \cellcolor{orange!20}100.0 \\
it & \cellcolor{gray!40}99.9 & \cellcolor{gray!40}\textbf{100.0} & \cellcolor{yellow!25}97.1 & 5.8 & 5.1 & 4.9 & \cellcolor{orange!20}100.0 \\
ja & \cellcolor{gray!40}96.0 & \cellcolor{gray!40}95.8 & \cellcolor{yellow!25}96.0 & 9.9 & 3.2 & 10.2 & \cellcolor{orange!20}\textbf{97.8} \\
ka & \cellcolor{gray!40}\textbf{98.2} & \cellcolor{gray!40}80.5 & \cellcolor{yellow!25}19.1 & 3.2 & 3.2 & 3.2 & \cellcolor{orange!20}62.0 \\
kk & \cellcolor{gray!40}\textbf{69.1} & \cellcolor{gray!40}46.1 & \cellcolor{yellow!25}13.8 & 13.8 & 13.8 & 13.8 & \cellcolor{orange!20}1.2 \\
ko & \cellcolor{gray!40}\textbf{100.0} & \cellcolor{gray!40}100.0 & \cellcolor{yellow!25}100.0 & 12.9 & 3.2 & 4.7 & \cellcolor{orange!20}100.0 \\
mt & \cellcolor{gray!40}\textbf{93.5} & \cellcolor{gray!40}11.9 & \cellcolor{yellow!25}30.6 & 3.0 & 2.6 & 2.6 & \cellcolor{orange!20}78.6 \\
nl & \cellcolor{gray!40}\textbf{100.0} & \cellcolor{gray!40}100.0 & \cellcolor{yellow!25}99.5 & 13.4 & 3.2 & 3.2 & \cellcolor{orange!20}100.0 \\
no & \cellcolor{gray!40}\textbf{99.5} & \cellcolor{gray!40}99.2 & \cellcolor{yellow!25}70.0 & 3.6 & 3.7 & 3.5 & \cellcolor{orange!20}85.1 \\
pl & \cellcolor{gray!40}99.8 & \cellcolor{gray!40}\textbf{100.0} & \cellcolor{yellow!25}96.7 & 20.6 & 3.2 & 3.2 & \cellcolor{orange!20}100.0 \\
pt & \cellcolor{gray!40}99.8 & \cellcolor{gray!40}\textbf{100.0} & \cellcolor{yellow!25}97.0 & 5.1 & 3.2 & 3.2 & \cellcolor{orange!20}100.0 \\
ro & \cellcolor{gray!40}\textbf{100.0} & \cellcolor{gray!40}99.7 & \cellcolor{yellow!25}97.0 & 10.7 & 3.2 & 3.2 & \cellcolor{orange!20}99.9 \\
ru & \cellcolor{gray!40}99.7 & \cellcolor{gray!40}99.1 & \cellcolor{yellow!25}96.5 & 5.8 & 3.9 & 4.6 & \cellcolor{orange!20}\textbf{100.0} \\
sk & \cellcolor{gray!40}85.0 & \cellcolor{gray!40}95.2 & \cellcolor{yellow!25}60.5 & 1.5 & 2.7 & 2.6 & \cellcolor{orange!20}\textbf{98.5} \\
sv & \cellcolor{gray!40}99.9 & \cellcolor{gray!40}\textbf{99.9} & \cellcolor{yellow!25}94.9 & 3.6 & 3.4 & 3.3 & \cellcolor{orange!20}99.8 \\
sw & \cellcolor{gray!40}98.8 & \cellcolor{gray!40}83.7 & \cellcolor{yellow!25}34.1 & 3.2 & 3.1 & 3.2 & \cellcolor{orange!20}\textbf{99.5} \\
th & \cellcolor{gray!40}99.9 & \cellcolor{gray!40}99.8 & \cellcolor{yellow!25}45.9 & 3.4 & 3.2 & 3.3 & \cellcolor{orange!20}\textbf{100.0} \\
tr & \cellcolor{gray!40}\textbf{100.0} & \cellcolor{gray!40}99.7 & \cellcolor{yellow!25}96.3 & 7.3 & 3.2 & 3.3 & \cellcolor{orange!20}100.0 \\
uk & \cellcolor{gray!40}100.0 & \cellcolor{gray!40}\textbf{100.0} & \cellcolor{yellow!25}96.9 & 6.9 & 3.3 & 3.3 & \cellcolor{orange!20}100.0 \\
vi & \cellcolor{gray!40}99.9 & \cellcolor{gray!40}99.9 & \cellcolor{yellow!25}95.9 & 8.3 & 3.2 & 3.3 & \cellcolor{orange!20}\textbf{100.0} \\
zh & \cellcolor{gray!40}99.8 & \cellcolor{gray!40}99.7 & \cellcolor{yellow!25}94.2 & 6.6 & 3.4 & 16.0 & \cellcolor{orange!20}\textbf{100.0} \\
\midrule
Avg. & \cellcolor{gray!40}\textbf{94.7} & \cellcolor{gray!40}90.6 & \cellcolor{yellow!25}80.7 & 8.6 & 3.7 & 7.0 & \cellcolor{orange!20}91.3 \\
\bottomrule
\end{tabular}
\caption{\textbf{Language forcing scores} across all methods for 32 ablation languages for \texttt{Aya-Expanse-8B}.}
\label{tab:forcing_aya}
\end{table*}

\begin{table*}[t!]
\centering
\small
\begin{tabular}{l|ccccccc}
\toprule
\textbf{Lang.} & $\color{blue}\mathcal{E}$ Base.-I & $\color{green}\mathcal{T}$ Base.-II & $\color{orange}\vec{\Delta}$ DiffM. & $\color{purple}\mathbf{w}$ Probe & $\color{cyan}\mathbf{u}$ PCA & $\color{blue}\mathbf{v}$ LDA & $\color{red}\odot$ LAPE \\
\midrule
ar & \cellcolor{gray!40}96.3 & \cellcolor{gray!40}\textbf{98.2} & \cellcolor{yellow!25}95.1 & 94.2 & 94.6 & 94.6 & \cellcolor{orange!20}94.9 \\
bo & \cellcolor{gray!40}11.6 & \cellcolor{gray!40}\textbf{95.0} & \cellcolor{yellow!25}89.3 & 94.0 & 94.4 & 94.7 & \cellcolor{orange!20}45.2 \\
cs & \cellcolor{gray!40}97.0 & \cellcolor{gray!40}\textbf{98.3} & \cellcolor{yellow!25}94.8 & 94.9 & 94.8 & 94.2 & \cellcolor{orange!20}94.3 \\
da & \cellcolor{gray!40}96.3 & \cellcolor{gray!40}\textbf{96.8} & \cellcolor{yellow!25}93.5 & 93.5 & 94.8 & 95.2 & \cellcolor{orange!20}87.9 \\
de & \cellcolor{gray!40}97.0 & \cellcolor{gray!40}\textbf{98.2} & \cellcolor{yellow!25}94.8 & 94.0 & 94.5 & 94.8 & \cellcolor{orange!20}95.1 \\
el & \cellcolor{gray!40}95.8 & \cellcolor{gray!40}\textbf{98.0} & \cellcolor{yellow!25}94.4 & 94.7 & 94.5 & 94.4 & \cellcolor{orange!20}94.1 \\
en & \cellcolor{gray!40}98.2 & \cellcolor{gray!40}\textbf{98.3} & \cellcolor{yellow!25}95.8 & 94.7 & 94.4 & 95.0 & \cellcolor{orange!20}96.7 \\
es & \cellcolor{gray!40}96.0 & \cellcolor{gray!40}\textbf{97.9} & \cellcolor{yellow!25}95.5 & 94.3 & 94.6 & 95.0 & \cellcolor{orange!20}95.5 \\
fa & \cellcolor{gray!40}96.1 & \cellcolor{gray!40}\textbf{97.6} & \cellcolor{yellow!25}94.4 & 94.3 & 94.3 & 94.7 & \cellcolor{orange!20}93.3 \\
fr & \cellcolor{gray!40}97.1 & \cellcolor{gray!40}\textbf{98.5} & \cellcolor{yellow!25}95.1 & 93.1 & 93.7 & 95.3 & \cellcolor{orange!20}95.2 \\
hi & \cellcolor{gray!40}96.2 & \cellcolor{gray!40}\textbf{97.2} & \cellcolor{yellow!25}94.7 & 93.9 & 93.8 & 94.3 & \cellcolor{orange!20}94.1 \\
id & \cellcolor{gray!40}96.3 & \cellcolor{gray!40}\textbf{98.3} & \cellcolor{yellow!25}94.9 & 93.8 & 93.9 & 94.9 & \cellcolor{orange!20}95.3 \\
it & \cellcolor{gray!40}97.1 & \cellcolor{gray!40}\textbf{97.4} & \cellcolor{yellow!25}94.8 & 94.4 & 94.5 & 94.5 & \cellcolor{orange!20}94.1 \\
ja & \cellcolor{gray!40}95.8 & \cellcolor{gray!40}\textbf{96.2} & \cellcolor{yellow!25}93.9 & 93.8 & 95.0 & 93.8 & \cellcolor{orange!20}94.5 \\
ka & \cellcolor{gray!40}37.2 & \cellcolor{gray!40}46.7 & \cellcolor{yellow!25}81.7 & 92.9 & \textbf{94.7} & 93.7 & \cellcolor{orange!20}31.7 \\
kk & \cellcolor{gray!40}47.8 & \cellcolor{gray!40}55.5 & \cellcolor{yellow!25}75.8 & 92.9 & \textbf{94.3} & 93.6 & \cellcolor{orange!20}13.8 \\
ko & \cellcolor{gray!40}96.0 & \cellcolor{gray!40}\textbf{96.2} & \cellcolor{yellow!25}94.0 & 93.5 & 93.5 & 93.2 & \cellcolor{orange!20}93.8 \\
mt & \cellcolor{gray!40}73.9 & \cellcolor{gray!40}92.5 & \cellcolor{yellow!25}81.2 & 92.8 & 93.3 & \textbf{93.7} & \cellcolor{orange!20}42.0 \\
nl & \cellcolor{gray!40}97.0 & \cellcolor{gray!40}\textbf{98.3} & \cellcolor{yellow!25}95.8 & 93.5 & 94.2 & 94.3 & \cellcolor{orange!20}95.3 \\
no & \cellcolor{gray!40}\textbf{96.4} & \cellcolor{gray!40}95.3 & \cellcolor{yellow!25}92.3 & 92.2 & 93.9 & 93.8 & \cellcolor{orange!20}85.5 \\
pl & \cellcolor{gray!40}96.8 & \cellcolor{gray!40}\textbf{97.5} & \cellcolor{yellow!25}95.5 & 94.3 & 94.4 & 93.7 & \cellcolor{orange!20}93.5 \\
pt & \cellcolor{gray!40}96.6 & \cellcolor{gray!40}\textbf{98.0} & \cellcolor{yellow!25}95.6 & 93.7 & 94.1 & 94.7 & \cellcolor{orange!20}95.5 \\
ro & \cellcolor{gray!40}96.6 & \cellcolor{gray!40}\textbf{98.0} & \cellcolor{yellow!25}95.6 & 94.1 & 93.9 & 93.0 & \cellcolor{orange!20}94.1 \\
ru & \cellcolor{gray!40}96.5 & \cellcolor{gray!40}\textbf{98.1} & \cellcolor{yellow!25}95.0 & 93.9 & 94.0 & 94.2 & \cellcolor{orange!20}94.3 \\
sk & \cellcolor{gray!40}96.0 & \cellcolor{gray!40}\textbf{96.9} & \cellcolor{yellow!25}94.7 & 94.6 & 95.0 & 93.9 & \cellcolor{orange!20}89.7 \\
sv & \cellcolor{gray!40}96.7 & \cellcolor{gray!40}\textbf{97.5} & \cellcolor{yellow!25}94.6 & 93.0 & 93.9 & 94.0 & \cellcolor{orange!20}87.4 \\
sw & \cellcolor{gray!40}48.0 & \cellcolor{gray!40}60.9 & \cellcolor{yellow!25}77.5 & 92.5 & 93.8 & \textbf{94.8} & \cellcolor{orange!20}14.2 \\
th & \cellcolor{gray!40}73.3 & \cellcolor{gray!40}71.9 & \cellcolor{yellow!25}78.3 & 92.5 & \textbf{94.7} & 93.5 & \cellcolor{orange!20}48.5 \\
tr & \cellcolor{gray!40}96.2 & \cellcolor{gray!40}\textbf{96.4} & \cellcolor{yellow!25}93.4 & 94.3 & 94.4 & 94.3 & \cellcolor{orange!20}93.8 \\
uk & \cellcolor{gray!40}96.5 & \cellcolor{gray!40}\textbf{97.4} & \cellcolor{yellow!25}94.7 & 95.1 & 95.2 & 94.3 & \cellcolor{orange!20}94.4 \\
vi & \cellcolor{gray!40}96.5 & \cellcolor{gray!40}\textbf{97.5} & \cellcolor{yellow!25}94.1 & 94.6 & 95.0 & 94.5 & \cellcolor{orange!20}94.1 \\
zh & \cellcolor{gray!40}\textbf{96.4} & \cellcolor{gray!40}96.2 & \cellcolor{yellow!25}94.7 & 93.6 & 93.9 & 94.3 & \cellcolor{orange!20}94.7 \\
\midrule
Avg. & \cellcolor{gray!40}87.5 & \cellcolor{gray!40}92.4 & \cellcolor{yellow!25}92.1 & 93.8 & \textbf{94.3} & 94.3 & \cellcolor{orange!20}82.1 \\
\bottomrule
\end{tabular}
\caption{\textbf{Output relevance scores} across all methods for 32 languages for  \texttt{Aya-Expanse-8B}.}
\label{tab:judge_aya}
\end{table*}

\begin{table*}[t!]
\centering
\small
\begin{tabular}{l|ccccccc}
\toprule
\textbf{Lang.} & $\color{blue}\mathcal{E}$ Base.-I & $\color{green}\mathcal{T}$ Base.-II & $\color{orange}\vec{\Delta}$ DiffM. & $\color{purple}\mathbf{w}$ Probe & $\color{cyan}\mathbf{u}$ PCA & $\color{blue}\mathbf{v}$ LDA & $\color{red}\odot$ LAPE \\
\midrule
ar & \cellcolor{gray!40}98.1 & \cellcolor{gray!40}\textbf{99.0} & \cellcolor{yellow!25}92.7 & 25.9 & 6.2 & 15.4 & \cellcolor{orange!20}97.2 \\
bo & \cellcolor{gray!40}7.8 & \cellcolor{gray!40}8.8 & \cellcolor{yellow!25}\textbf{16.7} & 8.8 & - & - & \cellcolor{orange!20}3.5 \\
cs & \cellcolor{gray!40}98.4 & \cellcolor{gray!40}\textbf{98.9} & \cellcolor{yellow!25}95.8 & 59.1 & 3.4 & 2.5 & \cellcolor{orange!20}97.0 \\
da & \cellcolor{gray!40}\textbf{97.9} & \cellcolor{gray!40}96.6 & \cellcolor{yellow!25}79.2 & 6.9 & 6.8 & 5.9 & \cellcolor{orange!20}93.4 \\
de & \cellcolor{gray!40}98.5 & \cellcolor{gray!40}\textbf{99.1} & \cellcolor{yellow!25}96.0 & 13.7 & 6.2 & 8.7 & \cellcolor{orange!20}97.5 \\
el & \cellcolor{gray!40}97.8 & \cellcolor{gray!40}\textbf{99.0} & \cellcolor{yellow!25}95.6 & 15.8 & 6.2 & 6.2 & \cellcolor{orange!20}97.0 \\
en & \cellcolor{gray!40}93.4 & \cellcolor{gray!40}93.5 & \cellcolor{yellow!25}96.2 & 12.2 & 9.5 & 85.6 & \cellcolor{orange!20}\textbf{98.3} \\
es & \cellcolor{gray!40}97.7 & \cellcolor{gray!40}\textbf{99.0} & \cellcolor{yellow!25}96.3 & 14.8 & 6.2 & 6.3 & \cellcolor{orange!20}97.7 \\
fa & \cellcolor{gray!40}97.7 & \cellcolor{gray!40}\textbf{98.2} & \cellcolor{yellow!25}96.3 & 15.9 & 6.1 & 6.2 & \cellcolor{orange!20}96.2 \\
fr & \cellcolor{gray!40}98.5 & \cellcolor{gray!40}\textbf{99.2} & \cellcolor{yellow!25}97.1 & 16.3 & 6.2 & 6.2 & \cellcolor{orange!20}97.5 \\
hi & \cellcolor{gray!40}98.0 & \cellcolor{gray!40}\textbf{98.3} & \cellcolor{yellow!25}97.2 & 12.5 & 10.8 & 12.1 & \cellcolor{orange!20}96.8 \\
id & \cellcolor{gray!40}98.0 & \cellcolor{gray!40}\textbf{99.2} & \cellcolor{yellow!25}96.8 & 8.5 & 6.5 & 6.3 & \cellcolor{orange!20}97.6 \\
it & \cellcolor{gray!40}98.5 & \cellcolor{gray!40}\textbf{98.7} & \cellcolor{yellow!25}96.0 & 11.0 & 9.7 & 9.4 & \cellcolor{orange!20}97.0 \\
ja & \cellcolor{gray!40}95.9 & \cellcolor{gray!40}96.0 & \cellcolor{yellow!25}95.0 & 17.9 & 6.2 & 18.4 & \cellcolor{orange!20}\textbf{96.1} \\
ka & \cellcolor{gray!40}53.9 & \cellcolor{gray!40}\textbf{59.1} & \cellcolor{yellow!25}30.9 & 6.2 & 6.2 & 6.2 & \cellcolor{orange!20}42.0 \\
kk & \cellcolor{gray!40}\textbf{56.5} & \cellcolor{gray!40}50.4 & \cellcolor{yellow!25}23.4 & 24.1 & 24.1 & 24.1 & \cellcolor{orange!20}2.2 \\
ko & \cellcolor{gray!40}98.0 & \cellcolor{gray!40}\textbf{98.0} & \cellcolor{yellow!25}96.9 & 22.6 & 6.2 & 8.9 & \cellcolor{orange!20}96.8 \\
mt & \cellcolor{gray!40}\textbf{82.5} & \cellcolor{gray!40}21.1 & \cellcolor{yellow!25}44.4 & 5.7 & 5.1 & 5.1 & \cellcolor{orange!20}54.8 \\
nl & \cellcolor{gray!40}98.5 & \cellcolor{gray!40}\textbf{99.2} & \cellcolor{yellow!25}97.6 & 23.4 & 6.2 & 6.2 & \cellcolor{orange!20}97.6 \\
no & \cellcolor{gray!40}\textbf{97.9} & \cellcolor{gray!40}97.2 & \cellcolor{yellow!25}79.6 & 6.9 & 7.2 & 6.8 & \cellcolor{orange!20}85.3 \\
pl & \cellcolor{gray!40}98.3 & \cellcolor{gray!40}\textbf{98.7} & \cellcolor{yellow!25}96.1 & 33.8 & 6.2 & 6.2 & \cellcolor{orange!20}96.6 \\
pt & \cellcolor{gray!40}98.2 & \cellcolor{gray!40}\textbf{99.0} & \cellcolor{yellow!25}96.3 & 9.6 & 6.2 & 6.2 & \cellcolor{orange!20}97.7 \\
ro & \cellcolor{gray!40}98.3 & \cellcolor{gray!40}\textbf{98.8} & \cellcolor{yellow!25}96.3 & 19.3 & 6.2 & 6.2 & \cellcolor{orange!20}96.9 \\
ru & \cellcolor{gray!40}98.1 & \cellcolor{gray!40}\textbf{98.6} & \cellcolor{yellow!25}95.8 & 11.0 & 7.5 & 8.7 & \cellcolor{orange!20}97.0 \\
sk & \cellcolor{gray!40}90.2 & \cellcolor{gray!40}\textbf{96.0} & \cellcolor{yellow!25}73.8 & 3.0 & 5.2 & 5.0 & \cellcolor{orange!20}93.9 \\
sv & \cellcolor{gray!40}98.3 & \cellcolor{gray!40}\textbf{98.7} & \cellcolor{yellow!25}94.8 & 6.9 & 6.6 & 6.4 & \cellcolor{orange!20}93.2 \\
sw & \cellcolor{gray!40}64.6 & \cellcolor{gray!40}\textbf{70.5} & \cellcolor{yellow!25}47.4 & 6.2 & 6.1 & 6.2 & \cellcolor{orange!20}24.8 \\
th & \cellcolor{gray!40}\textbf{84.6} & \cellcolor{gray!40}83.6 & \cellcolor{yellow!25}57.9 & 6.6 & 6.2 & 6.3 & \cellcolor{orange!20}65.3 \\
tr & \cellcolor{gray!40}\textbf{98.0} & \cellcolor{gray!40}98.0 & \cellcolor{yellow!25}94.8 & 13.5 & 6.2 & 6.3 & \cellcolor{orange!20}96.8 \\
uk & \cellcolor{gray!40}98.2 & \cellcolor{gray!40}\textbf{98.7} & \cellcolor{yellow!25}95.8 & 12.9 & 6.3 & 6.3 & \cellcolor{orange!20}97.1 \\
vi & \cellcolor{gray!40}98.2 & \cellcolor{gray!40}\textbf{98.7} & \cellcolor{yellow!25}95.0 & 15.2 & 6.2 & 6.3 & \cellcolor{orange!20}96.9 \\
zh & \cellcolor{gray!40}\textbf{98.1} & \cellcolor{gray!40}97.9 & \cellcolor{yellow!25}94.5 & 12.4 & 6.5 & 27.3 & \cellcolor{orange!20}97.3 \\
\midrule
Avg. & \cellcolor{gray!40}\textbf{90.2} & \cellcolor{gray!40}88.9 & \cellcolor{yellow!25}83.1 & 15.0 & 7.1 & 11.1 & \cellcolor{orange!20}84.2 \\
\bottomrule
\end{tabular}
\caption{Language steering scores (i.e. harmonic means of language forcing and output relevance scores) across all methods for 32 ablation languages for \texttt{Aya-Expanse-8B}. Language steering score is a harmonic mean of language forcing success and output relevance.}
\label{tab:steering_score_aya}
\end{table*}

\clearpage

\begin{figure*}[ht]
\section{Between-language Forcing Results}
\subsection{Llama-3.1-8B}
    \centering
    \includegraphics[width=0.7\linewidth]{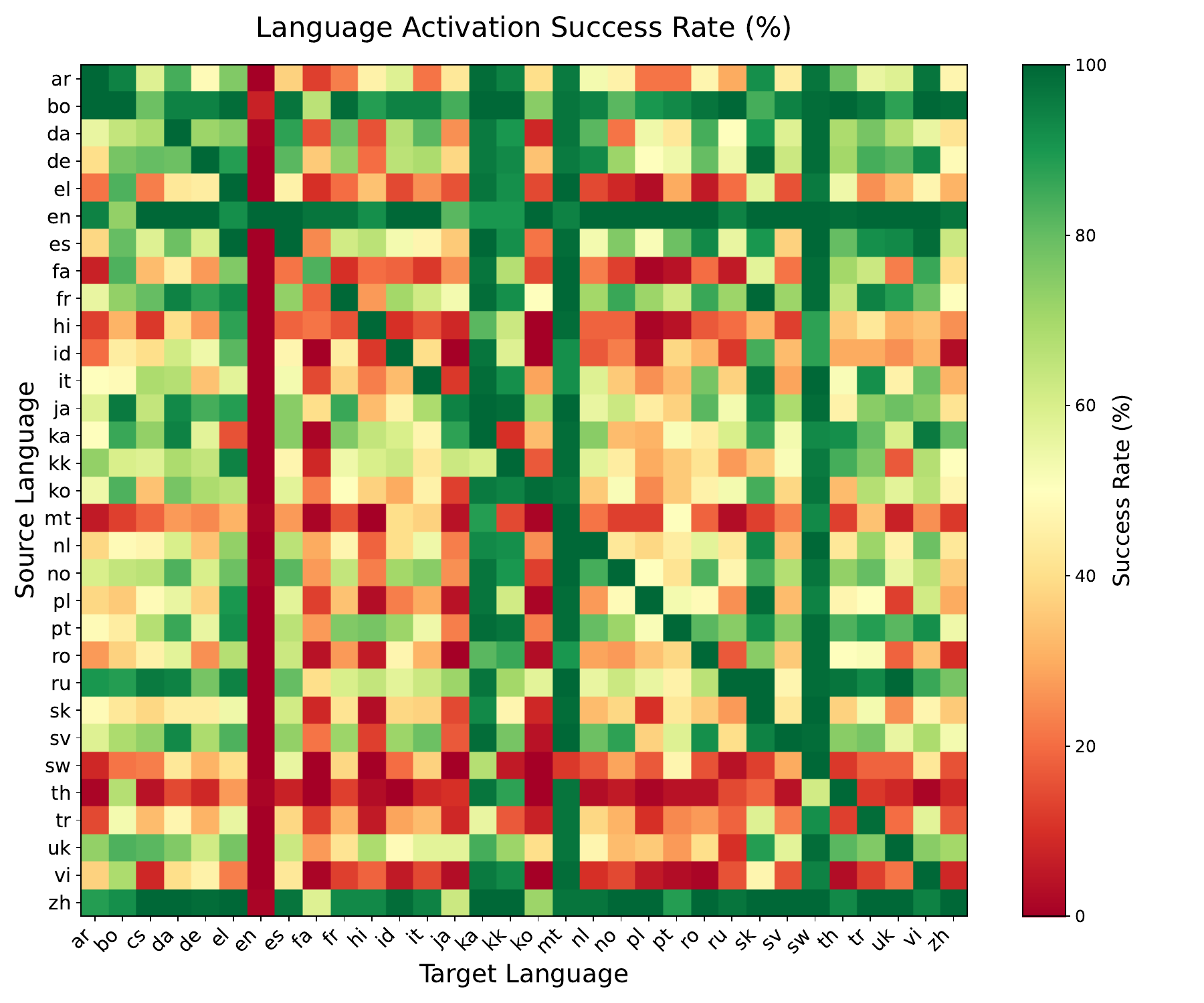}
    \caption{Between-language forcing scores for \textbf{Baseline-I} across 32 languages in \texttt{Llama-3.1-Instruct}. The matrix structure allows for tracing steerability in both directions: which languages are most amenable to being steered away from (rows) and which are most readily steered into (columns).}
    \label{fig:base_i_between}
\end{figure*}

\begin{figure*}[h]
    \centering
    \includegraphics[width=0.7\linewidth]{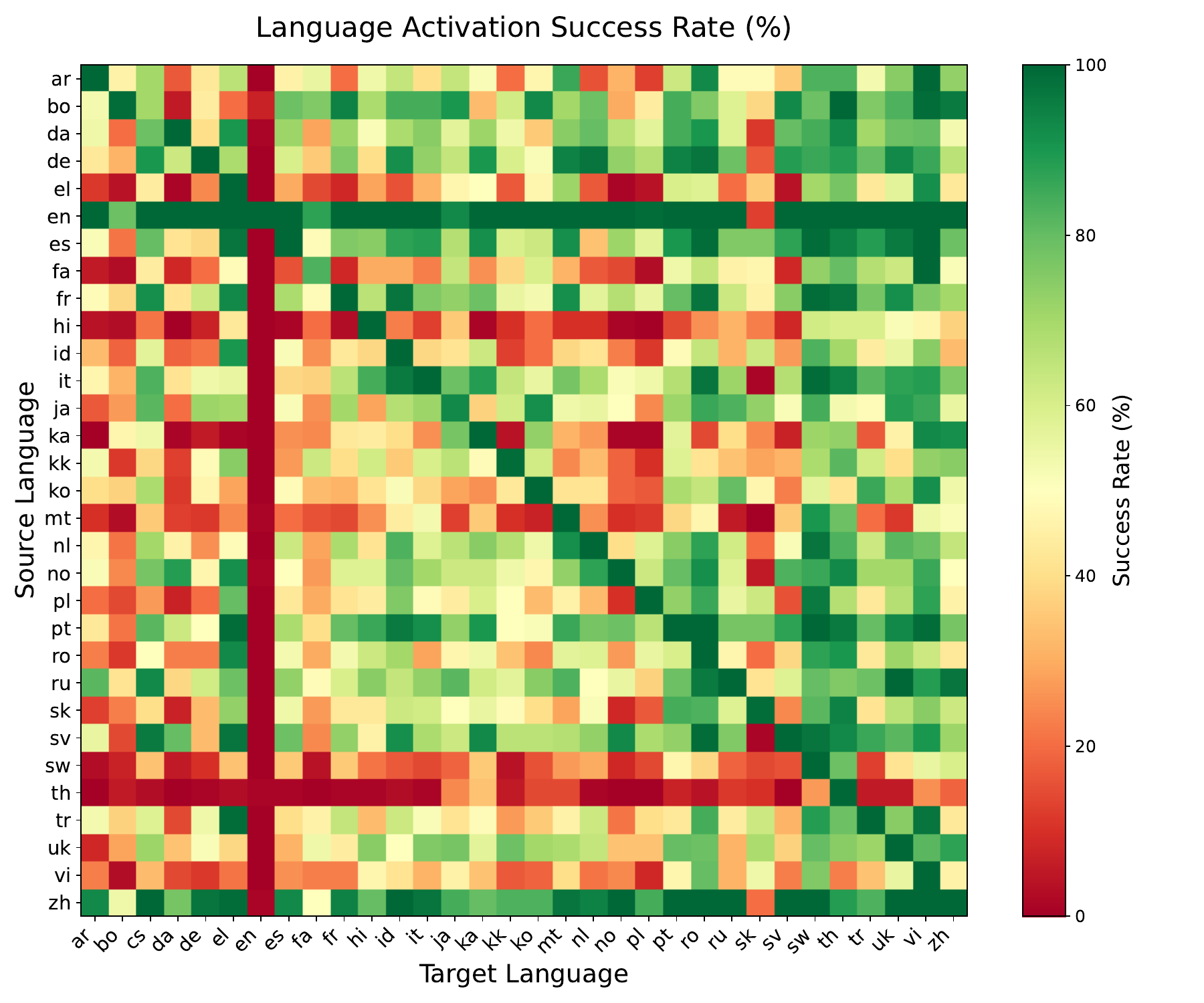}
    \caption{Between-language forcing scores for \textbf{Baseline-II} across 32 languages in \texttt{Llama-3.1-Instruct}. The matrix structure allows for tracing steerability in both directions: which languages are most amenable to being steered away from (rows) and which are most readily steered into (columns).}
    \label{fig:base_ii_between}
\end{figure*}

\begin{figure*}[ht]
    \centering
    \includegraphics[width=0.7\linewidth]{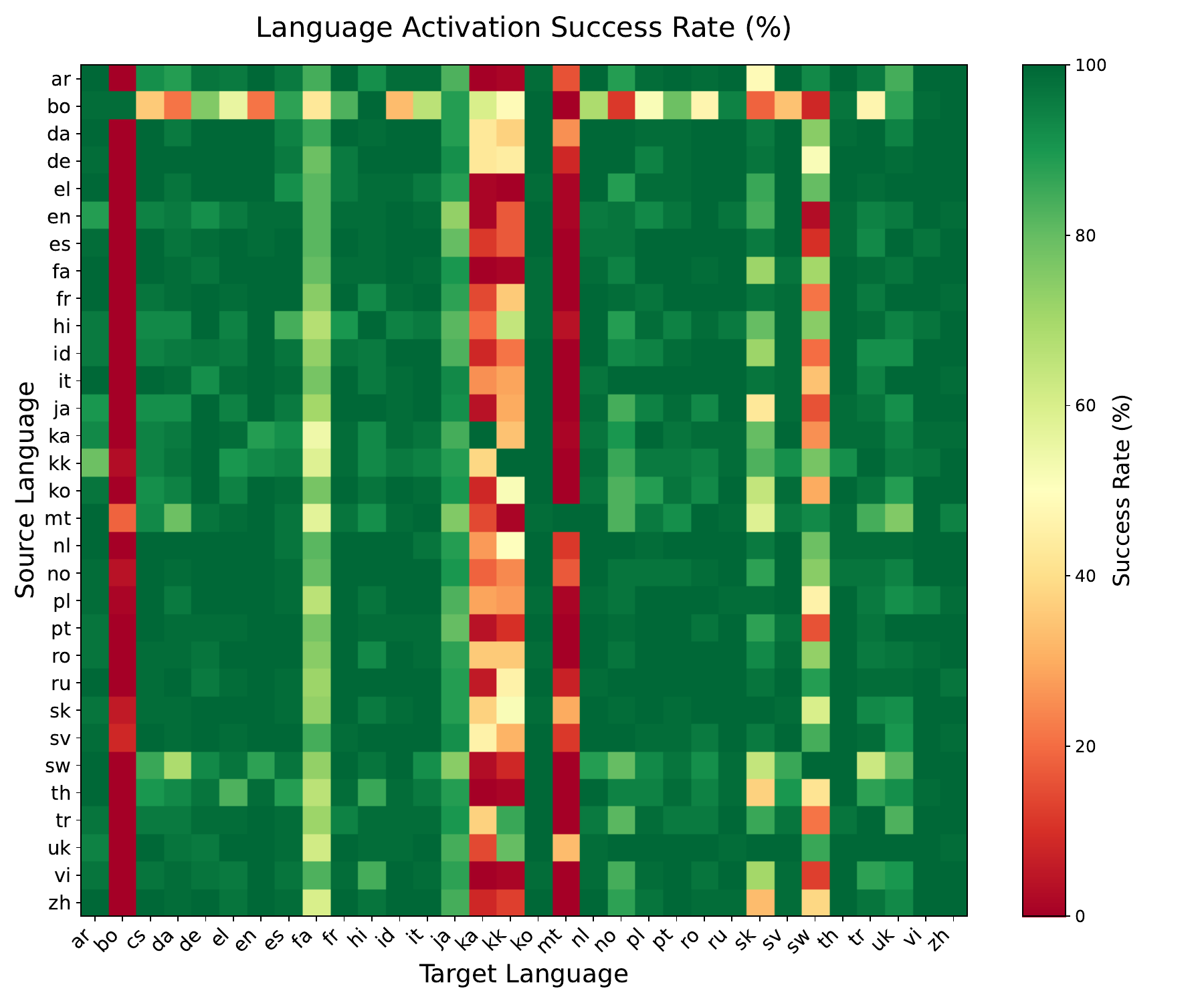}
    \caption{Between-language forcing scores for \textbf{DiffMean} across 32 languages in \texttt{Llama-3.1-Instruct}. The matrix structure allows for tracing steerability in both directions: which languages are most amenable to being steered away from (rows) and which are most readily steered into (columns).}
    \label{fig:diffmean_between}
\end{figure*}

\begin{figure*}[h]
    \centering
    \includegraphics[width=0.7\linewidth]{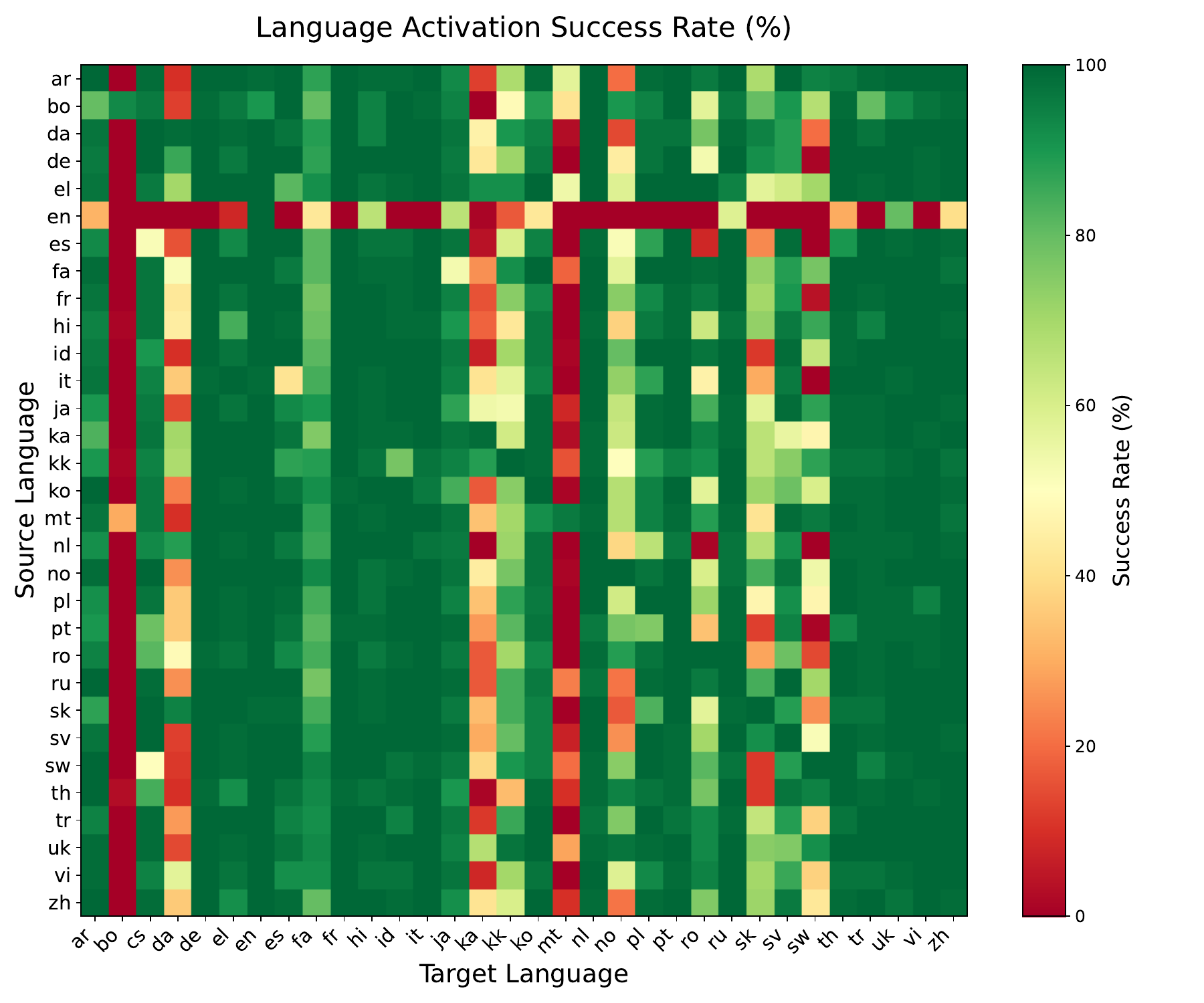}
    \caption{Between-language forcing scores for \textbf{LAPE} across 32 languages in \texttt{Llama-3.1-Instruct}. The matrix structure allows for tracing steerability in both directions: which languages are most amenable to being steered away from (rows) and which are most readily steered into (columns).}
    \label{fig:lape_between}
\end{figure*}

\begin{figure*}[ht]
\subsection{Aya-Expanse-8B}
    \centering
    \includegraphics[width=0.7\linewidth]{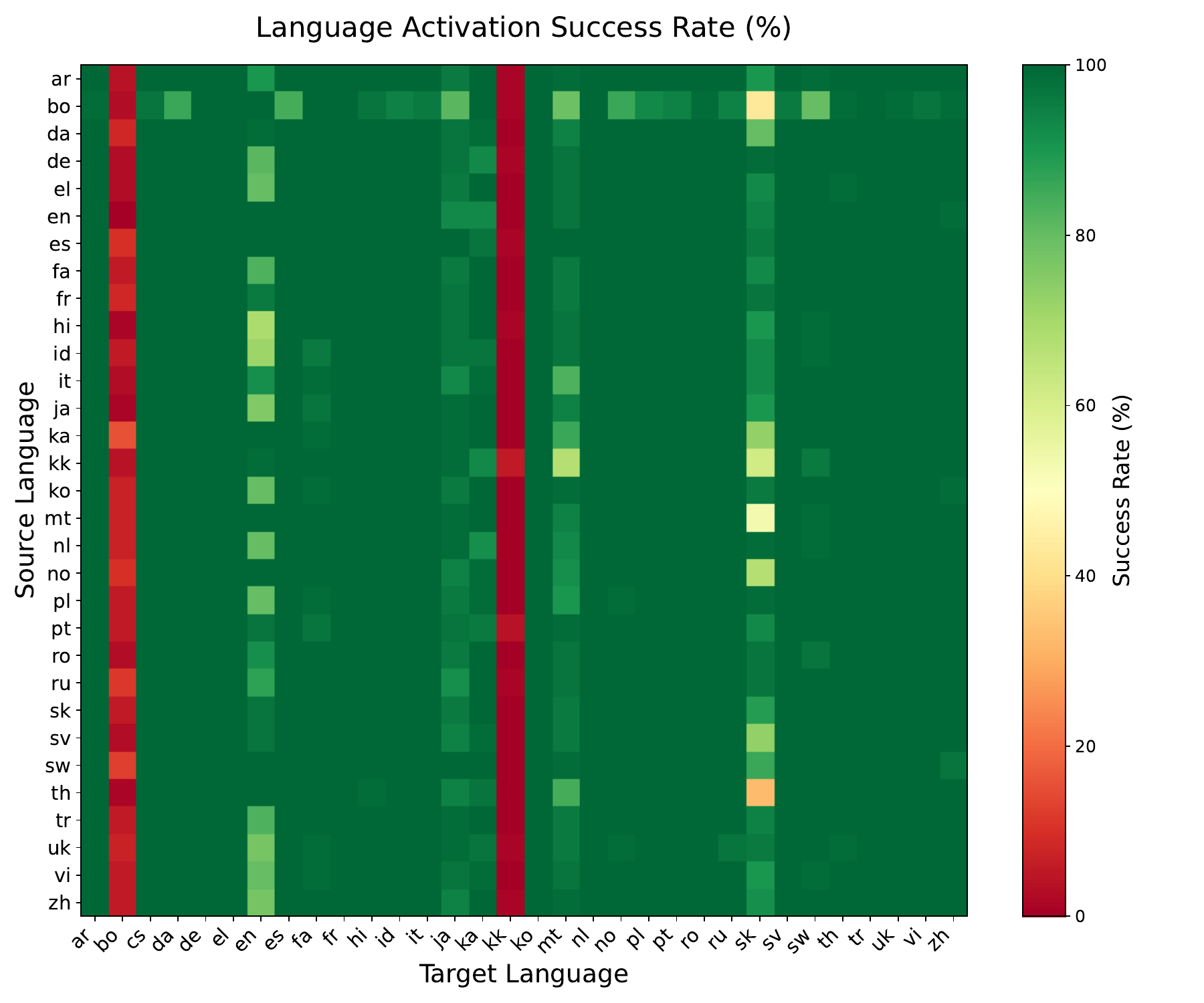}
    \caption{Between-language forcing scores for \textbf{Baseline-I} across 32 languages in \texttt{Aya-Expanse-8B}. The matrix structure allows for tracing steerability in both directions: which languages are most amenable to being steered away from (rows) and which are most readily steered into (columns).}
    \label{fig:base_i_between_aya}
\end{figure*}

\begin{figure*}[ht]
    \centering
    \includegraphics[width=0.7\linewidth]{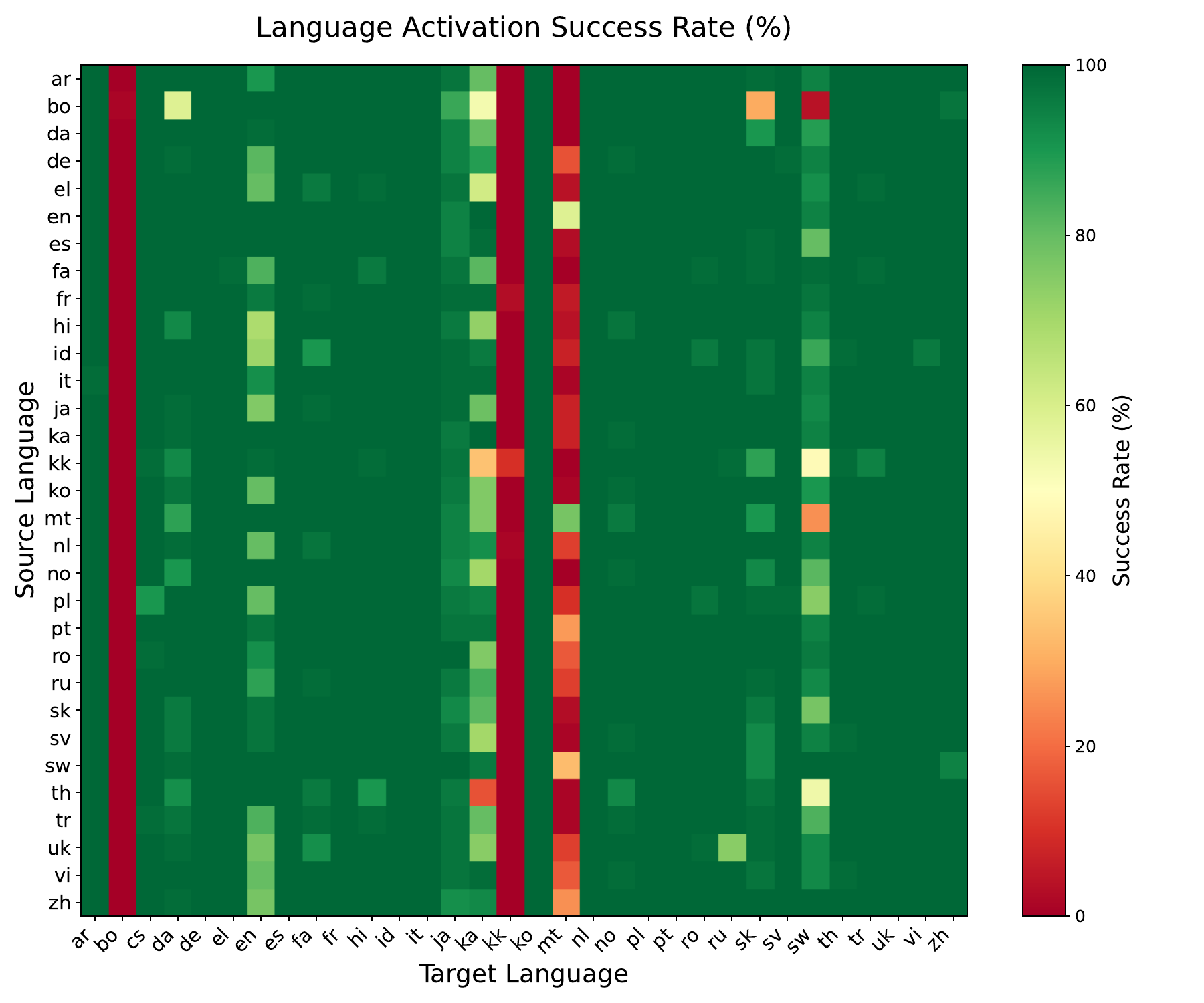}
    \caption{Between-language forcing scores for \textbf{Baseline-II} across 32 languages in \texttt{Aya-Expanse-8B}. The matrix structure allows for tracing steerability in both directions: which languages are most amenable to being steered away from (rows) and which are most readily steered into (columns).}
    \label{fig:base_ii_between_aya}
\end{figure*}

\begin{figure*}[ht]
    \centering
    \includegraphics[width=0.7\linewidth]{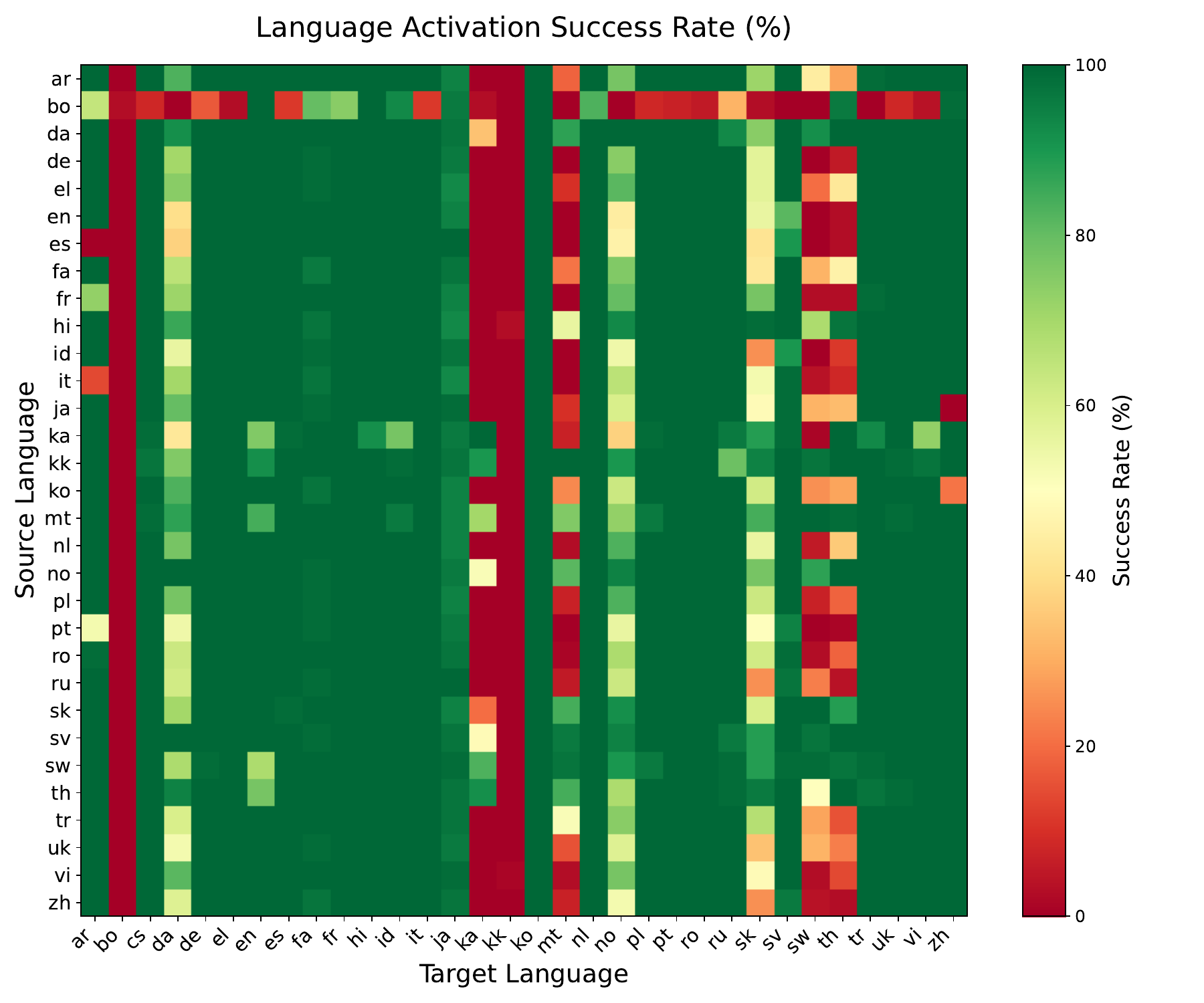}
    \caption{Between-language forcing scores for \textbf{DiffMean} across 32 languages in \texttt{Aya-Expanse-8B}. The matrix structure allows for tracing steerability in both directions: which languages are most amenable to being steered away from (rows) and which are most readily steered into (columns).}
    \label{fig:diffmean_between_aya}
\end{figure*}

\begin{figure*}[h]
    \centering
    \includegraphics[width=0.7\linewidth]{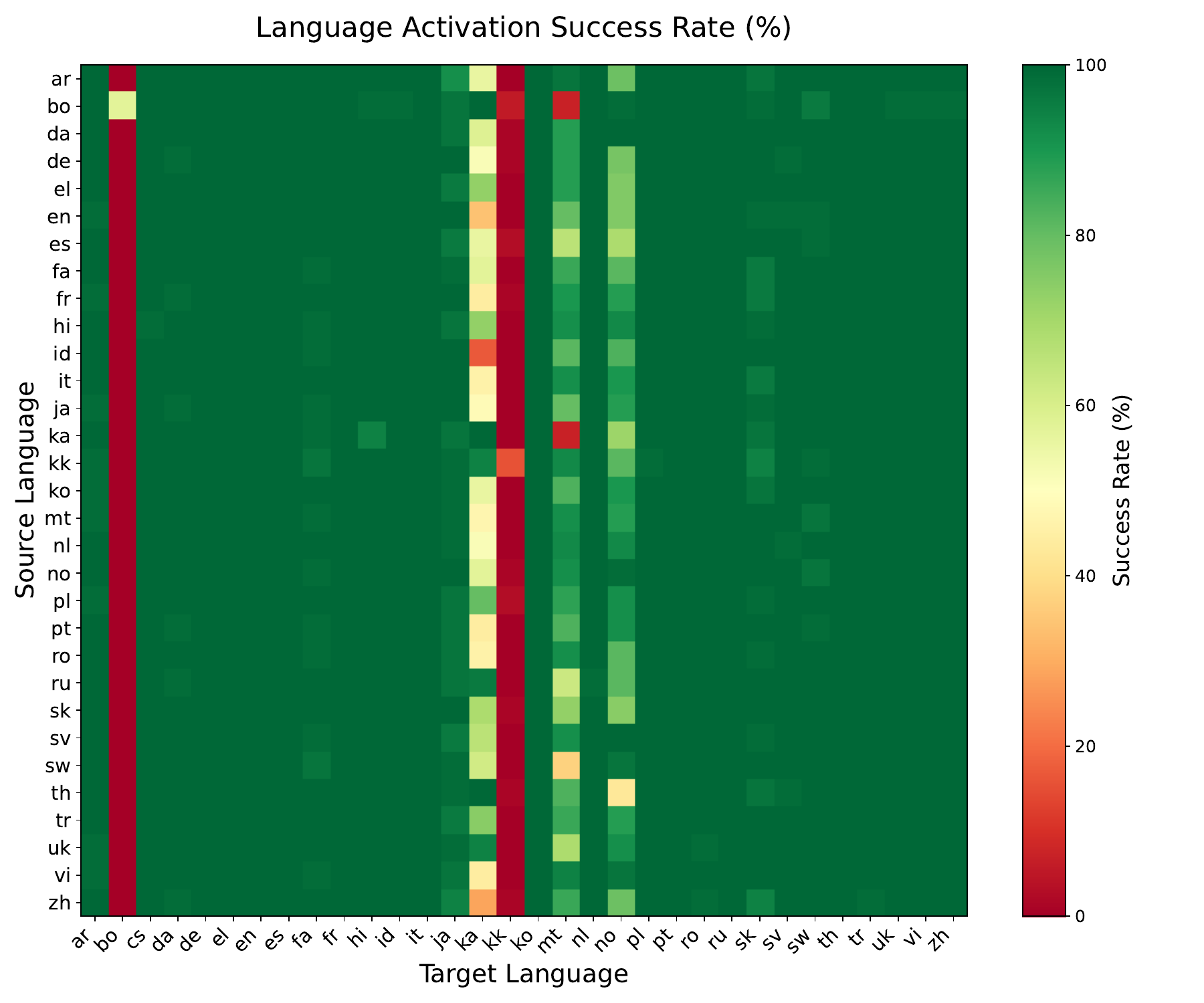}
    \caption{Between-language forcing scores for \textbf{LAPE} across 32 languages in \texttt{Aya-Expanse-8B}. The matrix structure allows for tracing steerability in both directions: which languages are most amenable to being steered away from (rows) and which are most readily steered into (columns).}
    \label{fig:lape_between_aya}
\end{figure*}

\clearpage

\begin{table*}[t]
\section{Sample Generations}
\label{app:examples}
\subsection{Llama-3.1-8B}
\centering
\footnotesize
\begin{tabular}{l|p{12cm}}
\toprule
\textbf{Method} & \textbf{Generated Output} \\
\midrule
\multicolumn{2}{p{14cm}}{\textbf{Input (Spanish):} ¿Cuáles son las formas más efectivas de lidiar con el estrés? \hfill \textbf{Target:} German} \\
\midrule
\cellcolor{red!20}$\mathcal{E}$ Baseline-I & Existen varias formas efectivas de lidiar con el estrés. Aquí te presento algunas de ellas: 1. **Meditación y respiración profunda**: La meditación... \\
\midrule
\cellcolor{red!20}$\mathcal{E}$ Baseline-II & Existen varias formas efectivas de lidiar con el estrés. Aquí te presento algunas de ellas: 1. **Meditación y respiración profunda**: La meditación... \\
\midrule
\cellcolor{green!20}$\vec{\Delta}$ DiffMean & Es ist wichtig, dass es verschiedene Strategien gibt, um mit Stress umzugehen. Hier sind einige der effektivsten Methoden: 1. **Atmungsübungen**: Die Atmung ist ein wichtiger Teil des Stress-Management-Systems... \\
\midrule
\cellcolor{red!20}$\mathbf{w}$ Probe & Existen varios Methodos efectivos para lidiar con el estrés. Aquí te presento algunos de los más comunes: 1. **Meditación und Yoga**... \\
\midrule
\cellcolor{green!20}$\mathbf{u}$ PCA & Existenten gibt es mehrere Strategien, um die Stressbewältigung zu verbessern. Hier sind einige der effektivsten Methoden: 1. 2023. \\
\midrule
\cellcolor{red!20}$\vec{\Delta}$ SAE-DM & Existen varias formas efectivas de lidiar con el estrés, algunas de las cuales incluyen: 1. **Praktikas de mindfulness**:... \\
\midrule
\cellcolor{red!20}$\mathbf{v}$ LDA & Lidiar con el estrés puede ser un desafío, pero hay varias formas efectivas de manejarlo. Aquí están algunas de las mejores estrategias: 1. **La respiración profunda**:... \\
\midrule
\cellcolor{green!20}$\odot$ LAPE & Es gibt verschiedene Möglichkeiten, mit Stress umzugehen. Hier sind einige der effektivsten Methoden: 1. **Meditation und tiefe Atmung**:... \\
\bottomrule
\end{tabular}
\caption{Sample outputs for steering from Spanish to German. Methods highlighted in \colorbox{green!20}{green} successfully produce German, and \colorbox{red!20}{red} indicates failure to switch from the source language.}
\label{tab:examples_1}
\end{table*}

\begin{table*}[t]
\centering
\footnotesize
\begin{tabular}{l|p{12cm}}
\toprule
\textbf{Method} & \textbf{Generated Output} \\
\midrule
\multicolumn{2}{p{14cm}}{\textbf{Input (Korean):} 시간 관리 능력을 향상시키는 방법은 무엇입니까? \hfill \textbf{Target:} English} \\
\midrule
\cellcolor{red!20}$\mathcal{E}$ Baseline-I & 시간 관리 능력을 향상시키는 방법은 다음과 같습니다. 1. **목표 설정**:... \\
\midrule
\cellcolor{red!20}$\mathcal{E}$ Baseline-II & 시간 관리 능력을 향상시키는 방법은 다음과 같습니다. 1. **목표 설정**:... \\
\midrule
\cellcolor{green!20}$\vec{\Delta}$ DiffMean & Time management skills to improve: 1. **Goal setting**: Clear goals help you focus on what's important and allocate your time accordingly. Set specific, measurable, achievable, relevant, and time-bound (SMART) goals... \\
\midrule
\cellcolor{green!20}$\mathbf{w}$ Probe & time management skills to improve. Here are some tips to help you improve your time management skills: 1. **Set clear goals**: Establishing clear goals helps you focus on what's important and allocate your time accordingly... \\
\midrule
\cellcolor{red!20}$\mathbf{u}$ PCA & 시간 관리는 시간을 관리하는 방법을 향상하는 방법을 향상하는 방법을...  \\
\midrule
\cellcolor{red!20}$\vec{\Delta}$ SAE-DM & 시간 관리 능력을 향상시키는 방법은 다음과 같습니다: 1. **목표 설정**:... \\
\midrule
\cellcolor{red!20}$\mathbf{v}$ LDA & 시간 관리 ability 향상시키는 방법은 다음과: 1. **Prioritize**: 할 일 목록을 만들고... \\
\midrule
\cellcolor{green!20}$\odot$ LAPE & Improving time management skills can be achieved through several strategies. Here are some effective methods:1. **Set clear goals**:... \\
\bottomrule
\end{tabular}
\caption{Sample outputs for steering from Korean to English. Methods highlighted in \colorbox{green!20}{green} successfully produce English, and \colorbox{red!20}{red} indicates failure to switch from the source language.}
\label{tab:examples_2}
\end{table*}

\begin{table*}[t]
\subsection{Aya-Expanse-8B}
\centering
\footnotesize
\begin{tabular}{l|p{12cm}}
\toprule
\textbf{Method} & \textbf{Generated Output} \\
\midrule
\multicolumn{2}{p{14cm}}{\textbf{Input (French):} Écrivez une revue de concert symphonique, en discutant de la performance de l'orchestre et de l'expérience globale du public. \hfill \textbf{Target:} Slovak} \\
\midrule
\cellcolor{green!20}$\mathcal{E}$ Baseline-I & Koncert symfonického orchestru: Hudobná cesta plná emocí. Minulý večer som málokoného zažil tak silný a zapamätateľný koncert ako bol ten, ktorý předvedol symfonický orchestr v našom meste... \\
\midrule
\cellcolor{green!20}$\mathcal{E}$ Baseline-II & **Recenzia symfonického koncerty: Hudobná podívaná plná emocí** Minulý večer som mala príjemnosť navštíviť symfonický koncert, ktorý sa stal nezabudnuteľným zážitkom pre všetkých přítomnych... \\
\midrule
\cellcolor{green!20}$\vec{\Delta}$ DiffMean & **Koncert symfonického orchestru: Hudba, ktorá uchvátila srdce** Minulý večer sa v našom meste konal neobyčajný hudební zážitek, ke ktorému sa sjednotili milovníci klasickej hudby z celého regionu... \\
\midrule
\cellcolor{red!20}$\mathbf{w}$ Probe & **Revue du Concert Symphonique : Une Soirée Musicale Inoubliable** Le week-end dernier, j'ai eu le privilège d'assister à un concert symphonique exceptionnel donné par l'Orchestre Philharmonique de la ville, qui a rempli la salle de concert jusqu'à sa capacité maximale... \\
\midrule
\cellcolor{red!20}$\mathbf{u}$ PCA & **Une soirée enivrante avec l'Orchestre Symphonique : Une fusion parfaite de la musique et de l'émotion** Le concert symphonique de la semaine dernière a été une expérience musicale inoubliable, offrant au public une performance exceptionnelle de l'Orchestre Symphonique local... \\
\midrule
\cellcolor{red!20}$\mathbf{v}$ LDA & **Revue du Concert Symphonique : Une Soirée Musicale Inoubliable** Le week-end dernier, j'ai eu le privilège d'assister à un concert symphonique exceptionnel donné par l'Orchestre Symphonique National dans le cadre de leur tournée annuelle... \\
\midrule
\cellcolor{green!20}$\odot$ LAPE & **Revúa koncert symfónický: Úžasná harmonía a vzrušujúce zážok pre všetkých** Posledný večer bol pre mne nepopirateľným zážitkom, kejý si pamätám na dĺžšie. Mám tu šťastnou príležitosť pozývať na koncert symfónického orchestru v nášho meste, a musím povedať, že to bolo úžasné preživenie umu a hudby... \\
\bottomrule
\end{tabular}
\caption{Sample outputs for steering from French to Slovak. Methods highlighted in \colorbox{green!20}{green} successfully produce Slovak, and \colorbox{red!20}{red} indicates failure to switch from the source language.}
\label{tab:examples_3}
\end{table*}

\begin{table*}[t]
\centering
\footnotesize
\begin{tabular}{l|p{12cm}}
\toprule
\textbf{Method} & \textbf{Generated Output} \\
\midrule
\multicolumn{2}{p{14cm}}{\textbf{Input (Italian):} Struttura un copione per un episodio di podcast che discuta l’influenza delle piattaforme di streaming sull’industria musicale. \hfill \textbf{Target:} Romanian} \\
\midrule
\cellcolor{green!20}$\mathcal{E}$ Baseline-I & Titlu: "Revoluția Streamingului: Cum modelează platformele digitală viitorul muzicii" Introducere (0:00 - 2:00) **Gazda:** Bună și bine ați venit la un nou episod al podcast-ului nostru, unde explorăm impactul tehnologiei asupra artei... \\
\midrule
\cellcolor{green!20}$\mathcal{E}$ Baseline-II & **Titlu: Revoluția Muzicală: Cum Platformele de Streaming Redescresc Industria** **Introducere (0:00 - 2:00)** **Prezentator:** Bună și bine ați venit la un nou episod al podcast-ului nostru, unde explorăm impactul tehnologiei asupra artei... \\
\midrule
\cellcolor{green!20}$\vec{\Delta}$ DiffMean & **Titlu: Revoluția Streamingului: Cum Platformele Online Redefinesc Industria Muzicală** **Introducere (0:00 - 2:00)** - **Prezentator (P):** Bună și bine ați venit la un nou episod al podcast-ului nostru, unde explorăm impactul tehnologiei asupra artei și culturii... \\
\midrule
\cellcolor{red!20}$\mathbf{w}$ Probe & **Titolo: La Rivoluzione dello Streaming: Come le Piattaforme Digitali Stanno Rimodellando la Musica** **Introduzione:** **Narratore:** Benvenuti in un nuovo episodio del nostro podcast, dove esploriamo l'impatto della tecnologia sulla musica... \\
\midrule
\cellcolor{red!20}$\mathbf{u}$ PCA & Titolo: "La Rivoluzione dello Streaming: Come le Piattaforme Online stanno Trasformando la Musica" Introduzione (0:00 - 2:00) **Presentatore:** Benvenuti a un nuovo episodio del nostro podcast, dove esploriamo l'impatto della tecnologia sulla musica... \\
\midrule
\cellcolor{red!20}$\mathbf{v}$ LDA & Titolo: "La Rivoluzione dello Streaming: Come le Piattaforme Online stanno Trasformando la Musica" Introduzione (0:00 - 2:00) **Presentatore:** Benvenuti a un nuovo episodio del nostro podcast, dove esploriamo l'impatto della tecnologia sulla musica... \\
\midrule
\cellcolor{green!20}$\odot$ LAPE & Titlu: "Revoluția Streamingului: Cum modelează platformele digitală viitorul muzicii" Introducere (0:00 - 2:00) **Gazda:** Bună și bine ați venit la un nou episod al podcast-ului nostru, unde explorăm impactul tehnologiei asupra artei... \\
\bottomrule
\end{tabular}
\caption{Sample outputs for steering from Italian to Romanian. Methods highlighted in \colorbox{green!20}{green} successfully produce Romanian, and \colorbox{red!20}{red} indicates failure to switch from the source language.}
\label{tab:examples_4}
\end{table*}